\documentclass{article} %
\usepackage{preprint,times}

\usepackage{url}
\usepackage{graphicx}   %
\usepackage{booktabs}   %
\usepackage{rotating}   %
\usepackage{placeins}
\usepackage{arydshln} %
\usepackage{mathtools}

\usepackage{gav_style}

\usepackage[normalem]{ulem}

\usepackage{xspace}
\newcommand{\testbed}{\textsc{TestBED}\xspace}

\usepackage{multirow}

\newcommand{\MeanSD}[2]{\ensuremath{#1 {\pm \scriptstyle #2}}}

\usepackage{enumitem}
\definecolor{tbframe}{RGB}{31,73,125}

\usepackage{xcolor}
\usepackage{listings}
\usepackage[listings,skins,most]{tcolorbox}

\usepackage{inconsolata}          %

\newcommand{\cls}[1]{{\ttfamily\color{codecls}#1}}

\definecolor{codebg}{HTML}{FBFBFD}
\definecolor{codeline}{HTML}{DFE1E8}
\definecolor{codefg}{HTML}{24292F}
\definecolor{codekw}{HTML}{8250DF}   %
\definecolor{codecls}{HTML}{B5560A}  %
\definecolor{codecom}{HTML}{6E7781}  %
\definecolor{codestr}{HTML}{0A7D40}
\definecolor{codenum}{HTML}{0550AE}

\lstdefinestyle{testbed}{
  language=Python,
  basicstyle=\ttfamily\scriptsize\color{codefg},
  keywordstyle=\color{codekw}\bfseries,
  commentstyle=\color{codecom}\itshape,
  stringstyle=\color{codestr},
  emph={LocationFinding,AttentivePolicyNet,PolicyDesigner,SPCE,%
        evaluate_checkpoint,Rollout},
  emphstyle=\color{codecls},
  showstringspaces=false,
  columns=fullflexible,
  keepspaces=true,
  upquote=true,
  aboveskip=0pt, belowskip=0pt,
  literate={0}{{\color{codenum}0}}1 {1}{{\color{codenum}1}}1
           {2}{{\color{codenum}2}}1 {3}{{\color{codenum}3}}1
           {4}{{\color{codenum}4}}1 {5}{{\color{codenum}5}}1
           {6}{{\color{codenum}6}}1 {7}{{\color{codenum}7}}1
           {8}{{\color{codenum}8}}1 {9}{{\color{codenum}9}}1,
}
\usepackage{titlesec}
\titlespacing*{\paragraph}
  {0pt}
  {0pt}
  {1em}

\newtcblisting{testbedcode}{
  listing only, listing style=testbed,
  colback=codebg, colframe=codeline,
  boxrule=0.6pt, arc=2.5pt,
  left=7pt, right=7pt, top=5pt, bottom=5pt,
}

\usepackage{times}
\usepackage{tikz}
\usetikzlibrary{arrows.meta, positioning, fit, backgrounds, calc}

\definecolor{taskfill}{HTML}{DCE8F4}
\definecolor{taskline}{HTML}{35618E}
\definecolor{polfill}{HTML}{D9EBE4}
\definecolor{polline}{HTML}{2A7361}
\definecolor{objfill}{HTML}{FBE6D2}
\definecolor{objline}{HTML}{B36B22}
\definecolor{evfill}{HTML}{E7E1F0}
\definecolor{evline}{HTML}{5F5290}
\definecolor{hubfill}{HTML}{2F3B52}
\definecolor{hookline}{HTML}{9A3B5E}
\definecolor{codebg}{HTML}{FAFAFC}
\definecolor{codeline}{HTML}{BFC3CE}
\definecolor{kw}{HTML}{1F6FB2}
\definecolor{cm}{HTML}{2A7361}
\definecolor{st}{HTML}{B3521F}

\newcommand{\sd}[1]{{\tiny\textcolor{black!62}{#1}}}

\title{TestBED: An Open-Source Toolkit for \\ Bayesian Experimental Design}

\author{Gavin Kerrigan\thanks{Correspondence to: \texttt{kerrigan@stats.ox.ac.uk}. Code for \testbed is available at \url{https://github.com/testbed-benchmark/testbed}.}, \, Marcel Hedman, Sahel Iqbal, \& Tom Rainforth \\
Department of Statistics, University of Oxford   \\
}

\iclrfinalcopy %
\begin{document}

\maketitle

\begin{abstract}

Bayesian experimental design (BED) is a principled framework for intelligent data acquisition with applications across a wide range of scientific and industrial domains. However, the development and adoption of BED remains hindered by a lack of standardised software and benchmarks. 
Existing methods are typically implemented in bespoke codebases and evaluated on a small subset of tasks, making methods difficult to apply, compare, or extend. To address this, we present \testbed, a general-purpose, open-source toolkit for Bayesian experimental design. 
\testbed decomposes the BED process into modular, reusable components spanning tasks, objectives, training mechanisms, policy architectures, and evaluation procedures, together with scalable implementations of core primitives such as expected information gain estimators.
We provide a wide range of these components, which can be freely composed, allowing existing methods to be reproduced within a common framework, while also enabling previously unexplored configurations and new methodologies to be easily implemented and evaluated.
Using \testbed, we further conduct a large-scale benchmark of contemporary BED methods, including controlled studies of under-appreciated but surprisingly critical methodological choices, such as the importance of time embeddings in design policy architectures. 
By providing a modular, extensible software foundation and a reproducible benchmarking suite, we hope that \testbed will accelerate methodological research in BED and lower the barrier to its real-world application.
\end{abstract}

\FloatBarrier

\section{Introduction}
\label{sect:intro}

In recent years, Bayesian experimental design~\citep[BED,][]{lindley1956measure,lindley1972bayesian,degroot1962uncertainty} methodology has developed rapidly, incorporating techniques from areas like amortised optimisation \citep{foster2021deep, ivanova2021implicit, rossa2026actionbed}, variational inference \citep{foster2019variational, huang2025aline, bracher2025jadai, phillips2026bayesian}, reinforcement learning \citep{huan2016sequential, blau2022optimizing, blau2023statistically,  barlas2025performance}, foundation models \citep{choudhury2026bed,huang2026efficient}, and generative models \citep{iollo2025bayesian, dong2025variational, saravanan2026diffbed, phillips2026bayesian}, amongst others. 
Beyond the research community, BED has seen uptake in a wide range of applications, including psychology \citep{watson2017quest+, aushev2023online, keurulainen2023amortised}, quantum physics \citep{varona2025computationally, belliardo2026multidimensional}, drug discovery \citep{park2013bayesian, chakraborty2025bridging, wang2026advanced}, and beyond.

However, many applications of BED have not leveraged these recent methodological advancements. A domain expert applying BED must often navigate multiple research codebases or re-implement methods from scratch, with little guidance as to which methods are best suited to a particular problem.
This lack of standardisation also bottlenecks progress within BED itself. 
Researchers developing new methods frequently re-build much of the surrounding infrastructure, %
incurring significant engineering costs. %
This burden is amplified by the range of settings that experimental design must support, like discrete outcomes \citep{kirby1996delay, cook2008optimal, vincent2017darc, valentin2024designing} and history-dependent models \citep{lim2022policy, iqbal2024nesting}, which can require special care for each method.
Furthermore, methods are often tested on different subsets or variants of benchmark problems, and details like experimental horizons, architectures, optimisation procedures, and others vary across papers. This makes results difficult to compare, and can obscure whether an apparent improvement is a result of methodological or implementation choices.

\definecolor{fitfill}{HTML}{ECECEF}
\definecolor{fitline}{HTML}{5C6170}

\lstnewenvironment{codeband}[2]{%
  \lstset{style=testbed,
    backgroundcolor=\color{#1},
    frame=l, framerule=1.4pt, rulecolor=\color{#2},
    framesep=4pt, xleftmargin=5.4pt, framexleftmargin=0pt,
    framextopmargin=1pt, framexbottommargin=1pt,
    aboveskip=0pt, belowskip=0pt,
    emph={LocationFinding,AttentivePolicyNet,PolicyDesigner,SPCE,Adam},
  }}{}

\begin{figure}[!t]
\centering
\begin{minipage}[c]{0.36\textwidth}
\centering
\begin{tikzpicture}[
  font=\sffamily,
  >={Stealth[length=1.8mm,width=1.3mm]},
  box/.style={rounded corners=2pt, draw, line width=0.6pt, align=center,
              inner sep=2.5pt, minimum width=2.5cm, minimum height=0.62cm,
              font=\sffamily\scriptsize},
  flow/.style={->, line width=0.7pt, draw=black!58},
  back/.style={->, line width=0.7pt, draw=objline, dashed,
               dash pattern=on 2pt off 1.5pt, rounded corners=2.5pt},
  hookarrow/.style={line width=0.5pt, draw=hookline!70, dotted},
  tag/.style={font=\sffamily\tiny, inner sep=1.2pt},
]
\node[box, fill=taskfill, draw=taskline] (task) at (0,0) {\texttt{Task}};
\node[box, fill=polfill,  draw=polline]  (des)  at (0,-1.0) {\texttt{Designer}};
\node[box, fill=hubfill,  draw=hubfill, text=white] (roll) at (0,-2.0)
  {\texttt{Rollout}\ {\tiny\textcolor{white!82}{$\theta, x_{1:T}, y_{1:T}$}}};
\node[box, fill=objfill,  draw=objline]  (obj)  at (0,-3.0) {\texttt{Objective}};

\draw[flow] (task) -- (des);
\draw[flow] (des) -- (roll);
\draw[flow] (roll) -- (obj);

\node[box, fill=fitfill, draw=fitline, minimum width=0.95cm, minimum height=0.62cm,
      font=\sffamily\tiny] (fit) at (1.95,-2.0) {\texttt{fit}\\$\nabla_\phi\widehat{\mathcal{I}}$};
\draw[back] (obj.east) -| (fit.south);
\draw[back] (fit.north) |- (des.east);

\node[font=\sffamily\tiny, text=hookline, align=right, anchor=east]
  (hook) at (-1.55,-1.0) {\texttt{hooks}};
\draw[hookarrow] (hook.east) -- (des.west);
\end{tikzpicture}
\end{minipage}\hfill
\begin{minipage}[c]{0.62\textwidth}
\begin{codeband}{taskfill}{taskline}
task = LocationFinding(n_sources=2, dim_theta=2)
\end{codeband}
\begin{codeband}{polfill}{polline}
policy_network = AttentivePolicyNet(
    task.event_shape_y, task.event_shape_design,
    event_shape_out=task.event_shape_design, dim_emb=64)
designer = PolicyDesigner(task, T=30, design_net=policy_network)
\end{codeband}
\begin{codeband}{objfill}{objline}
objective = SPCE(task, n_contrastive=2048)
\end{codeband}
\begin{codeband}{fitfill}{fitline}
optimiser = Adam(policy_network.parameters(), lr=1e-4)
for step in range(1000):
\end{codeband}
\begin{codeband}{taskfill}{taskline}
    theta = task.sample_prior(2048)
\end{codeband}
\begin{codeband}{polfill}{polline}
    rollout = designer.rollout(theta)
\end{codeband}
\begin{codeband}{objfill}{objline}
    loss = -objective(rollout)
\end{codeband}
\begin{codeband}{fitfill}{fitline}
    optimiser.zero_grad(); loss.backward(); optimiser.step()
\end{codeband}
\end{minipage}
\caption{\textbf{Left:} \testbed workflow. A \cls{Task} supplies the model and
design space, a \cls{Designer} produces a \cls{Rollout} of designs and outcomes, and an \cls{Objective} scores it. A fitting algorithm updates the designer's parameters. Hooks augment the rollout loop. \textbf{Right:} a complete DAD-style training loop~\citep{foster2021deep}, with each line shaded by the component it uses (imports omitted).}
\label{fig:testbed_system_diagram}
\label{fig:testbed_pseudocode}
\end{figure}

To address these challenges, we introduce \testbed, an open-source Python toolkit for BED. %
Built on PyTorch~\citep{paszke2019pytorch, ansel2024pytorch2}, \testbed is organised around two key design principles. The first is \emph{modularity}: we decompose the BED pipeline into a \textit{task} specifying the design problem to be solved, and three components which together define a BED methodology: a \textit{designer} specifying the strategy for producing designs, an \textit{objective} specifying what is optimised, and \textit{algorithms} which specify how the objective is optimised.
See \cref{fig:testbed_system_diagram}. Compatible components can be freely composed to obtain an end-to-end BED methodology for a given task. Beyond making existing methods easier to implement and compare, this decomposition enables new combinations not previously studied in the literature, several of which we study in our experiments, see~\cref{fig:coverage_grid}.

\looseness=-1
The second principle is \emph{extensibility}: each of these components is defined by a small abstract interface, so that new tasks, objectives, designers, architectures and methods can be developed without reimplementing the necessary surrounding infrastructure. A new expected information gain (EIG) estimator \citep{ao2024estimating, bartuska2024multilevel}, for instance, can be immediately evaluated across the available task suite, while a new application can make use of all compatible methods already implemented in \testbed. Thus, \testbed is not merely a collection of methods but rather a foundation on which new BED methodology and applications can be easily developed and tested.

Building on these principles, \testbed provides reference implementations of contemporary BED methods \citep{pmlr-v108-foster20a-unified, foster2021deep, blau2022optimizing, huang2025aline, rossa2026actionbed, phillips2026bayesian} in a common framework, together with reproducible evaluation protocols and recommended default configurations. We use \testbed to conduct what is, to our knowledge, the largest controlled empirical comparison of BED methods to date, exploiting its modularity to isolate the effect of individual architectural and computational choices, which existing evaluations cannot do.

\looseness=-1
While no single method dominates across tasks, our study yields several insights. %
First, time embeddings in the design policy have a large effect: earlier approaches that lack them \citep{foster2021deep} improve substantially once they are added. Second, a well-tuned static design is surprisingly strong on several popular tasks~\citep{iqbal2024nesting, phillips2026bayesian}, suggesting these tasks have limited power to discriminate between strategies. %
Third, variational joint posteriors~\citep{pmlr-v108-foster20a-unified,phillips2026bayesian}, a method rarely included as a baseline, is highly competitive when paired with modern architectures and comparable compute. %
These results show implementation choices can matter as much as methodological ones, and underscore the need for controlled, reproducible comparisons. Overall, \testbed provides the infrastructure needed to make Bayesian experimental design easier to use and improve. By providing a highly modular, extensible, and usable toolkit, \testbed lowers the cost of applying BED in new domains and can accelerate the development of new methodology.

\definecolor{covsame}{HTML}{6B7183}
\definecolor{covvar}{HTML}{2A7361}  %
\begin{figure}[!t]
\centering
\begin{tikzpicture}[
  font=\scriptsize,
  cell/.style={draw=white, line width=0.5pt, minimum width=0.56cm, minimum height=0.40cm, inner sep=0pt},
  same/.style={fill=covsame},
  new/.style={fill=covvar!30},
  combo/.style={fill=covvar, text=white, font=\tiny},
  na/.style={fill=black!3},
  collabel/.style={rotate=45, anchor=south west, inner sep=1pt},
  blocklabel/.style={font=\scriptsize\itshape, anchor=north, inner sep=2pt},
]
\node[collabel] at (0.280,0.04) {Location finding};
\node[collabel] at (0.840,0.04) {Temporal discounting};
\node[collabel] at (1.400,0.04) {CES};
\node[collabel] at (1.960,0.04) {Pharmacokinetic};
\node[collabel] at (2.520,0.04) {Michaelis--Menten};
\node[collabel] at (3.080,0.04) {Pendulum};
\node[collabel] at (3.640,0.04) {Cart pole};
\node[collabel] at (4.200,0.04) {Double pendulum};
\node[cell, same, anchor=south west] at (0.000,-0.400) {};
\node[cell, new, anchor=south west] at (0.560,-0.400) {};
\node[cell, same, anchor=south west] at (1.120,-0.400) {};
\node[cell, new, anchor=south west] at (1.680,-0.400) {};
\node[cell, new, anchor=south west] at (2.240,-0.400) {};
\node[cell, same, anchor=south west] at (2.800,-0.400) {};
\node[cell, same, anchor=south west] at (3.360,-0.400) {};
\node[cell, same, anchor=south west] at (3.920,-0.400) {};
\node[cell, same, anchor=south west] at (0.000,-0.800) {};
\node[cell, same, anchor=south west] at (0.560,-0.800) {};
\node[cell, new, anchor=south west] at (1.120,-0.800) {};
\node[cell, new, anchor=south west] at (1.680,-0.800) {};
\node[cell, new, anchor=south west] at (2.240,-0.800) {};
\node[cell, new, anchor=south west] at (2.800,-0.800) {};
\node[cell, new, anchor=south west] at (3.360,-0.800) {};
\node[cell, new, anchor=south west] at (3.920,-0.800) {};
\node[cell, same, anchor=south west] at (0.000,-1.200) {};
\node[cell, same, anchor=south west] at (0.560,-1.200) {};
\node[cell, same, anchor=south west] at (1.120,-1.200) {};
\node[cell, same, anchor=south west] at (1.680,-1.200) {};
\node[cell, new, anchor=south west] at (2.240,-1.200) {};
\node[cell, same, anchor=south west] at (2.800,-1.200) {};
\node[cell, same, anchor=south west] at (3.360,-1.200) {};
\node[cell, same, anchor=south west] at (3.920,-1.200) {};
\node[cell, same, anchor=south west] at (0.000,-1.600) {};
\node[cell, new, anchor=south west] at (0.560,-1.600) {};
\node[cell, same, anchor=south west] at (1.120,-1.600) {};
\node[cell, new, anchor=south west] at (1.680,-1.600) {};
\node[cell, new, anchor=south west] at (2.240,-1.600) {};
\node[cell, new, anchor=south west] at (2.800,-1.600) {};
\node[cell, new, anchor=south west] at (3.360,-1.600) {};
\node[cell, new, anchor=south west] at (3.920,-1.600) {};
\node[cell, same, anchor=south west] at (0.000,-2.000) {};
\node[cell, combo, anchor=south west] at (0.560,-2.000) {$\bigstar$};
\node[cell, combo, anchor=south west] at (1.120,-2.000) {$\bigstar$};
\node[cell, new, anchor=south west] at (1.680,-2.000) {};
\node[cell, new, anchor=south west] at (2.240,-2.000) {};
\node[cell, same, anchor=south west] at (2.800,-2.000) {};
\node[cell, same, anchor=south west] at (3.360,-2.000) {};
\node[cell, same, anchor=south west] at (3.920,-2.000) {};
\node[cell, same, anchor=south west] at (0.000,-2.400) {};
\node[cell, new, anchor=south west] at (0.560,-2.400) {};
\node[cell, same, anchor=south west] at (1.120,-2.400) {};
\node[cell, new, anchor=south west] at (1.680,-2.400) {};
\node[cell, new, anchor=south west] at (2.240,-2.400) {};
\node[cell, new, anchor=south west] at (2.800,-2.400) {};
\node[cell, new, anchor=south west] at (3.360,-2.400) {};
\node[cell, new, anchor=south west] at (3.920,-2.400) {};
\node[cell, same, anchor=south west] at (0.000,-2.800) {};
\node[cell, combo, anchor=south west] at (0.560,-2.800) {$\bigstar$};
\node[cell, combo, anchor=south west] at (1.120,-2.800) {$\bigstar$};
\node[cell, new, anchor=south west] at (1.680,-2.800) {};
\node[cell, new, anchor=south west] at (2.240,-2.800) {};
\node[cell, same, anchor=south west] at (2.800,-2.800) {};
\node[cell, new, anchor=south west] at (3.360,-2.800) {};
\node[cell, same, anchor=south west] at (3.920,-2.800) {};
\draw[black!40, line width=0.4pt] (0.000,-2.870) -- ++(4.480,0);
\node[blocklabel] at (2.240,-2.870) {Task};
\node[collabel] at (5.360,0.04) {Attention + time};
\node[collabel] at (5.920,0.04) {Attention};
\node[collabel] at (6.480,0.04) {DeepSets + time};
\node[collabel] at (7.040,0.04) {DeepSets};
\node[cell, na, anchor=south west] at (5.080,-0.400) {};
\node[cell, na, anchor=south west] at (5.640,-0.400) {};
\node[cell, na, anchor=south west] at (6.200,-0.400) {};
\node[cell, na, anchor=south west] at (6.760,-0.400) {};
\node[cell, na, anchor=south west] at (5.080,-0.800) {};
\node[cell, na, anchor=south west] at (5.640,-0.800) {};
\node[cell, na, anchor=south west] at (6.200,-0.800) {};
\node[cell, na, anchor=south west] at (6.760,-0.800) {};
\node[cell, new, anchor=south west] at (5.080,-1.200) {};
\node[cell, same, anchor=south west] at (5.640,-1.200) {};
\node[cell, new, anchor=south west] at (6.200,-1.200) {};
\node[cell, same, anchor=south west] at (6.760,-1.200) {};
\node[cell, new, anchor=south west] at (5.080,-1.600) {};
\node[cell, new, anchor=south west] at (5.640,-1.600) {};
\node[cell, new, anchor=south west] at (6.200,-1.600) {};
\node[cell, same, anchor=south west] at (6.760,-1.600) {};
\node[cell, na, anchor=south west] at (5.080,-2.000) {};
\node[cell, na, anchor=south west] at (5.640,-2.000) {};
\node[cell, na, anchor=south west] at (6.200,-2.000) {};
\node[cell, na, anchor=south west] at (6.760,-2.000) {};
\node[cell, na, anchor=south west] at (5.080,-2.400) {};
\node[cell, na, anchor=south west] at (5.640,-2.400) {};
\node[cell, na, anchor=south west] at (6.200,-2.400) {};
\node[cell, na, anchor=south west] at (6.760,-2.400) {};
\node[cell, same, anchor=south west] at (5.080,-2.800) {};
\node[cell, same, anchor=south west] at (5.640,-2.800) {};
\node[cell, new, anchor=south west] at (6.200,-2.800) {};
\node[cell, new, anchor=south west] at (6.760,-2.800) {};
\draw[black!40, line width=0.4pt] (5.080,-2.870) -- ++(2.240,0);
\node[blocklabel] at (6.200,-2.870) {Policy architecture};
\node[anchor=east, inner sep=3pt] (r0) at (0,-0.200) {Random};
\node[anchor=east, inner sep=3pt] (r1) at (0,-0.600) {Static {\color{black!55}\citep{foster2021deep}}};
\node[anchor=east, inner sep=3pt] (r2) at (0,-1.000) {DAD {\color{black!55}\citep{foster2021deep}}};
\node[anchor=east, inner sep=3pt] (r3) at (0,-1.400) {RL-BOED {\color{black!55}\citep{blau2022optimizing}}};
\node[anchor=east, inner sep=3pt] (r4) at (0,-1.800) {JointVarPost {\color{black!55}\citep{pmlr-v108-foster20a-unified}}};
\node[anchor=east, inner sep=3pt] (r5) at (0,-2.200) {ALINE {\color{black!55}\citep{huang2025aline}}};
\node[anchor=east, inner sep=3pt] (r6) at (0,-2.600) {ActionBED {\color{black!55}\citep{rossa2026actionbed}}};
\end{tikzpicture}
\par\vspace{2pt}
\begin{tikzpicture}[
  font=\scriptsize,
  cell/.style={draw=white, line width=0.5pt, minimum width=0.56cm, minimum height=0.40cm, inner sep=0pt},
  same/.style={fill=covsame},
  new/.style={fill=covvar!30},
  combo/.style={fill=covvar, text=white, font=\tiny},
  na/.style={fill=black!3},
  sw/.style={cell, minimum width=0.32cm, minimum height=0.22cm},
  lab/.style={anchor=west, inner sep=0pt},
]
\node[sw, same] (s0) at (0,0) {};
\node[lab, right=3pt of s0] (t0) {Prior work};
\node[sw, new, right=12pt of t0] (s1) {};
\node[lab, right=3pt of s1] (t1) {New in \testbed};
\node[sw, combo, right=12pt of t1] (s2) {$\bigstar$};
\node[lab, right=3pt of s2] (t2) {New methodological contribution};
\end{tikzpicture}
\vspace{-4pt}  %
\caption{\textbf{\testbed evaluates every method on every task, and enables new methodological contributions.} Columns show tasks (left) and design policy architectures (right). Grey: evaluated in prior work. Light green: an existing method, first evaluated in \testbed. Green ($\bigstar$): new methods, namely REINFORCE estimators extending \mbox{ActionBED} and \mbox{JointVarPost} to  discrete outcomes.}
\label{fig:coverage_grid}
\end{figure}
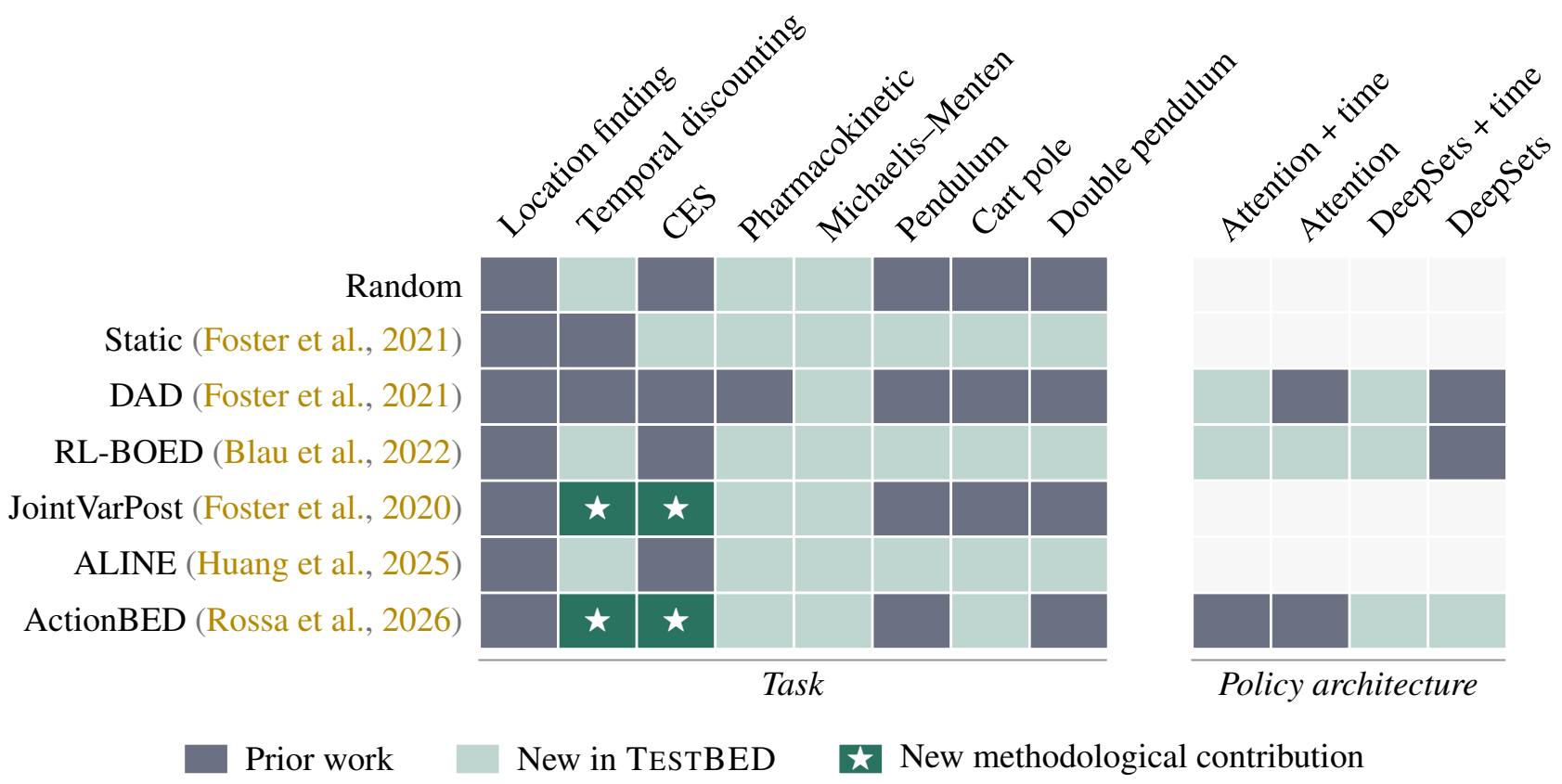

\section{Background}
\label{sect:background}

\looseness=-1
In BED, an experimenter seeks to learn about an unknown quantity $\theta \in \Theta$ through observations $y \in \YY$ which are influenced by a controllable design $x \in \XX$~\citep{lindley1956measure,lindley1972bayesian,degroot1962uncertainty,bernardo1979expected,chaloner1995bayesian,RYAN201445,rainforth2024modern,huan2024optimal}. The experimenter specifies a prior $p(\theta)$ and an observation model $p(y \mid \theta, x, h_t)$, where $h_t = \{ (x_s, y_s) : s  \leq t \}$ denotes a history of previous designs and observations.

\looseness=-1
In sequential (or adaptive) BED, $T$ experiments are performed adaptively, using the data $h_t$ from previous rounds to choose the next design $x_{t+1}$. This can be encoded as a \textit{policy} $\pi: h_t \mapsto x_{t+1}$.
Historically, BED methods followed an \textit{implicit} policy, where we calculate a posterior $p(\theta \mid h_t)$ then optimise the next design $x_{t+1}$ when it is needed~\citep{lindley1972bayesian,RYAN201445,foster2019variational}.
However, it has recently become the norm to learn an \emph{explicit} policy before the experiment~\citep{huan2016sequential, foster2021deep}, typically using a deep neural network.
This allows experiments to be run nearly instantaneously, while the learned policies often also outperform traditional approaches by learning to plan ahead and avoiding direct posterior approximation~\citep{foster2021deep}. 
Many techniques for fitting policies have been proposed, such as end-to-end training \citep{foster2021deep, ivanova2021implicit}, reinforcement learning \citep{huan2016sequential, blau2022optimizing, blau2023statistically, barlas2025performance}, or by joint training with a downstream action policy \citep{rossa2026actionbed}. These explicit policy-based approaches are the primary focus of our experiments, due to their dominance in the literature, but \testbed supports a broad range of alternative design procedures as well. 

\looseness=-1
Training a design policy requires an objective that assigns a value to the data it collects. The most common choice is the  expected information gain (EIG) in $\theta$ over the entire trajectory \citep{lindley1956measure}, %
\begin{equation} \label{eqn:eig}
    \II(\pi) \coloneq \E_{p(\theta, h_T \mid \pi)}\left[ \log p(h_T \mid \theta, \pi ) - \log p(h_T \mid \pi) \right],
\end{equation}
which is equivalently the mutual information between $\theta$ and the data $h_T$ collected by the policy $\pi$.
More generally, BED objectives may be defined as an expected utility~\citep{huan2024optimal} 
\begin{equation}
    \label{eq:expected-utility}
    \mathcal{U}(\pi) \coloneq \E_{p(\theta, h_T \mid \pi)}[U(\theta, h_T)],
\end{equation}
where the utility $U$ scores how useful the data $h_T$ are when the true parameter is $\theta$. The EIG is the special case $U(\theta, h_T) = \log p(h_T \mid \theta, \pi) - \log p(h_T \mid \pi)$. Other choices of $U$ recover many other notions of an experiment's value from the BED literature. Examples include the expected reward of a downstream action taken Bayes-optimally~\citep{raiffa1961applied, degroot1962uncertainty, lindley1972bayesian}, and the negative generalised posterior entropy induced by a proper scoring rule~\citep{degroot1962uncertainty, savage1971elicitation, dawid1998coherent, bickfordsmith2025rethinking, rossa2026actionbed}.

\looseness=-1
Many BED objectives are, however, difficult to compute. For example, the EIG depends on the typically intractable marginal likelihood $p(h_T \mid \pi)$. This has motivated a large literature on EIG estimators~\citep{ryan2003estimating, ba_bound, huan2013simulation, oord2018representation, belghazi2018mine, foster2019variational, poole2019variational}, which vary widely in what they require: some draw additional samples from the prior~\citep{rainforth2018nesting, foster2021deep}, while others train auxiliary networks~\citep{ba_bound, foster2019variational, kleinegesse2021gradient}. While accommodating this variety of objectives can be challenging, \testbed provides modular abstractions which enable any utility-based objective to be implemented without changing the rest of the pipeline.

\section{TestBED: An Open-Source Toolkit for BED}
\label{sect:TestBED}

\vspace{-2pt}
The core interface of \testbed consists of a \textit{task} specifying the underlying Bayesian model, and three components defining a BED method: a \textit{designer} specifying the strategy for producing designs (e.g., static or policy-based), an \textit{objective} specifying what is optimised (e.g., sPCE), and \textit{algorithms} which specify how the objective is optimised.
\cref{fig:testbed_system_diagram} illustrates each of these components and their interactions. The designer, objective, and algorithm can be freely composed and applied to any compatible task, allowing a single interface to be used across different problems.

This composability also extends hierarchically within each component. Each is itself built from modular subcomponents, for which \testbed provides reusable implementations. For example, \testbed provides several off-the-shelf and highly performant neural architectures which can be used to instantiate policy designers. These subcomponents are also highly extensible, allowing a user to develop new tools and methods without reimplementing the surrounding infrastructure.
We describe the main interface below, and refer to the \testbed documentation for lower-level details.

\begin{figure}[!t]
\centering
\begin{testbedcode}
class LocationFinding(Task):
    def __init__(self, n_sources, dim_theta, ..., prior=None, design_prior=None):
        super().__init__(
            event_shape_y = [1], 
            event_shape_theta = [n_sources, dim_theta],
            event_shape_design = [dim_theta],
            prior = prior or LocationFindingPrior(n_sources, dim_theta),
            design_prior = design_prior  # defaults to unconstrained, standard Gaussian
        )

        # task-specific initialisation (hyperparameters, ...)
        ...

    # inputs are a flat batch: inputs of shape [b, *event_shape]
    # histories are ignored as location finding is memoryless
    def _outcome_logp(self, y, theta, design, y_history, design_history):
        ...
    
    def _sample_outcome(self, theta, design, y_history, design_history):
        ...

# Example usage: TestBED handles arbitrary batch shapes
task = LocationFinding(n_sources=2, dim_theta=2)
theta = torch.randn(2, 2)                             # a single theta
design = torch.randn(10, 2)                           # a batch of 10 designs
y = task.sample_outcome(theta, design, n_samples=100) # [100, 10, 1]
\end{testbedcode}
\vspace{-4pt}
\caption{A simplified implementation of the location finding task in \testbed. Implementing a new task requires passing metadata (event shapes, prior, etc.) to the base \cls{Task}, and implementing methods for sampling outcomes and evaluating log-probabilities. Everything else, including broadcasting and safety checks, is handled by the base \cls{Task}, making new problems straightforward to implement.}
\vspace{-4pt}
\label{fig:task_pseudocode}
\end{figure}

\subsection{Core \testbed Interfaces}
\label{subsect:testbed-interfaces}

\paragraph{Tasks.} A \cls{Task} specifies the experimental design problem to be solved. This consists of a Bayesian model, including a prior $p(\theta)$, observation model $p(y \mid \theta, x, h_t)$. Tasks also specify a design prior $p(x)$, which can be used to produce random designs and whose support specifies the space of admissible designs. Tasks are highly flexible: parameters, designs, and observations may be continuous or discrete, models may be explicit or implicit, and may depend on the history.

Implementing a new task minimally requires specifying a mechanism for sampling observations $y \sim p(y \mid \theta, x, h_t)$. For explicit models, users may also provide the corresponding log-probability. Users may use PyTorch distributions or custom implementations, and \testbed provides re-usable implementations of common priors over designs and parameters. Computational machinery, such as broadcasting, is handled by \testbed, so that introducing a new task is straightforward. See \cref{fig:task_pseudocode}.

\looseness=-1
\paragraph{Designers and Rollouts.} A designer specifies how designs are produced over a sequence of $T$ experiments. The base \cls{Designer} class implements a common rollout procedure, where at each step $t$, it produces a design $x_t$, obtains an observation $y_t$ from the task, and updates a \cls{RolloutState} containing the current history $h_t$ of designs and observations via \texttt{state.y\_history} and \texttt{state.design\_history}, as well as any designer-specific information, e.g., stochastic policy probabilities. Concrete designers thus only need to implement the \texttt{\_step\_design(state)} function, specifying how the next design  is selected from the rollout state. After the final step, a designer returns a \cls{Rollout} which discards any information that is no longer needed, but keeps aspects needed downstream, like the history.

\looseness=-1
\testbed currently provides a \cls{RandomDesigner}, a \cls{StaticDesigner} that provides all $T$ designs jointly, a \cls{SequentialDesigner} that optimises a design at each step, and a \cls{PolicyDesigner} in which designs are produced using a neural network that maps the relevant parts of the rollout state to the next design.
The latter supports any contemporary policy-based method, including DAD~\citep{foster2021deep}, RL-BOED \citep{blau2022optimizing}, ALINE \citep{huang2025aline}, and ActionBED \citep{rossa2026actionbed}, amongst others. We additionally provide several off-the-shelf choices for policy architectures, including Attention \citep{vaswani2017attention} and DeepSets \citep{zaheer2017deepsets} based models, while allowing users to implement their own if so desired.

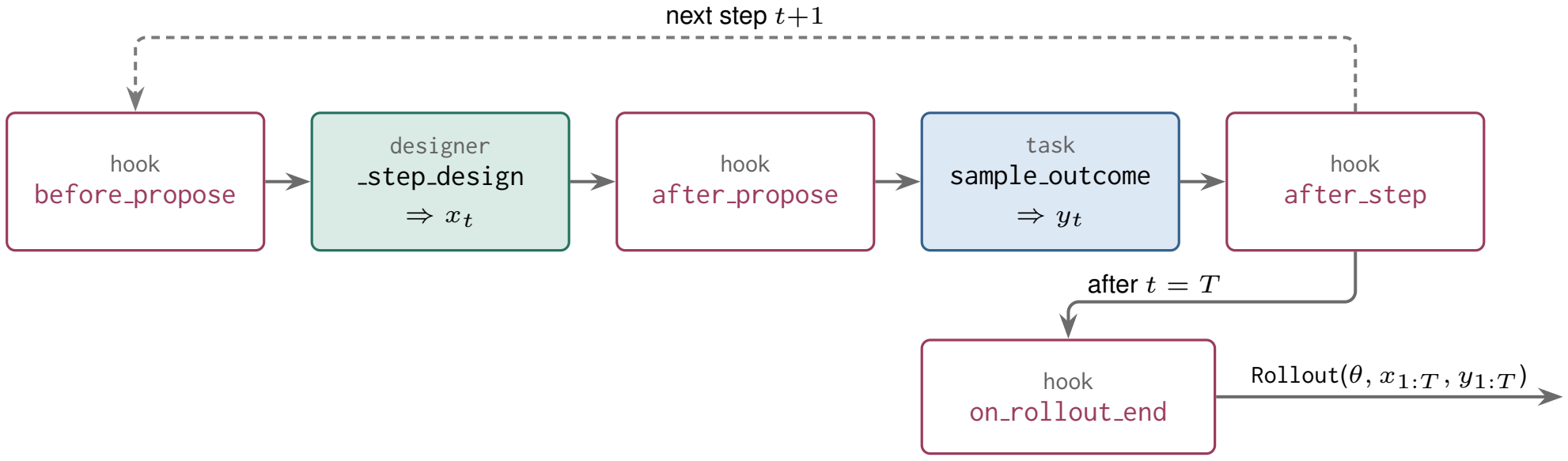
\begin{figure}[!t]
\centering
\begin{tikzpicture}[
  font=\sffamily,
  >={Stealth[length=2.1mm,width=1.5mm]},
  box/.style={
    rounded corners=2pt,
    draw,
    line width=0.6pt,
    align=center,
    inner sep=3.5pt,
    text centered,
    font=\sffamily\scriptsize,
    minimum width=2.15cm,
    minimum height=1.15cm
  },
  hookbox/.style={box, fill=white,    draw=hookline},
  desbox/.style={ box, fill=polfill,  draw=polline},
  taskbox/.style={box, fill=taskfill, draw=taskline},
  flow/.style={->, line width=0.75pt, draw=black!58},
  back/.style={
    ->,
    line width=0.75pt,
    draw=black!52,
    dashed,
    dash pattern=on 2pt off 1.5pt,
    rounded corners=2.5pt
  },
  tag/.style={font=\sffamily\tiny, inner sep=1.2pt}
]

\node[hookbox] (before) at (0,0)
{\sd{\texttt{hook}}\\[-0.5pt]
 \textcolor{hookline}{\texttt{before\_propose}}};

\node[desbox] (design) at (2.55,0)
{\sd{\texttt{designer}}\\[-0.5pt]
 \texttt{\_step\_design}\\[1pt]
 {\tiny$\Rightarrow x_t$}};

\node[hookbox] (afterprop) at (5.10,0)
{\sd{\texttt{hook}}\\[-0.5pt]
 \textcolor{hookline}{\texttt{after\_propose}}};

\node[taskbox] (outcome) at (7.65,0)
{\sd{\texttt{task}}\\[-0.5pt]
 \texttt{sample\_outcome}\\[1pt]
 {\tiny$\Rightarrow y_t$}};

\node[hookbox] (afterstep) at (10.20,0)
{\sd{\texttt{hook}}\\[-0.5pt]
 \textcolor{hookline}{\texttt{after\_step}}};

\draw[flow] (before.east)    -- (design.west);
\draw[flow] (design.east)    -- (afterprop.west);
\draw[flow] (afterprop.east) -- (outcome.west);
\draw[flow] (outcome.east)   -- (afterstep.west);

\draw[back]
  (afterstep.north) -- ($(afterstep.north)+(0,0.62)$)
  -- node[tag, above=0.4pt] {{next step $t{+}1$}}
     ($(before.north)+(0,0.62)$)
  -- (before.north);

\node[hookbox, minimum width=2.45cm, minimum height=0.95cm] (end) at (7.80,-1.80)
{\sd{\texttt{hook}}\\[-0.5pt]
 \textcolor{hookline}{\texttt{on\_rollout\_end}}};

\draw[flow, rounded corners=2.5pt]
  (afterstep.south) -- ($(afterstep.south)+(0,-0.42)$)
  -| node[tag, above=0.5pt, pos=0.35] {{after $t = T$}}
  (end.north);

\draw[flow]
  (end.east) -- node[tag, above=0.5pt] {\texttt{Rollout}{($\theta, x_{1:T}, y_{1:T}$})}
  ($(end.east)+(2.9,0)$);

\end{tikzpicture}
\vspace{-4pt}
\caption{\looseness=-1 Expanded view of the \texttt{Rollout} block of \cref{fig:testbed_system_diagram}. Each timestep alternates designer and
task calls, with hooks interleaved. All calls receive the current rollout state. After the final step and hook, the completed rollout is returned, including any extra fields produced by the hooks, e.g., actions.}
\vspace{-10pt}
\label{fig:rollout_expanded}
\end{figure}

For methods that modify the standard rollout, \testbed additionally exposes hooks at key stages of the design process.  \cref{fig:rollout_expanded} illustrates this. Hooks allow designers to incorporate additional behaviours and modifications by modifying the \cls{RolloutState} without needing to re-implement the complete design loop. For example, ActionBED~\citep{rossa2026actionbed} is implemented by attaching a hook that produces an action at the end of the rollout. %
Methods requiring substantial changes to the design loop can implement a new designer subclass, as we do for ALINE \citep{huang2025aline}.

\paragraph{Design Objectives.} Optimising designs requires an objective function. %
As described in~\cref{sect:background}, BED objectives can in general be described as an expected utility $\mathcal{U}(\pi) = \E_{p(\theta, h_T \mid \pi)}[U(\theta, h_T)]$. As these are expectations over the observed history, in \testbed, our objectives generally accept as input a \cls{Rollout} which at minimum contains the history $h_T$ alongside the true $\theta$ that can be used to estimate $\UU(\pi)$. For example, we provide highly efficient implementations of the standard sPCE and sNMC utilities \citep{pmlr-v108-foster20a-unified}, which in expectation give bounds on the EIG. See \cref{appendix_sect:spce_snmc}.

Utility functions are often complex, and may depend on additional information beyond the history. In this case, any auxiliary quantities needed to estimate the utility can be recorded in the \cls{Rollout}, or otherwise calculated within the design objective itself. Generally, computations that a user may want to control or modify (e.g., an action for loss-based utilities \citep{raiffa1961applied, degroot1962uncertainty, lindley1972bayesian, rossa2026actionbed} or a posterior approximation \citep{foster2019variational, phillips2026bayesian}) are calculated during the design loop and appended to the \cls{Rollout}, which the objective then directly uses. On the other hand, quantities a user is unlikely to want to configure (e.g,. contrastive samples when using the sPCE objective) are typically calculated within the objective itself. 
For methods which depend on utilities calculated before the terminal step $T$, an intermediate rollout may be produced using $h_t$ and passed in to the \cls{Objective} to calculate the intermediary utility. 

For non-reparameterisable models, \testbed also provides a general-purpose REINFORCE-style gradient estimator \citep{williams1992simple} that wraps these utilities. See \cref{appendix_sect:reinforce}. 
We emphasize that the modularity of \testbed allow a wide range of utilities, including 
alternative EIG estimators \citep{ao2024estimating, bartuska2024multilevel} or new objectives \citep{helin2025bayesian, kerrigan2026a} to be easily incorporated within a unified framework.

\paragraph{Algorithms.} \testbed additionally provides several algorithms for optimising the parameters of a designer, including end-to-end optimisation \citep{foster2021deep} and RL-based training \citep{blau2022optimizing}. We separate these from the designer, as optimisation introduces a further set of choices. Even for a static design, the design quality can depend on hyperparameter choices like initialisation, learning rates, treatment of constraints, and stopping criteria. Keeping these choices explicit avoids committing to a specific optimisation procedure, and allows users to develop alternative fitting procedures for the same underlying designer as needed. Our implementations also expose common reusable infrastructure, for checkpointing, logging and hyperparameter tuning, allowing new methods to reuse the surrounding machinery while only modifying the aspects of interest.

\looseness=-1
\paragraph{Recipes and Reproducibility.} Finally, \testbed provides a set of recipes, i.e., standardised configurations which combine the aforementioned components and specify choices such as policy architecture, optimiser settings, and other method-specific hyperparameters. \testbed includes reference implementations of methods from recent and representative policy-based methods, such as DAD \citep{foster2021deep}, RL-BOED \citep{blau2022optimizing}, ALINE \citep{huang2025aline}, ActionBED~\citep{rossa2026actionbed}, and variational posteriors \citep{foster2019variational, pmlr-v108-foster20a-unified, phillips2026bayesian}. The recipes used to produce the results in \cref{sect:experiments} are available within \testbed and provide reproducible, default configurations for each method. Configurations can be modified or swept independently, and are compatible with external experiment management tools like Hydra~\citep{yadan2019Hydra} and Weights \& Biases~\citep{wandb}.

In short, the abstractions in \testbed provide interfaces for the probabilistic model, design strategy, optimisation objective, and training procedure that specify a full BED pipeline, while providing the necessary infrastructure supporting them. This separation allows for existing approaches to be reproduced and compared within a common framework, while new tasks and methodologies can be introduced by modifying only the components in which they differ.

\section{Related Work}
\label{sect:related_work}

The closest work to a general-purpose BED library is the probabilistic programming language~(PPL) Pyro~\citep{bingham2019pyro}, whose experimental design module~\citep{pyro_oed_docs} implements several EIG estimators, including those of \citet{foster2019variational}. However, this module is limited to estimating the EIG of fixed designs, and has seen little development since 2020. Some of the early amortised methods were built on Pyro~\citep{foster2021deep, ivanova2021implicit}, but as standalone research code rather than as part of the library. More broadly, Pyro is a general-purpose PPL rather than a BED library, and provides no abstractions for design tasks or policies. %

Although many BED papers release code, general-purpose BED software remains scarce. Beyond Pyro, BED software is restricted to specific algorithms or model classes~\citep{overstall2020acebayes, nyberg2012poped}, classical non-amortised design~\citep{mcmichael2021optbayesexpt}, or inverse problems under Gaussian or linearised approximations~\citep{chowdhary2024pyoed}.
None of these supports amortised, policy-based BED, despite its now-central role in sequential design~\citep{rainforth2024modern}.

\section{Benchmarks and Experiments}
\label{sect:experiments}

We now use \testbed to conduct several controlled studies of contemporary BED methodology. Our aim is not primarily to introduce new methods, but rather to use \testbed to compare existing approaches under consistent conditions and to isolate methodological and implementation choices that are otherwise difficult to study across disparate codebases. 
We first conduct a large-scale benchmark of existing BED methods across eight diverse tasks. %
We then use the modularity of \testbed to study two factors that emerge as key, namely, the architecture of the design policy and the computational trade-offs involved in EIG-based training.
\cref{appendix_sect:additional_results} contains additional empirical results. %

\subsection{Benchmarking BED Algorithms}
\label{subsect:experiments_benchmark}

\looseness=-1
For our large-scale benchmark, we compare Joint Variational Posteriors~\citep[JointVarPost,][]{pmlr-v108-foster20a-unified, phillips2026bayesian}, Deep Adaptive Design~\citep[DAD,][]{foster2021deep}, RL-BOED~\citep{blau2022optimizing}, ALINE~\citep{huang2025aline}, and ActionBED~\citep{rossa2026actionbed}. %
We train ActionBED using two downstream losses (MSE and CMSE; see~\cref{appendix_sect:downstream_losses}), while the others all target EIG. Here, note that JointVarPost can in fact be seen as a special case of ActionBED, where actions are distributions on $\theta$ and the loss is targetting the EIG. We additionally study random and static (sPCE) \citep {pmlr-v108-foster20a-unified} baselines. 
Further details of each algorithm and their implementations are given in~\cref{appendix_sect:methods}.

All methods use the same tasks, with two exceptions. First, ALINE requires a finite pool of candidate designs, and handles continuous spaces by sampling candidate designs uniformly. %
Second, we found RL-BOED training becomes unstable when the design space is unbounded. Therefore, in line with the reference implementations \citep{blau2022optimizing, huang2025aline}, we modify the design space for both methods as needed, choosing bounds wide enough that they do not meaningfully restrict the method. For other hyperparameters, such as learning rate, weight decay and gradient clipping, we follow the original authors' recommendations where available.

\begin{figure}[!t]
    \centering
    \includegraphics[]{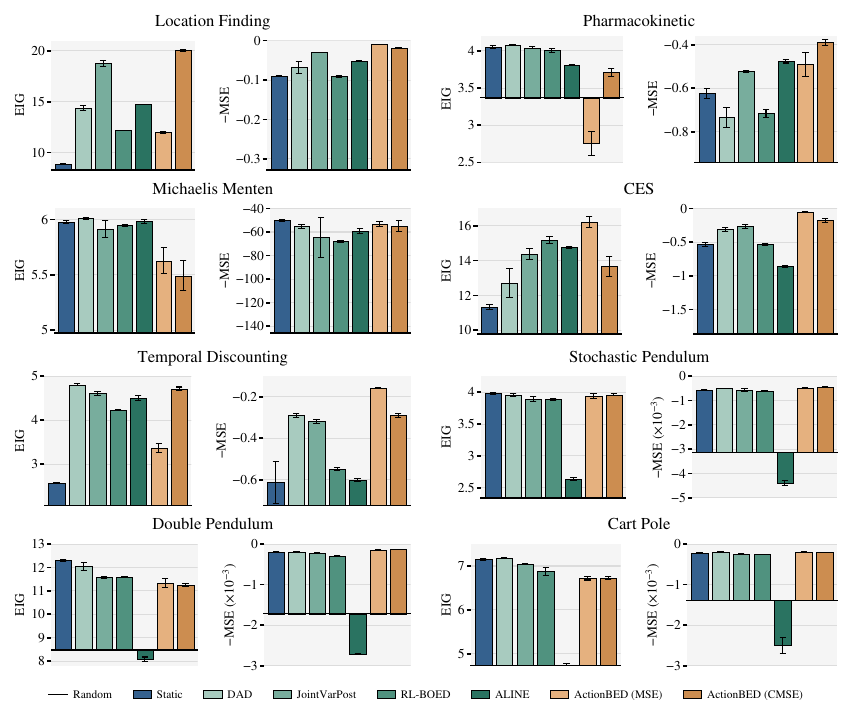}
    \caption{Benchmark (mean $\pm$ standard error across five seeds) of BED methods under a shared policy architecture and wall-clock training budget. We report EIG upper bounds ($L=10^9, N=2{,}048$), together with downstream MSE (negated so  higher is better). The horizontal black line indicates performance of a random policy.}
    \label{fig:benchmark_barchart}
    \vspace{-4ex}
\end{figure}

\looseness=-1

\looseness=-1
Following recent work~\citep{huang2025aline, huang2026efficient, rossa2026actionbed, phillips2026bayesian}, we use a common Attention-based policy architecture~\citep{vaswani2017attention} wherever possible. ALINE~\citep{huang2025aline} is also Attention-based but requires a specialised architecture, so we closely match its parameter count instead. Some methods train additional networks alongside the design policy, so we cannot match total parameter counts exactly. However, giving every design policy the same capacity still provides a fair basis for comparison. We also normalise computational cost by training every method for 12 hours on identical hardware~(an NVIDIA GH200 Superchip). This is a generous budget by the standards of the BED literature, so that no method is disadvantaged by under-training.

Results are shown in \cref{fig:benchmark_barchart}, with full numerical results in \cref{tab:benchmark} and \cref{tab:posthoc_actions} in the Appendix. We plot the sNMC upper bound~\citep{foster2019variational, foster2021deep} of the EIG~(\cref{appendix_sect:spce_snmc}) with $L=10^9$ contrastive samples and $N=2{,}048$ outer samples, and \cref{tab:benchmark} contains matching sPCE lower bounds. This evaluation budget is far larger than is typical, giving EIG bounds substantially tighter than those previously reported and hence high-confidence estimates of the true information gained by each method.
We additionally measure downstream utility for each design policy by fitting a small MLP to predict $\theta$ from the complete rollout $h_T$ using a mean-squared-error (MSE) loss. Although ActionBED fits such an MLP during training, we re-fit it at evaluation time for the sake of a fair comparison. See \cref{appendix_sect:downstream_losses} for further details. %

\looseness=-1
No single method dominates across all tasks and metrics. Notably, overlooked methods like JointVarPost~\citep{pmlr-v108-foster20a-unified, phillips2026bayesian} and even the static baseline perform strongly on several tasks. Rankings by EIG and loss do not always agree, reinforcing the complementarity of both measures. More surprisingly, a method trained on a different objective can outperform methods directly trained on the evaluation metric. On pharmacokinetics, e.g., ActionBED (CMSE) has a lower downstream MSE than ActionBED (MSE). Understanding which training objectives yield good designs, and how they relate to downstream objectives, is an important direction for future work.

We see that ALINE often struggles, attaining worse performance than random on several tasks. Note that \cref{tab:aline-reproduction} in the Appendix shows our implementation closely matches its published results. While ALINE introduced several advances to BED, including the inspiration for the Attention-based architecture we use here, 
we hypothesize that ALINE's restriction to a finite pool of candidate designs can be limiting in tasks with continuous design spaces, like the $3$ dynamical systems tasks. In addition, although ALINE's stated objective is a lower bound on the total information gain over the horizon, its policy-gradient loss~\citep[Eq. 11]{huang2025aline} weights each design only by its immediate one-step reward, and the trained policy is hence myopic.%

\subsection{The Effects of Architecture}
\label{subsect:arch_experiments}

\begin{table}[t]
\centering
\caption{Architecture ablation on the location finding task. Entries report the mean $\pm$ standard error across five seeds of the sPCE and sNMC bounds of the EIG ($L=10^9, N=2\,048$).}
\label{tab:benchmark-architecture}
\vspace{1ex}
\resizebox{\columnwidth}{!}{%
\begin{tabular}{l cccc cccc}
\toprule
& \multicolumn{4}{c}{Attention} & \multicolumn{4}{c}{DeepSets} \\
\cmidrule(lr){2-5} \cmidrule(lr){6-9}
& \multicolumn{2}{c}{$+$ time} & \multicolumn{2}{c}{$-$ time}
& \multicolumn{2}{c}{$+$ time} & \multicolumn{2}{c}{$-$ time} \\
\cmidrule(lr){2-3} \cmidrule(lr){4-5} \cmidrule(lr){6-7} \cmidrule(lr){8-9}
Method & Lower & Upper & Lower & Upper & Lower & Upper & Lower & Upper \\
\midrule
DAD
& \MeanSD{14.334}{0.223} & \MeanSD{14.359}{0.228}
& \MeanSD{5.654}{1.061} & \MeanSD{5.654}{1.061}
& \MeanSD{15.423}{0.265} & \MeanSD{15.587}{0.307}
& \MeanSD{12.051}{0.028} & \MeanSD{12.052}{0.028} \\
RL-BOED
& \MeanSD{12.188}{0.022} & \MeanSD{12.190}{0.022}
& \MeanSD{12.121}{0.034} & \MeanSD{12.122}{0.034}
& \MeanSD{12.017}{0.009} & \MeanSD{12.018}{0.009}
& \MeanSD{12.058}{0.016} & \MeanSD{12.059}{0.016} \\
ActionBED
& \MeanSD{18.563}{0.068} & \MeanSD{20.034}{0.114}
& \MeanSD{9.418}{0.496} & \MeanSD{9.418}{0.497}
& \MeanSD{15.818}{0.548} & \MeanSD{16.144}{0.632}
& \MeanSD{9.579}{0.091} & \MeanSD{9.579}{0.091} \\
\bottomrule
\end{tabular}%
}
\end{table}

Earlier BED methods~\citep[e.g.,][]{foster2021deep, blau2022optimizing} often used the DeepSets architecture~\citep{zaheer2017deepsets} for the design policy. More recent methods~\citep{huang2024amortized, huang2025aline, rossa2026actionbed, phillips2026bayesian} instead use an Attention-based architecture, first introduced by \citet{ivanova2021implicit}. Recent methods also add a time embedding to the design policy, even for exchangeable tasks~\citep{huang2024amortized}. Here, we investigate the degree to which recent advances in BED can be attributed to these architectural choices rather than to methodological advances.

We follow the setup from \cref{subsect:experiments_benchmark}, varying only the design network architecture on the location finding task, as this showed clear discrimination between methods. We compare both Attention- and DeepSets-based design policies, with and without time embeddings. To our knowledge, the combination of DeepSets with time embeddings has not been previously studied in the BED literature.

\looseness=-1
Results are reported in \cref{tab:benchmark-architecture}.
DAD performs better with DeepSets, whereas ActionBED
favours Attention. Throughput does not explain this, since Attention reduces the
number of updates by a similar factor for both methods. Rather, DAD's training
objective saturates: sPCE with $L=2{,}000$ is bounded by $\log(L+1)\approx 7.6$
nats, which both architectures approach within the first hour, so the objective can no longer resolve differences between policies. ActionBED's loss has no such ceiling.
With DeepSets it plateaus within four hours, whereas with Attention it continues
to improve throughout training. We study the related trade-off between EIG-estimation cost and optimisation throughput in \cref{subsect:L_experiments}.

\looseness=-1
Adding explicit timestep information clearly improves performance for both DAD and ActionBED. Although prior work has adopted time embeddings without isolating their effect, the improvement has a principled explanation: a sequential design problem with finite horizon $T$ is a finite-horizon Markov decision process (MDP)~\citep{huan2016sequential, blau2022optimizing, shen2023bayesian}, whose state at step $t$ is the belief $p(\theta \mid h_t)$. In finite-horizon MDPs, the optimal policy is generally \textit{time-dependent}, and an optimal stationary policy need not exist~\citep[Chapter~4]{puterman1994markov}. Intuitively, the value of an experiment depends not only on the current belief but also on how many experiments remain. Without explicit time information, the policy must infer $t$ from the number of observations in $h_t$, which common modules such as pooling and layer normalisation can make difficult~\citep{yehudai2024can}.

The performance of RL-BOED is similar across all combinations, but significantly worse than that of DAD or ActionBED with time embeddings. RL-BOED is trained with Soft Actor-Critic~\citep{haarnoja2018softa}, which uses dense per-step rewards and learned critic networks. These may allow it to perform time-specific credit assignment effectively even when the policy has no explicit time input.

\subsection{EIG as a Training Objective}
\label{subsect:L_experiments}

Finally, we study the computational trade-offs involved in using contrastive EIG objectives for policy learning. Methods optimising the EIG, like DAD, often rely on the sPCE estimator which is upper bounded by $\log(L+1)$~\citep{foster2021deep}. Increasing $L$ reduces saturation and potentially gives more signal, but also makes each optimisation step more expensive. Under a fixed compute budget, this creates a direct trade-off between estimator quality and the number of gradient updates.

To study this, we train DAD on location finding for a fixed budget of $3$ hours while varying $L$. The results in~\cref{tab:compute-normalized-spce-saturation} show that increasing $L$ substantially improves the learned policy up to a plateau. Policies trained with $L$ between $4{,}096$ and $16{,}384$ are statistically indistinguishable across five seeds, even though $L = 16{,}384$ completes roughly half as many gradient steps.

\looseness=-1
The results reveal an asymmetry. Under a fixed compute budget, choosing $L$ too small is far more damaging than choosing it too large. Training with $L = 256$ yields policies roughly $2$ nats below the plateau, whereas we observed no measurable penalty for the largest $L$ tested. %
The optimal value will depend on the task and compute budget, but these results suggest prioritising sufficiently accurate estimates of the EIG during training over more updates.

\begin{table}[!t]
    \vspace{-1ex}
    \small
    \centering
    \caption{Effect of the number of contrastive samples $L$ used for training DAD on location finding. Entries report mean $\pm$ standard error across five training seeds. All policies are evaluated using $2{,}048$ outer samples and $10^9$ contrastive samples.}
    \vspace{1ex}
    \begin{tabular}{cccc}
        \toprule
        $L$ & Lower & Upper & Average number of gradient steps \\
        \midrule
           256 & $12.568 \pm 0.084$ & $12.571 \pm 0.085$ & $28{,}355.8$ \\
           512 & $12.896 \pm 0.111$ & $12.898 \pm 0.111$ & $28{,}778.2$ \\
          1{,}024 & $13.683 \pm 0.184$ & $13.691 \pm 0.186$ & $27{,}448.0$ \\
          2{,}048 & $13.982 \pm 0.125$ & $13.992 \pm 0.127$ & $24{,}954.6$ \\
          4{,}096 & $14.615 \pm 0.175$ & $14.637 \pm 0.179$ & $20{,}571.2$ \\
          8{,}192 & $14.500 \pm 0.288$ & $14.521 \pm 0.291$ & $15{,}369.8$ \\
         16{,}384 & $14.600 \pm 0.059$ & $14.623 \pm 0.060$ & $10{,}216.6$ \\
        \bottomrule
    \end{tabular}
    \label{tab:compute-normalized-spce-saturation}
    \vspace{-2ex}
\end{table}

\section{Discussion and Conclusion}
\label{sect:conclusion}

We introduced \testbed, an open-source toolkit for Bayesian experimental design that provides a common modular foundation for implementing and comparing BED methods. By separating tasks, designers, objectives, and algorithms, \testbed makes individual components composable, so that new methods can be developed easily and isolated design choices studied in controlled experiments.%

\looseness=-1
Using \testbed, we conducted a benchmark of contemporary BED methods and ablations that expose several key considerations. First, EIG alone can give an incomplete picture of design quality: methods with similar information gain can differ substantially in downstream utility. Second, the choice of test problems matters, as performance varies widely across tasks, and simple baselines such as static designs remain highly competitive on several widely used problems. Finally, often-overlooked implementation details, such as policy architectures and time embeddings, can have effects that outweigh the differences between methods themselves. Architecture, estimator accuracy, and compute allocation should therefore be treated as part of the specification of a BED method rather than as incidental details. We hope that \testbed will serve as a flexible, easy-to-use foundation for both methodological and empirical work, accelerating the development and adoption of BED methods.

\bibliography{references}
\bibliographystyle{iclr2027_conference}

\clearpage
\newpage

\appendix

\section{Tasks}
\label{appendix_sect:tasks}

This section provides additional information about the \cls{Task} abstraction in \testbed. \cref{appendix_subsect:design_spaces} describes the design priors we use, which also control the space of admissible designs via the support of the prior. \cref{appendix_subsect:implemented_tasks} provides full details regarding the tasks we implement. Each task declares a tensor shape for a single design, observation and parameter. We write $d_\XX$, $d_\YY$ and $d_\Theta$ for the product of the entries of these shapes, i.e., the dimension once flattened to a vector, so that $\XX \subseteq \R^{d_\XX}$, $\YY \subseteq \R^{d_\YY}$ and $\Theta \subseteq \R^{d_\Theta}$. For example, location finding with $K = 2$ sources in two dimensions has parameter shape $[2, 2]$ and hence $d_\Theta = 4$.

\subsection{Design Priors}
\label{appendix_subsect:design_spaces}

In practice, it is often useful to specify a prior $p(x)$ over designs, and each \cls{Task} in \testbed specifies such a design prior. The design prior itself is a \cls{Distribution} object, which is a small wrapper around the standard PyTorch distribution class that enables interfacing with the rest of \testbed. The design prior provides two main functionalities. The first is to sample random designs, which can be used to create a \cls{RandomDesigner} baseline. The second is that the support of the design prior encodes the space of valid designs for the problem.

We directly extend the existing PyTorch distribution implementation, which includes utilities like transforming between raw/latent designs and constrained designs along with the corresponding change-of-variables Jacobian. While many methods do not require samples $x \sim p(x)$, the task itself may still crucially rely on its support which defines the valid designs. For this reason, \testbed provides a \cls{Box} distribution, which is uniform over the set $[a_1, b_1] \times \dots \times [a_{d_\XX}, b_{d_\XX}]$ where $a_k, b_k \in \R$ are user-specified bounds, which captures many constraints used in practice. For methods that require a finite set of designs, like ALINE \citep{huang2024amortized}, we provide an \cls{EmpiricalDistribution} class. For more complex constraint spaces, users can create custom distributions having the appropriate support, and can choose to not implement sampling if it is difficult or otherwise not required. By default, the design prior $p(x)$ is a standard Gaussian in $\R^{d_\XX}$, corresponding to unconstrained designs.

A policy may condition on either physical designs or the unconstrained values it emitted. When the reparametrisation is bijective these contain the same information, but the unconstrained representation often has a more convenient scale and avoids task-specific input normalisation.

\subsection{Implemented Tasks}
\label{appendix_subsect:implemented_tasks}

Here, we describe the various tasks currently implemented within \testbed.

\subsubsection{Location Finding}
\label{appendix_subsubsect:location_finding}

In this task, the experimenter must locate $K$ hidden sources in $\R^{d}$ from noisy measurements of the signal they emit, following the acoustic energy attenuation model of \citet{sheng2005maximum} as adapted for BED by \citet{foster2021deep}. Each source emits a signal whose intensity decays with distance, and a measurement returns the superposition of all $K$ contributions. The parameter of interest is the source locations $\theta = \{\theta_k\}_{k=1}^K$, where $K$ is assumed known ahead of time, and a design is the location $x \in \R^d$ at which the next measurement is taken.

Each source location is drawn independently from a standard Gaussian, so that
\begin{equation} \label{eqn:lf_prior}
    \theta_k \overset{\mathrm{i.i.d.}}{\sim} \mathcal{N}(0_d, I_d), \qquad k = 1, \dots, K,
\end{equation}
giving $\theta \in \R^{K \times d}$. The total intensity at a measurement location $x$ is
\begin{equation} \label{eqn:lf_intensity}
    \mu(\theta, x) = b + \sum_{k=1}^{K} \frac{\alpha}{m + \lVert \theta_k - x \rVert^{s}},
\end{equation}
and the outcome is the log total intensity observed with Gaussian noise,
\begin{equation} \label{eqn:lf_likelihood}
    y \mid \theta, x \sim \mathcal{N}\!\left(\log \mu(\theta, x), \, \sigma^2 \right).
\end{equation}
Here $\alpha$ is the amplitude of each source and $s$ the rate at which its signal decays, with $s = 2$ recovering the inverse-square law. The two remaining constants control the dynamic range of the observation. Setting $m > 0$ caps the contribution of a single source at $\alpha / m$, so that the intensity does not diverge as a measurement is taken arbitrarily close to a source and the EIG remains finite, while $b$ sets a background level that bounds how uninformative a measurement far from every source can be.

We adopt the configuration of \citet{foster2021deep}, namely $b = 0.1$, $m = 10^{-4}$, $\alpha = 1$ for every source, $\sigma = 0.5$ and $s = 2$. Our benchmark instantiates $K = 2$ sources in $d = 2$ dimensions over a horizon of $T = 30$ experiments, so $\theta$ is four-dimensional. The design space is unbounded: a measurement may be taken anywhere in $\R^{d}$, and the task therefore uses the default unconstrained space of \cref{appendix_subsect:design_spaces}.

 The sources are exchangeable, both in the prior and in the likelihood, since \cref{eqn:lf_intensity} depends on $\theta$ only through an unordered sum. Any labelling of the sources is therefore equally valid, which is why we report a permutation-invariant RMSE (\cref{appendix_sect:downstream_losses}) and canonicalise the source ordering in the variational posterior (\cref{appendix_subsect:jointvarpost}).

\subsubsection{Hyperbolic Temporal Discounting}
\label{appendix_subsubsect:temporal_discounting}

Temporal discounting describes the psychological phenomenon whereby the value a person ascribes to a reward falls as the delay before receiving it grows \citep{critchfield2001temporal, green2004discounting}. Following the notation of \citet{pmlr-v267-hedman25a}, an individual's time-delay preferences can be elicited through binary choices between an immediate reward and a larger delayed reward. In our setting, each experiment asks whether the participant would prefer $\$R_0$ today or a fixed \$100 in $D_1$ days. The experimental design is therefore $x=(D_1,R_0)$, and the outcome $y$ records the participant's choice between the immediate and delayed rewards.

\citet{mazur1987adjusting} introduced a widely used parametric model of temporal discounting. Following previous policy-based implementations \citep{foster2021deep, pmlr-v267-hedman25a}, we adopt the particular formulation given by \citet{vincent2016hierarchical}, which combines a discount curve with a probabilistic response rule and thereby defines a likelihood over observed choices. The individual participant is characterised by $\theta=(k,\alpha)$, where $k$ is their discount rate and $\alpha$ controls the scale of their choice noise, with prior
\begin{equation} \label{eqn:htd_prior}
    \log k \sim \mathcal{N}(-4.25, 1.5^2), \qquad \alpha \sim \mathrm{HalfNormal}(2).
\end{equation}
The two options carry subjective values
\begin{equation} \label{eqn:htd_value}
    V_0(x) = R_0, \qquad V_1(\theta, x) = \frac{100}{\left(1 + (k D_1)^{s}\right)^{b}},
\end{equation}
and the participant selects the delayed option with probability
\begin{equation} \label{eqn:htd_likelihood}
    p(y = 1 \mid \theta, x) = \epsilon + (1 - 2\epsilon)\,
        \Phi\!\left(\frac{V_1(\theta, x) - V_0(x)}{\alpha}\right),
\end{equation}
where $\Phi$ is the standard normal distribution function. Rather than choosing deterministically whichever option has the higher value, the participant compares them with sensitivity governed by $\alpha$, and the lapse rate $\epsilon$ bounds the response probability away from $0$ and $1$ to allow for a small chance of responding at random. We fix $\epsilon = 0.01$ and use a horizon of $T = 20$ questions.

The exponents $s$ and $b$ are both $1$ by default, recovering the standard hyperbolic model. 
A design must satisfy $D_1 \ge 0$ and $0 \le R_0 \le 100$. The design prior is constructed via the pushforward of standard Gaussian $(\tilde{x}_d, \tilde{x}_r) \sim \NN(0, 1)$ along the maps $D_1 = \exp(\tilde{x}_d - \log \bar{k})$ and $R_0 = 100\,\sigma(\tilde{x}_r)$, where the shift by the log prior mean of the discount rate keeps $k D_1$ at an informative scale for an untrained policy.

Unlike location finding, the outcome is a binary response rather than a continuous measurement, so the likelihood admits no pathwise gradient and methods that would otherwise differentiate through the rollout fall back to a score-function estimator (\cref{appendix_sect:reinforce}). \testbed also exposes a relaxed variant in which the response is drawn from a continuous relaxation of the Bernoulli, restoring a pathwise gradient at the cost of altering the model being scored; we retain the discrete model throughout so that reported EIG values remain comparable to existing literature.

\subsubsection{Constant Elasticity of Substitution}
\label{appendix_subsubsect:ces}

In economics, the constant elasticity of substitution (CES) model describes how a consumer trades one good off against another, with a single parameter controlling how readily goods substitute for one another \citep{Arrow1961CapitallaborSA}. Here the experimenter seeks to identify the preferences of an individual participant, who is shown two baskets of goods and reports the difference in utility they perceive between the two. The model was introduced to the BED setting by \citet{foster2019variational} and has since been used as a general BED benchmark \citep{blau2022optimizing, pmlr-v235-iollo24a, pmlr-v267-hedman25a, kerrigan2026a}.

A design corresponds to two baskets $x, x' \in [0, 100]^3$, each specifying non-negative quantities of three goods, so $d_\XX = 6$. The preferences of the participant are governed by $\theta = (\rho, \alpha, u)$, where $\rho$ is the substitution parameter, $\alpha = (\alpha_1, \alpha_2, \alpha_3)$ weights the three goods and $u$ scales the sensitivity of their response. The prior is
\begin{equation} \label{eqn:ces_prior}
    \rho \sim \mathrm{Beta}(1, 1), \qquad
    \alpha \sim \mathrm{Dirichlet}(1, 1, 1), \qquad
    \log u \sim \mathcal{N}(1, 3^2),
\end{equation}
so that $\alpha$ lies on the two-dimensional simplex and $\theta$ occupies a four-dimensional manifold within $\R^5$. A basket is valued through the CES utility
\begin{equation} \label{eqn:ces_utility}
    U(x) = \left( \sum_{i=1}^{3} \alpha_i x_i^{\rho} \right)^{1/\rho}.
\end{equation}
The participant compares the two baskets on a latent scale, which is Gaussian with a mean given by the difference in utility and a spread that grows with how far apart the baskets are,
\begin{equation} \label{eqn:ces_latent}
    \eta \mid \theta, x, x' \sim \mathcal{N}\!\left(
        u \left( U(x) - U(x') \right), \;
        \left( u \tau \left(1 + \lVert x - x' \rVert \right) \right)^2
    \right),
\end{equation}
and reports it on a sliding scale from $0$ to $1$ following a sigmoid transformation,
\begin{equation} \label{eqn:ces_likelihood}
    y = \mathrm{clip}\!\left( \sigma(\eta), \, \epsilon, \, 1 - \epsilon \right).
\end{equation}
We use $\tau = 0.005$ and $\epsilon = 2^{-22}$ over a horizon of $T = 10$ experiments, and the design prior is uniform over $[0, 100]^6$.

The censoring in \cref{eqn:ces_likelihood} is a unqiue characteristic of this task. Whenever one basket is clearly preferred, $\sigma(\eta)$ is driven to within $\epsilon$ of $0$ or $1$ and the response is clipped, so the observation distribution places atoms at both endpoints and a large part of the design space yields a saturated, uninformative answer. A policy must therefore learn to propose genuinely finely balanced pairs of baskets. This also has a consequence for optimisation: although $\eta$ is Gaussian and the outcome is formally reparametrisable, the log-likelihood switches branch at the censoring boundaries, and a pathwise gradient discards the term describing how the probability of censoring responds to the design. That term carries the signal telling a policy to leave the saturated region, so on this task we instead estimate the gradient with REINFORCE for DAD and JointVarPost (\cref{appendix_sect:reinforce}).

The support of $\theta$ also has consequences for methods that fit a posterior approximation. Because $\alpha$ is confined to the simplex and $\rho$ to an interval, $\theta$ does not fill $\R^5$, and a variational family placed directly on those coordinates would be a density with respect to the wrong measure. We instead push a distribution over unconstrained coordinates onto the support of the prior, as described in \cref{appendix_subsect:jointvarpost}.

\subsubsection{Pharmacokinetic}
\label{appendix_subsubsect:pharmacokinetic}

This task models a pharmacokinetic study, in which a drug is administered and blood subsequently drawn at chosen times to characterise how that drug is absorbed, distributed and eliminated. We use the one-compartment model with first-order absorption of \citet{RYAN201445}, which has become a standard BED benchmark \citep{Kleinegesse2020BayesianED, pmlr-v130-zhang21l}, and follow the configuration of \citet{ivanova2021implicit}.

The parameter of interest is $\theta = (k_a, k_e, V)$, comprising the absorption rate, the elimination rate and the volume of distribution, with a log-normal prior
\begin{equation} \label{eqn:pk_prior}
    \log \theta \sim \mathcal{N}\!\left( \log \left[1, \, 0.1, \, 20 \right], \; 0.05\, I_3 \right).
\end{equation}
A design is the time $x$ at which blood is sampled. Following a single dose $D_v$ given at the start of the experiment, the drug concentration at time $x$ is
\begin{equation} \label{eqn:pk_concentration}
    \mu(\theta, x) = \frac{D_v}{V} \cdot \frac{k_a}{k_a - k_e}
        \left( e^{-k_e x} - e^{-k_a x} \right),
\end{equation}
which rises while absorption dominates and then decays as elimination takes over. Measurements carry both multiplicative and additive error, so that
\begin{equation} \label{eqn:pk_likelihood}
    y \mid \theta, x \sim \mathcal{N}\!\left( \mu(\theta, x), \;
        \left( \sigma_{\mathrm{mult}}\, \mu(\theta, x) \right)^2 + \sigma_{\mathrm{add}}^2 \right).
\end{equation}
The multiplicative term grows in proportion to the quantity being measured, so the noise is heteroscedastic: readings taken near the peak of the concentration curve are inherently less precise than those taken in the tails.

We set $D_v = 400$, $\sigma_{\mathrm{mult}}^2 = 0.01$ and $\sigma_{\mathrm{add}}^2 = 0.1$. Blood may be sampled at any point in the first day, corresponding to a uniform design prior over $[0, 24]$ hours (\cref{appendix_subsect:design_spaces}), and our benchmark uses a horizon of $T = 10$ measurements. As in the references above, each measurement is taken from a different patient, all of whom are assumed to share the same $\theta$, so a single blood sample is drawn per patient.

The outcome is a differentiable function of $\theta$ and $x$ perturbed by Gaussian noise, so the likelihood is reparametrisable and methods may take pathwise gradients of the EIG. We note that \cref{eqn:pk_concentration} is singular when $k_a = k_e$, and the model is conventionally stated with the constraint $k_a > k_e$. The prior does not impose this constraint, but it separates the two rates by an order of magnitude relative to a log-scale standard deviation of $\sqrt{0.05}$, so draws that approach the singularity are rare.

\subsubsection{Michaelis-Menten}
\label{appendix_subsubsect:michaelis_menten}

This task models an enzyme kinetics experiment, in which the experimenter measures the velocity of a chemical reaction at a chosen substrate concentration in order to characterise the enzyme. The classical description of this relationship is the Michaelis-Menten model \citep{Michaelis2011Original}, under which the reaction velocity rises with substrate concentration before levelling off as the enzyme saturates. It has a substantial history as an optimal design problem \citep{Dette01092003, Overstall2022Bayesian}, and we follow the BED setup of \citet{forster2025improving}.

The parameter of interest is $\theta = (V_{\max}, V_{\mathrm{half}})$, where $V_{\max}$ is the velocity the reaction approaches once the enzyme is saturated and $V_{\mathrm{half}}$ is the substrate concentration at which half that velocity is attained. Both are given independent uniform priors,
\begin{equation} \label{eqn:mm_prior}
    V_{\max}, \, V_{\mathrm{half}} \overset{\mathrm{i.i.d.}}{\sim} \mathrm{Unif}(20, 200).
\end{equation}
A design is the substrate concentration $x$ at which the next measurement is made, giving a mean reaction velocity
\begin{equation} \label{eqn:mm_rate}
    \mu(\theta, x) = \frac{V_{\max}\, x^{s}}{V_{\mathrm{half}}^{s} + x^{s}},
\end{equation}
observed with additive Gaussian noise,
\begin{equation} \label{eqn:mm_likelihood}
    y \mid \theta, x \sim \mathcal{N}\!\left( \mu(\theta, x), \, \sigma^2 \right).
\end{equation}
We use $\sigma = 2.5$ and a horizon of $T = 20$ measurements. Concentrations are non-negative and the design prior is uniform over $[0, 400]$ (\cref{appendix_subsect:design_spaces}), whose upper bound is twice the largest $V_{\mathrm{half}}$ the prior admits and so comfortably covers the saturating regime.

The exponent $s$ defaults to $1$. Other values give the Hill model \citep{cornish2013fundamentals}, in which binding at one site alters the affinity of the remaining sites. \citet{forster2025improving} used this as a controlled setting for studying misspecification, since $s$ can be varied in the data-generating process while the assumed model is held at $s = 1$, tuning the severity of the mismatch continuously from none upwards.

The saturating response is the central tension of the task. Measurements at high concentration pin down $V_{\max}$ but say little about $V_{\mathrm{half}}$, whereas the concentrations that identify $V_{\mathrm{half}}$ lie on the rising part of the curve, whose position is itself unknown a priori. A policy must therefore locate the bend in the curve before it can resolve both parameters. The noise is homoscedastic, so unlike the pharmacokinetic task there is no additional penalty for measuring where the response is large, and the outcome is a differentiable function of $\theta$ and $x$ perturbed by Gaussian noise, so the likelihood is reparametrisable and pathwise gradients of the EIG are available.

\subsubsection{Stochastic Pendulum}
\label{appendix_subsubsect:stochastic_pendulum}

This task asks the experimenter to identify the physical properties of a pendulum by applying torques and observing how it moves. It is one of three dynamical systems in our \testbed suite, and a task whose likelihood is not memoryless: the system carries state between steps, so the outcome depends on the previous state as well as the current design, meaning the history is not exchangeable. We take the nonlinear formulation of \citet{iqbal2024nesting}, with the diffusion matched to \citet{phillips2026bayesian}.

The pendulum is modelled as a uniform rod of unknown mass $m$ and length $l$, so the parameter of interest is $\theta = (m, l)$ with a log-normal prior
\begin{equation} \label{eqn:pendulum_prior}
    \log \theta \sim \mathcal{N}\!\left(0, \, 0.01\, I_2 \right).
\end{equation}
The state $y = (q, \dot{q})$ collects the angle and the angular velocity, and the experiment begins from rest at $y_0 = (0, 0)$. A design is the torque $x$ applied over the next interval. Writing $g$ for gravitational acceleration and $d$ for a damping coefficient, the angular acceleration is
\begin{equation} \label{eqn:pendulum_dynamics}
    \ddot{q} = 3 \left( - \frac{g}{2 l} \sin q + \frac{x - d \dot{q}}{m l^{2}} \right),
\end{equation}
in which the factor of three and the factor of one-half in the gravitational term arise from modelling the pendulum as a uniform rod pivoted at one end, whose centre of mass lies at $l/2$ and whose moment of inertia is $ml^2/3$; the term $x-d\dot q$ represents the applied and damping torques, respectively.
The dynamics are integrated by an Euler-Maruyama step, so that
\begin{equation} \label{eqn:pendulum_likelihood}
    y_t \mid \theta, x_t, y_{t-1} \sim \mathcal{N}\!\left(
        y_{t-1} + \Delta t \cdot f(y_{t-1}, \theta, x_t), \;
        \Delta t \, \Sigma^2 \right),
\end{equation}
where $f(y, \theta, x) = (\dot{q}, \ddot{q})$ is the time derivative of the state, with $\ddot{q}$ given by \cref{eqn:pendulum_dynamics} and $\Sigma$ is the diffusion scale.

We use $\Delta t = 0.05$, $d = 0.1$, $g = 9.81$ and $\Sigma = 0.1 I_2$, over a horizon of $T = 50$ steps.
 The torque is bounded to $[-1, 1]$, and we use a uniform design prior over this set (\cref{appendix_subsect:design_spaces}).

The loss of exchangeability changes what a policy must do. A design is informative in the context of the state it acts upon, so the policy has to reason about the effects of its earlier torques.
At the initial equilibrium the model is degenerate: $\sin q = 0$ removes the gravitational term from \cref{eqn:pendulum_dynamics}, leaving a response that depends on $\theta$ only through the product $m l^{2}$, so no torque applied at rest can separate mass from length. Breaking that degeneracy requires displacing the pendulum, since the gravitational term is the only one that depends on $l$ alone, and with the torque bounded to $[-1, 1]$ against gravity this takes several steps of deliberate excitation.
 This makes the task a control and an estimation problem. The Euler-Maruyama step is a differentiable drift plus additive Gaussian noise, so the outcome is reparametrisable and pathwise gradients of the EIG are available.

\subsubsection{Cart Pole}
\label{appendix_subsubsect:cart_pole}
This task places a pole on a cart that may be pushed along a track, and asks the experimenter to identify the masses of the pole and cart alongside the pole length by choosing a force to be applied at each step. As with the stochastic pendulum it is a non-memoryless dynamical system taken from \citet{iqbal2024nesting}, with the diffusion matched to \citet{phillips2026bayesian}, but the state is four-dimensional and, as we show below, the pole mass is not identifiable until the pole has been displaced from vertical.

The parameter of interest is $\theta = (m_p, m_c, l)$, the pole mass, the cart mass and the pole length, with prior
\begin{equation} \label{eqn:cartpole_prior}
    \log \theta \sim \mathcal{N}\!\left(0, \, 0.01\, I_3 \right).
\end{equation}
The state $y = (s, q, \dot{s}, \dot{q})$ collects the cart position, the pole angle and their velocities, and the system starts at rest with $y_0 = 0$. A design is the force $x$ applied to the cart. Writing $\Delta := m_c + m_p \sin^2 q$, the accelerations are
\begin{align}
    \ddot{s} &= \frac{1}{\Delta}\left(
        x + m_p \sin q \left( l \dot{q}^2 + g \cos q \right)
        - \left( k_1 s + d_1 \dot{s} \right)
        - \frac{\left( k_2 q + d_2 \dot{q} \right)\cos q}{l}
    \right),
    \label{eqn:cartpole_cart}
    \\[2pt]
    \ddot{q} &=
    \begin{aligned}[t]
        -\frac{1}{l \Delta}\Bigg(
        x \cos q
        &+ m_p l \dot{q}^2 \cos q \sin q
        + \left( m_c + m_p \right) g \sin q \\
        &+ \left( k_1 s + d_1 \dot{s} \right)\cos q
        + \frac{\left( k_2 q + d_2 \dot{q} \right)\cos^2 q}{l}
        \Bigg)
    \end{aligned}
    \label{eqn:cartpole_pole}
\end{align}
where $k_1, k_2$ are stiffness coefficients and $d_1, d_2$ damping coefficients acting on the cart and the pole respectively. The dynamics are integrated by an Euler-Maruyama step,
\begin{equation} \label{eqn:cartpole_likelihood}
    y_t \mid \theta, x_t, y_{t-1} \sim \mathcal{N}\!\left(
        y_{t-1} + \Delta t \cdot f(y_{t-1}, \theta, x_t), \;
        \Delta t \, \Sigma^2 \right),
\end{equation}
with $f = (\dot{s}, \dot{q}, \ddot{s}, \ddot{q})$ the time derivative of the state.

We use $\Delta t = 0.05$, $g = 9.81$, $k_1 = k_2 = d_1 = d_2 = 0.01$ and $\Sigma = 0.1\, I_4$, over a horizon of $T = 50$ steps. The force is bounded to $[-5, 5]$, and we use a uniform design prior over this set (\cref{appendix_subsect:design_spaces}). The stiffness and damping terms are small and act as weak restoring forces towards $s = 0$ and $q = 0$.

The identifiability structure is what makes this task harder than the pendulum. Every occurrence of $m_p$ in \cref{eqn:cartpole_cart} and \cref{eqn:cartpole_pole}, including the one inside $\Delta$, is multiplied by $\sin q$ or $\sin^2 q$, so at the initial state the equations collapse to $\ddot{s} = x / m_c$ and $\ddot{q} = -x / (l m_c)$. A force applied at rest therefore reveals the cart mass and the pole length, but carries no information about the pole mass: $m_p$ becomes visible only once the pole is away from vertical. A policy must consequently excite the pole, and it can do so only indirectly, by accelerating the cart beneath it. The outcome remains a differentiable drift perturbed by additive Gaussian noise, so pathwise gradients of the EIG are available. At $T = 50$ with a four-dimensional state this is also one of the more expensive tasks in the suite, and is where the contrastive estimators used by DAD required a reduced batch size (\cref{appendix_subsect:dad}).

\subsubsection{Double Pendulum}
\label{appendix_subsubsect:double_pendulum}

This task consists of two links joined end to end, each of unknown mass and length, driven by a torque at each joint. It is the third and largest of the dynamical system tasks, and it carries the most parameters, the widest design space and the richest dynamics of the three \citep{iqbal2024nesting, phillips2026bayesian}.

The parameter of interest is $\theta = (m_1, m_2, l_1, l_2)$, the masses and lengths of the two links, with prior
\begin{equation} \label{eqn:doublelink_prior}
    \log \theta \sim \mathcal{N}\!\left(0, \, 0.01\, I_4 \right).
\end{equation}
The state $y = (q_1, q_2, \dot{q}_1, \dot{q}_2)$ collects the two joint angles and their velocities, starting from rest at $y_0 = 0$, and a design $x \in \R^2$ applies a torque at each joint. The dynamics take the manipulator form
\begin{equation} \label{eqn:doublelink_dynamics}
    M(\theta, q)\, \ddot{q} + C(\theta, q, \dot{q})\, \dot{q} = x + \tau(\theta, q),
\end{equation}
in which the mass matrix, the Coriolis and centrifugal matrix, and the gravitational torque are
\begin{gather}
    M = \begin{pmatrix}
        (m_1 + m_2) l_1^2 + m_2 l_2^2 + 2 m_2 l_1 l_2 \cos q_2 &
        m_2 l_2^2 + m_2 l_1 l_2 \cos q_2 \\
        m_2 l_2^2 + m_2 l_1 l_2 \cos q_2 & m_2 l_2^2
    \end{pmatrix}, \label{eqn:doublelink_mass} \\[3pt]
    C = \begin{pmatrix}
        0 & a \left( 2\dot{q}_1 + \dot{q}_2 \right) \\
        -\tfrac{1}{2} a \left( 2\dot{q}_1 + \dot{q}_2 \right) & \tfrac{1}{2} a \dot{q}_1
    \end{pmatrix}, \qquad a := -m_2 l_1 l_2 \sin q_2, \label{eqn:doublelink_coriolis} \\[3pt]
    \tau = -g \begin{pmatrix}
        (m_1 + m_2) l_1 \sin q_1 + m_2 l_2 \sin(q_1 + q_2) \\
        m_2 l_2 \sin(q_1 + q_2)
    \end{pmatrix}. \label{eqn:doublelink_gravity}
\end{gather}
Solving \cref{eqn:doublelink_dynamics} for $\ddot{q}$ gives the drift, which is integrated by the same Euler-Maruyama step as the other dynamical tasks,
\begin{equation} \label{eqn:doublelink_likelihood}
    y_t \mid \theta, x_t, y_{t-1} \sim \mathcal{N}\!\left(
        y_{t-1} + \Delta t \cdot f(y_{t-1}, \theta, x_t), \;
        \Delta t \, \Sigma^2 \right).
\end{equation}

We use $\Delta t = 0.05$, $g = 9.81$ and $\Sigma = 0.1\, I_4$, over a horizon of $T = 50$ steps. The two torques are bounded differently, to $[-4, 4]$ at the first joint and $[-2, 2]$ at the second, giving a box design space whose coordinates are on distinct scales; this is the task for which conditioning a policy on unconstrained rather than physical designs is most clearly useful (\cref{appendix_subsect:design_spaces}). We use a uniform design prior over this space.

As with the other two dynamical systems, the initial state is degenerate, and here the degeneracy is exact. At $y_0 = 0$ both the Coriolis matrix and the gravitational torque vanish, leaving $\ddot{q} = M(\theta, 0)^{-1} x$. Since $M$ is symmetric it contributes only three independent entries, which cannot determine four parameters; concretely, the map $(m_1, m_2, l_1, l_2) \mapsto (m_1 / c^2, m_2 / c^2, c\, l_1, c\, l_2)$ leaves $M(\theta, 0)$ unchanged for any $c > 0$, so an entire scaling family of parameters is indistinguishable from rest. It is the gravitational term of \cref{eqn:doublelink_gravity} that breaks this family, and it is visible only once the links swing away from the vertical. The outcome is again a differentiable drift with additive Gaussian noise, so pathwise gradients are available.

\section{Policy Gradients and REINFORCE}
\label{appendix_sect:reinforce}

This section provides additional details regarding our REINFORCE-style objectives, which enable us to train BED policies even in the presence of non-reparametrisable likelihoods. Much of this section is not new and detailed only for the sake of completeness. However, \cref{appendix_subsect:jointvarpost_reinforce} derives the necessary details for training JointVarPost \citep{pmlr-v108-foster20a-unified, phillips2026bayesian} on non-reparametrisable likelihoods, and \cref{appendix_subsect:actionbed_reinforce} does the same for ActionBED \citep{rossa2026actionbed} on non-reparametrisable likelihoods, both of which are novel methodological approach not previously studied in the literature.

\subsection{sPCE and REINFORCE}

One of the key design objectives against which policies are optimised is the sPCE lower bound on EIG \citep{foster2021deep}. Here we derive its REINFORCE-style gradient estimator for deterministic and stochastic policies, as used when the likelihood cannot be reparametrised.

Let $\phi$ denote the policy parameters and $\pi_\XX^\phi$ the design policy. We draw
$\theta_0 \sim p(\theta)$ and contrastive samples
$\theta_{1:L} \overset{\mathrm{i.i.d.}}{\sim} p(\theta)$, and generate a history
$h_T = (x_{1:T},y_{1:T})$. The sPCE bound of \cref{eqn:pce} can then be rewritten as
\begin{align}
\label{eqn:spce_objective}
    \LL_T(\phi)
    &= \E_{p(\theta_{0:L})p(h_T \mid \theta_0,\pi_\XX^\phi)}
    \left[R_\phi(h_T)\right] + \log(L+1),\\
    R_\phi(h_T)
    &:= \log
    \frac{p(h_T\mid\theta_0,\pi_\XX^\phi)}
    {\sum_{\ell=0}^L p(h_T\mid\theta_\ell,\pi_\XX^\phi)}.
\end{align}

The gradient estimator then takes slightly different forms based on whether the design policy is stochastic or deterministic. 

\subsubsection{Deterministic design policies}

This is a reiteration of the derivation within \citet{foster2021deep}.
For a deterministic policy,
$
    x_t = \pi_\XX^\phi(h_{t-1}),
$
so the only randomness in the rollout is in the observations. Its density is therefore
\begin{equation}
\label{eqn:deterministic_rollout}
    p(h_T\mid\theta,\pi_\XX^\phi)
    =
    \prod_{t=1}^T
    p\!\left(y_t\mid\theta,x_t,h_{t-1}\right),
    \qquad
    x_t=\pi_\XX^\phi(h_{t-1}),
\end{equation}
where dependence on $\phi$ enters through the designs $x_t$. Hence while the likelihood may not be reparametrisable in $y_t$, its log-density will be differentiable with respect to the designs.

Writing $p_0(h_T)=p(h_T\mid\theta_0,\pi_\XX^\phi)$ and differentiating \cref{eqn:spce_objective} directly,
\begin{align}
\frac{\d \LL_T}{\d\phi}
&=
\E_{p(\theta_{0:L})}
\left[
\int
\frac{\d}{\d\phi}
\left\{
p_0(h_T)R_\phi(h_T)
\right\}
\d h_T
\right] \\
&=
\E_{p(\theta_{0:L})p_0(h_T)}\left[
R_\phi(h_T)\frac{\d}{\d\phi}\log p_0(h_T)
+
\frac{\d}{\d\phi}R_\phi(h_T)
\right],
\label{eqn:general_reinforce}
\end{align}
where we used the score-function identity
$\frac{\d}{\d\phi}p_0=p_0\frac{\d}{\d\phi}\log p_0$. Expanding the second term gives
\begin{align}
\frac{\d \LL_T}{\d\phi}
=
\E\left[
R_\phi(h_T)\frac{\d}{\d\phi}\log p_0(h_T)
+
\frac{\d}{\d\phi}\log p_0(h_T)
-
\frac{\d}{\d\phi}
\log\sum_{\ell=0}^L
p(h_T\mid\theta_\ell,\pi_\XX^\phi)
\right].
\end{align}
The middle term vanishes because
\[
\E_{p_0}\!\left[\frac{\d}{\d\phi}\log p_0(h_T)\right]
=
\int \frac{\d}{\d\phi}p_0(h_T)\,\d h_T
=
\frac{\d}{\d\phi}1
=
0.
\]
Thus
\begin{equation}
\label{eqn:reinforce_deterministic}
    \frac{\d \LL_T}{\d\phi}
    =
    \E_{p(\theta_{0:L})p(h_T\mid\theta_0,\pi_\XX^\phi)}\left[
        R_\phi(h_T)
        \frac{\d}{\d\phi}
        \log p(h_T\mid\theta_0,\pi_\XX^\phi)
        -
        \frac{\d}{\d\phi}
        \log\sum_{\ell=0}^L
        p(h_T\mid\theta_\ell,\pi_\XX^\phi)
    \right]
\end{equation}
The first term is the score-function contribution from the sampled outcomes, while the second differentiates the contrastive objective directly through the deterministic designs. This is the estimator of \citet{foster2021deep}. We note that the resulting gradient resembles the deterministic policy gradient used in DDPG \citep{ddpg}. The important distinction is that, rather than differentiating through a learned critic \(Q(s,a)\), we differentiate the sPCE objective directly. Since the distribution over observations entering this objective depends on the design, its derivative contains a score-function term.

\subsubsection{Stochastic policies}

For a stochastic, non-reparametrised policy,
$x_t\sim\pi_\XX^\phi(\cdot\mid h_{t-1})$, and the rollout density is
\begin{equation}
\label{eqn:stochastic_rollout}
    p(h_T\mid\theta,\pi_\XX^\phi)
    =
    \prod_{t=1}^T
    \pi_\XX^\phi(x_t\mid h_{t-1})
    p(y_t\mid\theta,x_t,h_{t-1}).
\end{equation}

Substituting this into the sPCE likelihood ratio, the policy factors cancel because they are identical for every $\theta_\ell$:
\[
R(h_T)
=
\log
\frac{
\prod_{t=1}^T p(y_t\mid\theta_0,x_t,h_{t-1})
}{
\sum_{\ell=0}^L
\prod_{t=1}^T p(y_t\mid\theta_\ell,x_t,h_{t-1})
}.
\]

This cancellation removes the explicit policy density from the reward, but it is not what prevents direct differentiation through $R$. The key distinction from the deterministic case is that the designs are now sampled from a non-reparametrised distribution. Once a rollout has been realised, the sampled $x_{1:T}$ are treated as fixed when differentiating with respect to $\phi$, so there is no pathwise derivative from $\phi$ through the realised designs to the likelihood terms in $R(h_T)$. Therefore
\[
\frac{\d}{\d\phi}R(h_T)=0
\]
for the realised trajectory, and the dependence of the objective on $\phi$ is instead captured through how $\phi$ changes the distribution over sampled designs.

Using \cref{eqn:general_reinforce}, the remaining derivative is the score of the stochastic rollout distribution. Holding the realised history fixed,
\[
\frac{\d}{\d\phi}
\log p(h_T\mid\theta_0,\pi_\XX^\phi)
=
\sum_{t=1}^T
\frac{\d}{\d\phi}
\log \pi_\XX^\phi(x_t\mid h_{t-1}),
\]
since the observation likelihood contains no explicit $\phi$-dependence when the sampled designs are held fixed. Hence
\begin{equation}
\label{eqn:reinforce_stochastic}
    \frac{\d \LL_T}{\d\phi}
    =
    \E_{p(\theta_{0:L})p(h_T\mid\theta_0,\pi_\XX^\phi)}
    \left[
        R(h_T)
        \sum_{t=1}^T
        \frac{\d}{\d\phi}
        \log\pi_\XX^\phi(x_t\mid h_{t-1})
    \right],
\end{equation}
which is the standard REINFORCE policy-gradient estimator used by general reinforcement-learning approaches.

The distinction between the two cases is therefore where the dependence on $\phi$ enters. For a deterministic policy, the realised designs remain differentiable functions of $\phi$, so both the rollout likelihood and the contrastive reward can be differentiated through the designs. For a stochastic, non-reparametrised policy, the realised designs provide no pathwise gradient, and the effect of $\phi$ is captured entirely through the score of the policy distribution.

\subsection{JointVarPost and REINFORCE}
\label{appendix_subsect:jointvarpost_reinforce}

JointVarPost (\cref{appendix_subsect:jointvarpost}) co-optimises a variational posterior and a design policy against the Barber-Agakov bound of \cref{eqn:ba}. As with sPCE, a pathwise gradient of that bound is unavailable when the outcomes cannot be reparametrised. We introduce the corresponding score-function estimator as a methodological contribution to the BED literature. We derive it here, which is what allows us to run JointVarPost on the tasks with discrete or censored outcomes.

Write $\psi$ for the variational parameters and $\phi$ for those of the design policy, so that \cref{eqn:ba} reads
\begin{align}
\label{eqn:ba_objective}
    \LL_{\mathrm{BA}}(\phi, \psi)
    &= \E_{p(\theta) p(h_T \mid \theta, \pi_\XX^\phi)}
        \left[ R_{\psi,\phi}(\theta, h_T) \right], \\
    R_{\psi,\phi}(\theta, h_T)
    &:= \log q_\psi(\theta \mid h_T) - \log p(\theta).
\end{align}
The prior term depends on neither set of parameters. When applying REINFORCE, the realised outcomes are treated as constants, having no derivative path back to the designs that produced them, as holds automatically whenever the outcomes are not reparametrisable. The designs themselves remain differentiable functions of $\phi$, which is what makes the second term of \cref{eqn:ba_grad_phi} non-zero.

\begin{prop}[REINFORCE for JointVarPost] \label{prop:reinforce_jointvarpost}
    Assume the design policy $\pi_\XX^\phi$ is deterministic and that $\log p(y \mid \theta, x)$ is differentiable with respect to $x$. Then, with both expectations taken over $p(\theta) p(h_T \mid \theta, \pi_\XX^\phi)$,
    \begin{align}
        \frac{\d \LL_{\mathrm{BA}}}{\d \psi} &= \E\left[
            \frac{\d}{\d \psi} \log q_\psi(\theta \mid h_T) \right],
            \label{eqn:ba_grad_psi} \\
        \frac{\d \LL_{\mathrm{BA}}}{\d \phi} &= \E\left[
            R_{\psi,\phi}(\theta, h_T)
            \frac{\d}{\d \phi} \log p(h_T \mid \theta, \pi_\XX^\phi)
            + \frac{\d}{\d \phi} R_{\psi,\phi}(\theta, h_T) \right].
            \label{eqn:ba_grad_phi}
    \end{align}
\end{prop}
\begin{proof}
    The variational parameters do not enter the sampling distribution, giving \cref{eqn:ba_grad_psi} immediately. For \cref{eqn:ba_grad_phi}, differentiating under the integral and applying the score-function identity,
    \begin{align}
        \frac{\d \LL_{\mathrm{BA}}}{\d \phi}
        &= \E_{p(\theta)} \left[ \int \frac{\d}{\d \phi} \left\{
            p(h_T \mid \theta, \pi_\XX^\phi) R_{\psi,\phi}(\theta, h_T)
            \right\} \d h_T \right] \\
        &= \E \left[
            R_{\psi,\phi}(\theta, h_T)
            \frac{\d}{\d \phi} \log p(h_T \mid \theta, \pi_\XX^\phi)
            + \frac{\d}{\d \phi} R_{\psi,\phi}(\theta, h_T) \right].
    \end{align}
\end{proof}

The two sets of parameters behave quite differently. The variational parameters never enter the sampling distribution, so \cref{eqn:ba_grad_psi} is an ordinary pathwise gradient, and maximising it with respect to $\psi$ amounts to amortised maximum likelihood: fitting $q_\psi$ to the posterior on the histories the current policy generates. Neither the form of the policy nor the reparametrisability of the outcomes affects the $\psi$ gradient, since both only affect how $\phi$ acts on the sampling distribution; the score-function estimator is therefore needed for $\phi$ alone.

The two gradients are nonetheless coupled. The reward $R_{\psi,\phi}$ is itself learned, depending on the current variational parameters, whereas the sPCE reward of \cref{eqn:spce_objective} is determined by the model alone. Early in training the policy is therefore rewarded for histories on which an under-fitted $q_\psi$ happens to score well. The bound of \cref{eqn:ba} remains valid throughout, since it holds for any $q_\psi$, but its tightness, and hence the quality of the signal reaching the policy, improves only as the posterior is fitted.

For a stochastic, non-reparametrised policy the realised designs provide no pathwise gradient through the reward, so the reward, now written $R_\psi$, loses its dependence on $\phi$, the second term of \cref{eqn:ba_grad_phi} vanishes, and the score becomes that of the policy,
\begin{equation} \label{eqn:ba_grad_phi_stochastic}
    \frac{\d \LL_{\mathrm{BA}}}{\d \phi} = \E\left[
        R_\psi(\theta, h_T)
        \sum_{t=1}^{T} \frac{\d}{\d \phi} \log \pi_\XX^\phi(x_t \mid h_{t-1}) \right].
\end{equation}
In both cases the score term admits a baseline, and we use a leave-one-out control variate.

\subsection{ActionBED and REINFORCE}
\label{appendix_subsect:actionbed_reinforce}

The original ActionBED work of \citet{rossa2026actionbed} mentions that non-reparametrisable likelihoods (e.g., when $y$ is discrete) can be handled using a REINFORCE estimator \citep{williams1992simple} of the policy gradients. However, \citet{rossa2026actionbed} do not include the necessary details for actually carrying this out (which are nontrivial), and do not perform any experiments in the non-reparametrisable setting. In this section, we derive a novel REINFORCE estimator that is compatible with ActionBED, enabling us to apply ActionBED to tasks with non-reparametrisable likelihoods.

Write $\phi$ for the collection of all policy parameters, including both the design and action policies, so that $\pi_\XX^\phi$ is our design policy and  $\pi_\AA^\phi$ is our action policy. For a downstream loss $\ell(a, \theta)$, we define the loss $\LL(\phi)$ on parameters via 
\begin{equation}
    \LL(\phi) = \E_{p(\theta)p(h_T \mid \theta, \pi_\XX^\phi)}\left[ \ell(\pi_\AA^\phi(h_T), \theta) \right].
\end{equation}
We do not assume the likelihood is reparametrisable, but we will assume that the log-likelihood evaluations are differentiable with respect to the designs. This is frequently the case in many BED settings of interest. We have the following REINFORCE style gradient expression.

\begin{prop}[REINFORCE for ActionBED]    \label{prop:reinforce_actionbed}
    Assume that the design policy $\pi_\XX$ is deterministic and the log-likelihood $p(y \mid \theta, x)$ is differentiable with respect to $x$. Then,
    \begin{equation}
        \frac{\d \LL}{\d \phi} = \E_{p(\theta) p(h_T \mid \theta, \pi_\XX^\phi)} \left[ \ell\left(\pi_\AA^\phi (h_T), \theta\right) \left(\frac{\d}{\d \phi} \log p(h_T \mid \theta, \pi_\XX^\phi) \right) + \frac{\d }{\d \phi} \ell\left(\pi_\AA^\phi(h_T), \theta\right) \right].
    \end{equation}
\end{prop}
\begin{proof}
    Using the log-derivative trick,
    \begin{align}
        \frac{\d \LL}{\d \phi} &= \E_{p(\theta)} \left[ \int \frac{\d}{\d \phi} \left( \ell\left( \pi_\AA^\phi(h_T), \theta \right) p(h_T \mid \theta, \pi_\XX^\phi) \right) \d h_T \right] \\
        &= \E_{p(\theta)} \left[ \int \left(\ell\left( \pi_\AA^\phi(h_T), \theta \right) \frac{\d}{\d \phi} p(h_T \mid \theta, \pi_\XX^\phi) + p(h_T \mid \theta, \pi_\XX^\phi) \frac{\d}{\d \phi} \ell\left( \pi_\AA^\phi(h_T), \theta \right) \right) \d h_T \right] \\
        &= \E_{p(\theta)} \left[ \int p(h_T \mid \theta, \pi_\XX^\phi) \left(\ell\left( \pi_\AA^\phi(h_T), \theta \right) \frac{\d}{\d \phi} \log p(h_T \mid \theta, \pi_\XX^\phi) + \frac{\d}{\d \phi} \ell\left( \pi_\AA^\phi(h_T), \theta \right) \right) \d h_T \right] \\
        &= \E_{p(\theta) p(h_T \mid \theta, \pi_\XX^\phi)} \left[ \ell\left(\pi_\AA^\phi (h_T), \theta\right) \left(\frac{\d}{\d \phi} \log p(h_T \mid \theta, \pi_\XX^\phi) \right) + \frac{\d }{\d \phi} \ell\left(\pi_\AA^\phi(h_T), \theta\right) \right].
    \end{align}
    
\end{proof}

Note that the second summand in this expression is a total derivative, and that $h_T$ itself depends on $\phi$ through the policy $\pi_\XX^\phi$.

In the special case that the two policies do not share any parameters, i.e., $\phi = (\phi_\XX, \phi_\AA)$ separates into two subsets such that our policies are $\pi_\XX^{\phi_\XX}$ and $\pi_\AA^{\phi_\AA}$, we see from \cref{prop:reinforce_actionbed} that our loss gradients take the simpler form
\begin{equation}
    \frac{\d \LL}{\d \phi_\AA} = \E_{p(\theta) p(h_T \mid \theta, \pi_\XX^{\phi_\XX})} \left[ \frac{\d}{\d \phi_\AA} \ell\left(\pi_\AA^{\phi_\AA}(h_T), \theta\right) \right]
\end{equation}
    
\begin{equation}
    \frac{\d \LL}{\d \phi_\XX} = \E_{p(\theta) p(h_T \mid \theta, \pi_\XX^{\phi_\XX})} \left[ \ell\left(\pi_\AA^{\phi_\AA}(h_T), \theta\right) \left(\frac{\d}{\d \phi_\XX} \log p(h_T \mid \theta, \pi_\XX^{\phi_\XX}) \right) + \frac{\d }{\d \phi_\XX} \ell\left(\pi_\AA^{\phi_\AA}(h_T), \theta\right) \right].
\end{equation}

That is, gradients for the action policy are the usual pathwise gradients as $\phi_\AA$ does not affect the sampling distribution. On the other hand, the design policy gradient includes terms coming from the REINFORCE estimator.

\section{Downstream Losses}
\label{appendix_sect:downstream_losses}

One way in which we evaluate design policies in this work is through fitting a downstream action policy on top of a frozen design policy. This section contains additional details regarding this evaluation. These losses are additionally used to train ActionBED \citep{rossa2026actionbed} policies.

We consider two closely related downstream decisions: our action $a$ is either a prediction of $\theta$ or of $\log \theta$. On some tasks, $\theta$ is naturally on a log-scale, making the latter a more appropriate target. Both can be handled under the same framework via the reparametrisation $\theta \mapsto \log \theta$. We predict $\theta$ on all tasks except for temporal discounting and CES, where we predict $\log \theta$. Note that in \cref{tab:benchmark}, MSE and CMSE on temporal discounting and CES refer to errors on $\log \theta$. We define $\vartheta$ as the prediction target.

In both cases, $a, \vartheta \in \R^{d_\Theta}$ have the same dimensionality, so that if  $a = (a_1, \dots, a_{d_\Theta})$ and $\vartheta = (\vartheta, \dots, \vartheta_{d_\Theta})$, we may define the coordinate-wise error $e_i = a_i - \vartheta_i$. We consider two losses. The first is the usual mean-squared-error (MSE) loss
\begin{equation}
    \ell_{\mathrm{MSE}}(a, \vartheta) = \frac{1}{d_\Theta} \lVert a - \vartheta \rVert^2  = \frac{1}{d_\Theta}\sum_{i=1}^{d_\Theta} e_i^2.
\end{equation}
We also consider the combined MSE (CMSE) loss \citep{rossa2026actionbed}, which simultaneously measures predictive accuracy on the scale of the prediction target as well as accuracy on a finer-grained log scale, via
\begin{equation}
    \ell_{\mathrm{CMSE}}(a, \vartheta) = \frac{1}{d_\Theta} \sum_{i=1}^{d_\Theta} \left(e_i^2 + \log(e_i^2 + \epsilon) \right)
\end{equation}
where $0 < \epsilon \ll 1 $ is a small positive constant added for numerical stability; we use $\epsilon = 10^{-6}$ throughout. For the sake of consistent reporting, we follow ActionBED \citep{rossa2026actionbed} and normalise by the dimensionality $d_\Theta$ of the space, which allows for comparison across tasks of different dimensions.  

\paragraph{Permutation Invariant Lossess.} For some tasks, the parameter $\theta$ consists of a set of objects $\theta = \{ \theta_1', \dots, \theta_K' \}$ whose order does not matter. For example, in location finding, $K$ represents the number of sources, and each $\theta_k' \in \R^{d'_\Theta}$ is a source location. In this case, we do not want to penalise an action which recovers the object but happens to be in a different order from $\theta$. In this case, we report permutation-invariant versions of the MSE and CMSE by calculating the loss for all $K!$ permutations of $\theta$ and reporting the minimal value. For small values of $K$, this comes with negligible overhead. Location Finding is currently the only task for which this applies. An intuitive alternative is to sort the objects in $\theta$ e.g. by their norm, but this requires the action policy to reproduce that ordering, which is an unnecessarily additional challenge. %

\paragraph{Training and Parametrisation.} Our action policy is parametrised by an MLP having widths $[T \cdot(d_{\XX} + d_\YY), 512, 256, 128, d_\Theta]$ and using GELU activations \citep{hendrycks2016gaussian}. We use the same MLP architecture for our action policy across all tasks. Note this is the same action policy architecture used for ActionBED \citep{rossa2026actionbed} as well. We train for $25\,000$ gradient steps using Adam \citep{kingma2014adam} with $\beta =(0.8, 0.998)$ and an initial learning rate of $5 \times 10^{-4}$ which we decay by a factor of $\gamma = 0.95$ every $1\,000$ steps. We use a batch size of $2\,000$ during training and final evaluation uses $8\,192$ rollouts.

\section{Methods}
\label{appendix_sect:methods}

In this section, we provide additional details on each of the methods implemented in \testbed. As discussed in \cref{sect:TestBED}, new methods can be readily implemented within our framework. 

\subsection{Random}

As a simple baseline, \testbed implements \cls{RandomDesigner}, which draws designs i.i.d. from a specified distribution. By default, this distribution is inherited from the design prior of the task the designer is constructed with (\cref{appendix_subsect:design_spaces}). For example, unconstrained spaces use a standard Gaussian, box design spaces use a uniform distribution, and more general reparametrised spaces (e.g., temporal discounting) sample a latent Gaussian before applying the reparametrisation. While \testbed supports the use of general distributions within this designer, and task-specific heuristics could potentially serve as stronger baselines, these defaults provide a convenient starting point.

\subsection{Static}

For a non-adaptive baseline, we consider static design optimisation, where all $T$ designs are jointly optimised up-front. This is provided in \testbed via \cls{StaticDesigner}. In general, many different techniques could be applied towards optimising a static batch of designs, and \testbed is agnostic to this choice. 

In our benchmarks, we train the static designs by maximising the sPCE lower bound on the EIG (\cref{eqn:pce}). For the CES and temporal discounting task we use REINFORCE (\cref{appendix_sect:reinforce}) with a leave-one-out baseline and for all other directly take the gradient of our sPCE estimator. %
During training we use $L = 4\,096$ inner samples on all tasks, except for Location Finding, CES, and Double Pendulum where we use $L = 8\,192$ as the sPCE lower bound saturated for these tasks during training.

All designs are optimised using Adam \citep{kingma2014adam} for $10\,000$ gradient steps at a fixed learning rate of $5 \times 10^{-3}$, except for Double Pendulum where the cost of training at a high values of $T, L$ limited us to $5\, 000$ gradient steps. This was sufficient for convergence across all tasks. To avoid local optima, we train five random restarts in parallel and select the restart that achieved the highest SPCE in a held-out evaluation with $L=10^{6}$ and $2\,048$ rollouts (which is independent of the final evaluation reported in our results). Each random restart is initialised via a standard Gaussian which is reparametrised through the design space of the task if necessary.

\subsection{Deep Adaptive Design (DAD)}
\label{appendix_subsect:dad}

DAD was a method introduced by \citet{foster2021deep} and was the seminal work transitioning BED to a policy-based paradigm. Conceptually, it describes the idea of amortising adaptive design by optimising a (possibly deep) policy against a non-myopic objective, taking the whole $T$-step rollout into account rather than only one step at a time. The key conceptual idea of DAD to use the history of designs and observations itself as the state, rather than relying on any posterior approximations throughout the design process; a number of policy-based approaches in the literature directly apply this key idea: iDAD \citep{ivanova2021implicit}, RL-BOED \citep{blau2022optimizing}, Step-DAD \citep{pmlr-v267-hedman25a}, ActionBED \citep{rossa2026actionbed}, JointVarPost \citep{phillips2026bayesian} and ALINE \citep{huang2025aline} are all, in this sense, DAD methods. Conflating the general idea with the particular implementation details in the original paper (e.g., choice of policy network architecture) is a common source of confusion in the literature.

Throughout this work, when referring to DAD in results we are referring to the specific instantiation of \citet{foster2021deep}, namely a policy trained to maximise the EIG through the sPCE lower bound (\cref{eqn:pce}), with contrastive samples drawn from the prior and one gradient step taken per sPCE evaluation. We emphasize that DAD as a method is not tied to the particular DeepSets architecture used in the original paper, as shown in our experiments. Our hyperparameter settings are given in \cref{tab:dad-hparams}.

\begin{table}[thbp]
    \centering
    \caption{Training hyperparameters for DAD. Values apply to all eight benchmark tasks with exceptions explicitly called out.}
    \vspace{1ex}
    \small
    \begin{tabular}{lp{0.62\linewidth}}
        \toprule
        Parameter & Value \\
        \midrule
        Gradient estimator       & Pathwise (REINFORCE with leave-one-out baseline on CES, Temporal Discounting) \\
        Contrastive samples, $L$ & $2\,000$ \\
        Outer samples, $B$       & $2\,048$ ($1\,024$ on Double Pendulum, Cart Pole due to memory) \\
        Initial learning rate    & $10^{-4}$ \\
        Betas                    & $(0.9, 0.999)$ \\
        Gradient clipping        & $1.0$ (none on Temporal Discounting) \\
        Decay factor, $\gamma$   & None (CES: $0.9938$; Temporal Discounting: $0.96$) \\
        Annealing frequency      & $1\,000$ steps (used on CES and Temporal Discounting only) \\
        \bottomrule
    \end{tabular}
    \label{tab:dad-hparams}
\end{table}

Two points of note. Firstly, within~\cref{tab:benchmark} we use the shared attention design network of \cref{sect:experiments} rather than the DeepSets network variants of the original paper. Secondly, where outcomes are not reparametrisable we estimate the gradient with REINFORCE (\cref{appendix_sect:reinforce}) adding an additional leave-one-out baseline compared to existing literature. This applies to temporal discounting because the responses are discrete, and to CES because its outcome, a censored sigmoid of a Gaussian, places much of its mass on the two censoring atoms, where the pathwise gradient discards the term describing how the censoring probability responds to the design.

\subsection{Joint Variational Posterior (JointVarPost)}
\label{appendix_subsect:jointvarpost}

This approach optimises a design policy by targeting a variational lower bound on EIG (\cref{eqn:ba}).  
\begin{equation} \label{eqn:ba}
    \mathcal{I}_T(\pi) \geq \E_{p(\theta) p(h_T \mid \theta, \pi)} \left[ \log q_\psi(\theta \mid h_T) - \log p(\theta) \right],
\end{equation}

which is tight when $q_\psi$ recovers the true posterior but remains a valid lower bound regardless of the variational fit. \citet{ba_bound} introduced \cref{eqn:ba} as a bound on mutual information and \citet{foster2019variational} first brought this machinery to BED, recognising the equivalence of EIG and MI, thereby introducing a practical EIG objective alongside a complementary marginal-based upper bound. \citet{pmlr-v108-foster20a-unified} proposed jointly optimising the design and variational parameters in a single stochastic gradient procedure. \citet{phillips2026bayesian} generalised this to a policy-based procedure, and we follow their implementation closely in this work, including choices of variational family.

Our implementation co-optimises $q_\psi$ and the policy from scratch, with no separate posterior-pretraining phase \citep{phillips2026bayesian}. The policy network is the shared attention architecture of \cref{sect:experiments} and is identical to the one DAD uses (\cref{appendix_subsect:dad}), so on each task the two methods differ only in the objective that trains the policy. Training hyperparameter settings are provided in \cref{tab:jointvarpost-hparams}. Note that the Barber-Agakov bound has no contrastive axis, relieving a memory pressure that exists within DAD's sPCE-based training (\cref{appendix_subsect:dad}). The gradient estimator follows the same split as DAD: pathwise where the outcomes are reparametrisable, and REINFORCE with a leave-one-out baseline on temporal discounting and CES.

\begin{figure}[!t]
\centering
\resizebox{\linewidth}{!}{%
\begin{tikzpicture}[
  font=\sffamily,
  >={Stealth[length=2.0mm,width=1.4mm]},
  box/.style={
    rounded corners=2pt,
    draw,
    line width=0.6pt,
    align=center,
    inner sep=3pt,
    text centered,
    font=\sffamily\scriptsize,
    minimum width=2.7cm,
    minimum height=0.62cm
  },
  bar/.style={box, fill=polfill, draw=polline},
  embbox/.style={box, fill=taskfill, draw=taskline},
  attnbox/.style={box, fill=objfill, draw=objline, minimum height=0.76cm},
  ffbox/.style={box, fill=taskfill, draw=taskline, minimum height=0.76cm},
  clsbox/.style={box, fill=evfill, draw=evline, minimum width=1.4cm, minimum height=0.60cm},
  headbox/.style={box, fill=evfill, draw=evline, minimum width=1.35cm, minimum height=0.52cm},
  outbox/.style={box, fill=polfill, draw=polline},
  sumnode/.style={
    circle, draw=black!55, line width=0.6pt, fill=white,
    inner sep=0pt, minimum size=4.2mm, font=\sffamily\tiny
  },
  flow/.style={->, line width=0.7pt, draw=black!58},
  res/.style={->, line width=0.7pt, draw=black!58, rounded corners=2.5pt},
  back/.style={->, line width=0.75pt, draw=black!52, rounded corners=2.5pt},
  lbl/.style={font=\sffamily\tiny, text=black!62},
  tag/.style={font=\sffamily\tiny, inner sep=1.2pt}
]

\node[lbl] (inp) at (0,-0.72) {history $(x_t, y_t)_{t \le T}$};

\draw[rounded corners=4pt, fill=black!3, draw=taskline, line width=0.7pt]
  (-2.10,-0.30) rectangle (3.66,11.30);
\node[lbl, anchor=west, text=taskline] at (-2.10,11.52) {Encoder};

\draw[rounded corners=3pt, draw=black!45, line width=0.5pt, dashed,
      dash pattern=on 1.6pt off 1.4pt]
  (-1.58,-0.02) rectangle (1.58,2.62);
\node[lbl, anchor=west] at (1.70,1.30) {Token Embedding};

\node[embbox, minimum height=0.52cm] (lin1) at (0,0.45) {Linear \, {\tiny $[d_\XX{+}d_\YY, 256]$}};
\node[embbox, minimum height=0.52cm] (gelu) at (0,1.31) {GELU};
\node[embbox, minimum height=0.52cm] (lin2) at (0,2.17) {Linear \, {\tiny $[256, 128]$}};

\node[sumnode] (plus1) at (0,2.98) {$+$};
\node[lbl, anchor=east] (temb) at (3.52,2.98) {Time Embedding};

\node[embbox, minimum height=0.52cm] (cat) at (0,3.79) {Append CLS Token};
\node[clsbox] (cls) at (2.55,3.79) {\sd{learned}\\[-1pt]CLS};

\node[bar, minimum height=0.52cm] (ln0) at (0,4.80) {LayerNorm};

\draw[flow] (inp.north)   -- (lin1.south);
\draw[flow] (lin1.north)  -- (gelu.south);
\draw[flow] (gelu.north)  -- (lin2.south);
\draw[flow] (lin2.north)  -- (plus1.south);
\draw[flow] (plus1.north) -- (cat.south);
\draw[flow] (temb.west)   -- (plus1.east);
\draw[flow] (cls.west)    -- (cat.east);
\draw[flow] (cat.north)   -- node[tag, right=1pt] {$T{+}1$ tokens} (ln0.south);

\draw[rounded corners=3pt, fill=black!9, draw=black!30, line width=0.5pt]
  (-1.90,5.33) rectangle (2.34,10.77);
\node[lbl, anchor=west] at (-1.90,10.99) {\texttt{num\_blocks}$\times$};

\node[bar, minimum height=0.52cm] (ln1)  at (0,5.88) {LayerNorm};
\node[attnbox, minimum height=0.64cm] (mha)  at (0,6.80)
{Multi-Head Attention\\[-1pt]{\tiny \texttt{num\_heads}$\times$ heads, $[128, 128]$}};
\node[sumnode] (add1) at (0,7.67) {$+$};
\node[bar, minimum height=0.52cm] (ln2)  at (0,8.48) {LayerNorm};
\node[ffbox, minimum height=0.64cm] (ff)   at (0,9.40) {Feed Forward\\[-1pt]{\tiny $[128, 512, 128]$, GELU}};
\node[sumnode] (add2) at (0,10.27) {$+$};

\draw[flow] (ln0.north)  -- (ln1.south);
\draw[flow] (ln1.north)  -- (mha.south);
\draw[flow] (mha.north)  -- (add1.south);
\draw[flow] (add1.north) -- (ln2.south);
\draw[flow] (ln2.north)  -- (ff.south);
\draw[flow] (ff.north)   -- (add2.south);

\draw[res] (0,5.47) -- (1.82,5.47) -- (1.82,7.67) -- (add1.east);
\draw[res] (0,8.05) -- (1.82,8.05) -- (1.82,10.27) -- (add2.east);

\draw[rounded corners=4pt, fill=black!3, draw=evline, line width=0.7pt]
  (5.50,-0.05) rectangle (10.95,8.62);
\node[lbl, anchor=east, text=evline] at (10.95,8.86) {Gaussian mixture decoder};

\node[bar] (read) at (7.80,0.45) {CLS Readout};
\node[bar] (ln3)  at (7.80,1.60) {LayerNorm};

\draw[rounded corners=3pt, draw=black!45, line width=0.5pt, dashed,
      dash pattern=on 1.6pt off 1.4pt]
  (6.06,2.38) rectangle (9.54,4.57);
\node[lbl, anchor=west] at (9.68,3.48) {Trunk};

\node[outbox] (tlin)  at (7.80,2.90) {Linear \, {\tiny $[128, 128]$}};
\node[outbox] (tgelu) at (7.80,4.05) {GELU};

\node[headbox, minimum height=0.70cm] (hmean)  at (6.35,5.35) {means\\[-2pt]{\tiny $mD$}};
\node[headbox, minimum height=0.70cm] (hscale) at (7.80,5.35) {log std.\\[-2pt]devs.\\[-2pt]{\tiny $mD$}};
\node[headbox, minimum height=0.70cm] (hlogit) at (9.25,5.35) {logits\\[-2pt]{\tiny $m$}};

\node[attnbox, minimum width=3.3cm] (mix) at (7.80,6.70)
{Gaussian Mixture\\[-1pt]{\tiny $q_\psi(z \mid h_T)$, $m = 10$ components}};

\node[outbox, minimum width=3.0cm] (push) at (7.80,8.00) {Pushforward $z \mapsto \theta$};
\node[lbl] (out) at (7.80,9.30) {$q_\psi(\theta \mid h_T)$};

\draw[flow] (read.north)   -- (ln3.south);
\draw[flow] (ln3.north)    -- (tlin.south);
\draw[flow] (tlin.north)   -- (tgelu.south);
\draw[flow] (tgelu.north)  -- (hmean.south);
\draw[flow] (tgelu.north)  -- (hscale.south);
\draw[flow] (tgelu.north)  -- (hlogit.south);
\draw[flow] (hmean.north)  -- (mix.south);
\draw[flow] (hscale.north) -- (mix.south);
\draw[flow] (hlogit.north) -- (mix.south);
\draw[flow] (mix.north)    -- (push.south);
\draw[flow] (push.north)   -- (out.south);

\draw[back]
  (add2.north) -- (0,11.70)
  -- node[tag, above=0.5pt, pos=0.45] {encoded history}
     (4.75,11.70)
  -- (4.75,-0.45) -- (7.80,-0.45) -- (read.south);

\end{tikzpicture}%
}
\caption{The variational posterior network used by JointVarPost. Each
history pair is embedded, a time embedding is added, and a learned CLS token is
appended as an extra position; the resulting $T+1$ tokens are normalised and
then pass through the pre-norm transformer blocks, each applying
$x \mapsto x + \mathrm{MHA}(\mathrm{LN}(x))$ and then
$x \mapsto x + \mathrm{MLP}(\mathrm{LN}(x))$. The CLS position is read out and
passed to the mixture head, whose three linear heads parametrise a
$m$-component Gaussian mixture over unconstrained coordinates $z$, which a
task-specific bijection pushes onto the support of $\theta$. Note, each history token $t$
carries the time embedding of $t$ while the CLS token carries the embedding of index $t$, the position it occupies at the end of the history so that the readout knows how many experiments have been run. Additional details
are provided in \cref{tab:jointvarpost-posterior-net}.}
\label{fig:posterior_net}
\end{figure}
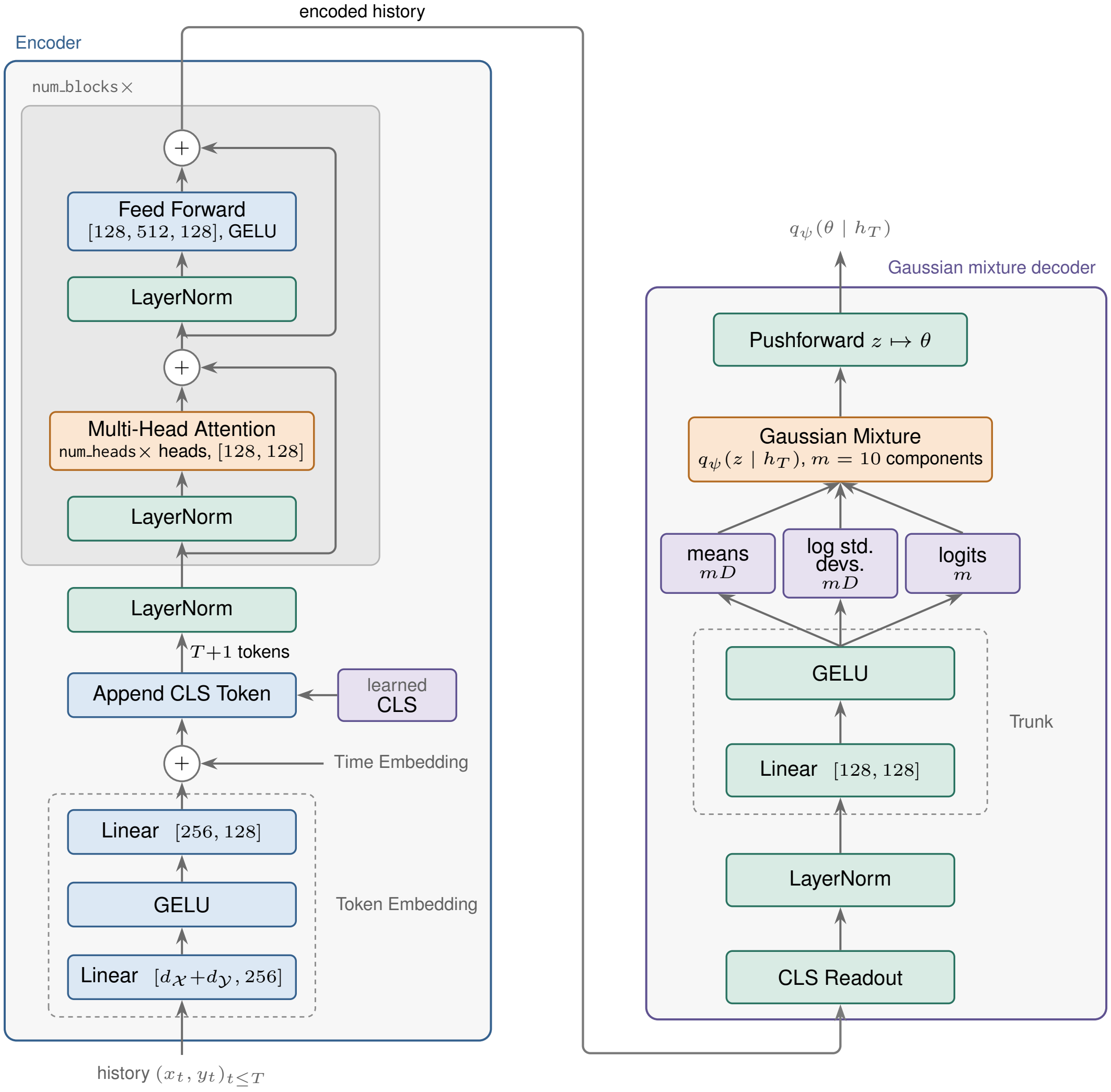

\looseness=-1
The variational posterior is a $10$-component conditional mixture of Gaussians with an attention encoder that mirrors the design network \citep{phillips2026bayesian}. Its architecture is shown in \cref{fig:posterior_net} and set out in full in \cref{tab:jointvarpost-posterior-net}. Since the encoder dominates the parameter count and only the output heads differ, the posterior network has $\sim\!670$k trainable parameters across the eight tasks, closely matching the $\sim\!650$k of the attentive design network it is trained alongside. When $\theta$ is constrained, the mixture is defined over unconstrained coordinates that are pushed forward onto the support of the prior through a task/$\theta$-specific bijection: an exponential map where $\theta$ is strictly positive, a squashed affine map where it is bounded to an interval, and a stick-breaking map for the simplex-valued component of CES. Placing a Gaussian mixture in $\theta$'s own coordinate space would otherwise put mass outside the support and, where $\theta$ occupies a lower-dimensional manifold as in CES, give a density with respect to the wrong measure, leaving the Barber-Agakov gain unbounded above and the bound meaningless. On location finding we additionally canonicalise the source ordering, since the $K = 2$ sources are exchangeable and the label-switching symmetry would otherwise have to be learned.

\begin{table}[thbp]
    \centering
    \caption{Training hyperparameters for JointVarPost. Values apply to all eight benchmark tasks with exceptions explicitly called out. The policy and the posterior networks are optimised jointly from scratch, so there is no pretraining phase unlike variants within \citet{phillips2026bayesian}.}
    \vspace{1ex}
    \small
    \begin{tabular}{lp{0.6\linewidth}}
        \toprule
        Parameter & Value \\
        \midrule
        Gradient estimator            & Pathwise (REINFORCE with leave-one-out baseline on CES, Temporal Discounting) \\
        Outer samples, $B$            & $2\,048$ \\
        Posterior pretraining steps   & $0$ \\
        Mixture components            & $10$ \\
        Policy learning rate          & $10^{-5}$, cosine decay to $5 \times 10^{-6}$ \\
        Policy betas                  & $(0.9, 0.999)$ \\
        Policy gradient clipping      & $1.0$ \\
        Policy EMA decay              & $0.99$ \\
        Posterior net learning rate       & Peak $10^{-3}$, warmup over the first $0.1\%$ of steps, cosine decay to $10^{-6}$ \\
        Posterior net betas               & $(0.9, 0.999)$ \\
        Posterior net gradient clipping   & $1.0$ \\
        Schedule horizon              & $200\,000$ steps ($280\,000$ on Michaelis-Menten, Temporal Discounting; $600\,000$ on CES, Pharmacokinetic) \\
        \bottomrule
    \end{tabular}
    \label{tab:jointvarpost-hparams}
\end{table}

\begin{table}[thbp]
    \centering
    \caption{Architecture of the variational posterior network used within the JointVarPost method. The encoder mirrors the shared design network of \cref{sect:experiments}; the mixture readout replaces its design head. $D$ denotes the dimension of the space the mixture is defined over, which is $d_\Theta$ where $\theta$ is unconstrained and the dimension of the unconstrained coordinates otherwise.}
    \vspace{1ex}
    \small
    \begin{tabular}{lp{0.58\linewidth}}
        \toprule
        Component & Value \\
        \midrule
        Token encoder        & MLP with widths $[d_\XX + d_\YY, 256, 128]$ and GELU after each hidden layer (no activation on the output) \\
        Time embedding       & Learned, dimension $128$ \\
        Transformer blocks   & $3$ blocks, width $128$, $4$ heads \\
        Pooling              & Learned CLS token \\
        Trunk                & Linear $128 \to 128$ followed by GELU \\
        Output heads         & Three linear heads from the trunk: component means ($m D$), diagonal log standard deviations ($m D$), mixing logits ($m$) \\
        Mixture components, $m$ & $10$ \\
        Log std.\ dev.\ range & Clamped to $[-7, 3]$ \\
        Support transform    & Identity (Location Finding); exponential (Pharmacokinetic, Stochastic Pendulum, Cart Pole, Double Pendulum); exponential and affine (Temporal Discounting); squashed affine (Michaelis-Menten); squashed affine, stick-breaking and exponential (CES) \\
        Canonicalisation     & Source ordering sorted on Location Finding only \\
        Trainable parameters & $669$k to $675$k depending on the task \\
        \bottomrule
    \end{tabular}
    \label{tab:jointvarpost-posterior-net}
\end{table}

\subsection{RL-BOED}
\label{appendix_subsect:rlboed}

RL-BOED~\citep{blau2022optimizing} formulates sequential BED as a Markov decision process~\citep[MDP,][]{bellman1957markovian, puterman1994markov} and learns the design policy with reinforcement learning. It was introduced to address two shortcomings of DAD, that of requiring differentiable likelihoods and continuous design spaces. 
Beyond the use of RL algorithms for policy learning, the main difference to DAD is a dense reward over time steps: the sPCE is decomposed into a telescoping sum where the $t$-th term is the contribution of the $t$-th experiment. This keeps the terminal objective unchanged while improving credit assignment.

The implementation in \testbed is written in plain PyTorch without any dependencies on third-party RL libraries. The only methodological difference to the authors' original code\footnote{Available at \url{https://github.com/csiro-mlai/RL-BOED}.} is that we use Soft Actor-Critic~\citep[SAC,][]{haarnoja2018softa} instead of REDQ to optimize the policy.\footnote{Our implementation of SAC follows that of CleanRL~\citep{huang2022cleanrl}.} REDQ is itself a modification of SAC~(it adds an ensemble of critics and a
high update-to-data ratio) designed to improve sample efficiency in
settings where environment interaction is costly. In sequential BED the
environment is the probabilistic model itself, so transitions are~(usually) cheap to
simulate and sample efficiency is not the binding constraint; the additional
critic updates therefore buy little at considerable computational cost. Consistent
with this, our implementation matches or exceeds the performance reported by
\citet{blau2022optimizing} on all tasks they consider.

While we only implement SAC, \testbed already provides the building blocks
common to modern off-policy RL algorithms: actor and critic networks, a replay
buffer, and the environment wrapper that exposes the sequential BED problem as
an MDP~(including the telescoping stage rewards). Implementing an alternative algorithm therefore amounts to overriding the train step for RL-BOED, leaving the remainder of the
training loop untouched. REDQ, for instance, differs from SAC only in the number
of critics and the update-to-data ratio, and hence requires minimal changes to the existing code.

Our implementation also uses the attention-based architecture for the design policy~(the actor) described in \cref{appendix_sect:architectures}, but the same critic networks as the original implementation, and the hyperparameters are also the same, with some task-specific variations. Our hyperparameter settings are given in \cref{tab:rlboed-hparams}, with the settings that vary by task given in \cref{tab:rlboed-task-hparams}. For instance, on Stochastic Pendulum and Cart Pole we use a discount factor $\gamma=0.99$ instead of the default $\gamma=0.9$ from the paper.

\begin{table}[thbp]
    \centering
    \caption{Training setup and hyperparameters for RL-BOED. Values apply to all eight benchmark tasks with exceptions explicitly mentioned. The bounded design space and outcome normalisation range used for each task are given in \cref{tab:rlboed-task-hparams}.}
    \vspace{1ex}
    \small
    \begin{tabular}{lp{0.58\linewidth}}
        \toprule
        Parameter & Value \\
        \midrule
        RL algorithm                  & Soft Actor-Critic with twin critics and a learned temperature \\
        Actor network                 & Shared attention architecture (\cref{appendix_sect:architectures}): encoder width $256$, embedding dimension $128$, $4$ heads, $3$ blocks \\
        Critic networks               & DeepSets encoder and readout MLP with hidden widths $[128, 128]$, embedding dimension $64$, as in the original implementation \\
        Policy head                   & Gaussian squashed by $\tanh$ into the design box, with $\log \sigma \in [-20, 0]$ and initial $\sigma = 3^{-1/2}$ before squashing \\
        Reward                        & Dense (telescoping) sPCE reward with $L = 100\,000$ contrastive samples \\
        Episodes per iteration        & $100$ (parallel environments) \\
        Gradient steps per iteration  & $64$ \\
        Batch size                    & $4{,}096$ transitions \\
        Replay buffer size            & $10^{6}$ transitions ($10^{7}$ on Location Finding) \\
        Learning starts after         & $100{,}000$ transitions \\
        Optimiser                     & Adam \\
        Critic learning rate          & $3 \times 10^{-4}$ \\
        Actor learning rate           & $3 \times 10^{-4}$ ($10^{-4}$ on Location Finding); also used for the temperature \\
        Discount factor, $\gamma$     & $0.9$ ($0.99$ on Stochastic Pendulum, Cart Pole) \\
        Target smoothing, $\tau$      & $0.001$ ($0.005$ on Temporal Discounting, CES, Pharmacokinetic, Michaelis-Menten) \\
        Initial temperature           & $1$ \\
        Target entropy                & Initially $-d_\XX$ (the negative design dimension), reduced by $7.143 \times 10^{-5}$ per iteration up to a total reduction of $1.429$ (reached after about $20{,}000$ iterations) \\
        \bottomrule
    \end{tabular}
    \label{tab:rlboed-hparams}
\end{table}

\begin{table}[thbp]
    \centering
    \caption{Task-specific RL-BOED settings. \emph{Design box} is the bounded design space used to train and evaluate RL-BOED (\cref{appendix_subsect:rlboed}), and equals the task's own design space except where marked $^\dagger$: on Location Finding the task is unbounded, on Temporal Discounting it is $D_1 \ge 0$, $R_0 \in [0, 100]$, and on CES it is $[0, 100]^6$. \emph{Outcome range} is the reference range used to rescale outcomes before they enter the actor and critics; outcomes outside it are not clipped.}
    \vspace{1ex}
    \small
    \begin{tabular}{lll}
        \toprule
        Task & Design box & Outcome range \\
        \midrule
        Location Finding$^\dagger$        & $[-4, 4]^2$                          & $[-3, 10]$   \\
        Temporal Discounting$^\dagger$    & $D_1 \in [0.5, 200]$, $R_0 \in [1, 99]$ & $[0, 1]$     \\
        CES$^\dagger$                     & $[0.01, 100]^6$                      & $[0, 1]$     \\
        Pharmacokinetic                   & $[0, 24]$                            & $[-2, 70]$   \\
        Michaelis-Menten                  & $[0, 400]$                           & $[-10, 210]$ \\
        Stochastic Pendulum               & $[-1, 1]$                            & $[-10, 10]$  \\
        Cart Pole                         & $[-5, 5]$                            & $[-10, 10]$  \\
        Double Pendulum                   & $[-4, 4] \times [-2, 2]$             & $[-10, 10]$  \\
        \bottomrule
    \end{tabular}
    \label{tab:rlboed-task-hparams}
\end{table}

RL-BOED requires a bounded design space. We therefore train it on the boxes given in \cref{tab:rlboed-task-hparams}, which for Location Finding and Temporal Discounting are tighter than the design spaces of the tasks themselves, and for CES differ in having a lower bound of $0.01$ rather than $0$. These bounds restrict only which designs the policy is able to propose: the prior and likelihood of every task are unchanged, and the sPCE and sNMC estimators used in our final evaluation depend only on these. The reported bounds are therefore estimates of the same EIG as for the other methods, although the RL-BOED policy is trained and evaluated on the bounded space.

\subsection{ALINE}
\label{appendix_subsect:aline}

ALINE~\citep{huang2025aline} is an amortized algorithm for BED that jointly trains a design policy and an inference network for posterior distributions. It uses an Attention-based architecture, in particular a Transformer Neural Process~\citep{nguyen2022transformer}, that serves as a common spine for both the design and inference networks, which are learned jointly. The inference network is trained to minimize the negative log likelihood of true parameter values, while the design network is trained to maximize the sum of the incremental EIGs over a rollout.

The implementation of ALINE in \testbed is faithful to the authors' specification and their official code,\footnote{Available at \url{https://github.com/huangdaolang/aline}.} with the following exceptions:
\begin{itemize}
    \item The \testbed implementation only supports maximizing the EIG in the full set of hidden parameters of a task, just like the other methods. The original method is more general, and allows targeting the EIG in a subset of the hidden parameters as well as the expected predictive information gain~\citep[EPIG,][]{bickfordsmith2023prediction}. These cases can still be handled in \testbed, but require a new \cls{Task} definition rather than a change in the algorithm's configuration.
    \item The original implementation always samples the first design randomly, then the rest are sampled from the policy. The \testbed implementation samples all designs from the policy, to be consistent with the other methods. Consequently, the policy is rewarded for all $T$ designs rather than the last $T-1$, and the inference network is also trained on the empty history, i.e., to predict the prior.
    \item ALINE training is split into two phases, where in the first phase only the inference network is trained. During this phase, in the original code the designs are sampled from the untrained policy network. Here we choose to sample them from a uniform distribution over the design space instead.
    \item \testbed has an option to apply a fixed affine rescaling to the designs and outcomes before they are passed to the network, which is disabled by default.
\end{itemize}

Hyperparameters for our benchmark runs are given in \cref{tab:aline-hparams}, network sizes in \cref{tab:aline-sizes}, and settings that vary by task in \cref{tab:aline-task-hparams}. The reproduction below instead uses the original hyperparameters and network sizes. We enable the fixed rescaling of designs and outcomes on every task except Location Finding and Pharmacokinetic. ALINE selects each design from a finite candidate pool sampled from a box, so, as with RL-BOED (\cref{appendix_subsect:rlboed}), it requires a bounded design space and we train it on the boxes in \cref{tab:aline-task-hparams}. This restricts which designs the policy can propose without changing the prior or likelihood of any task, so its sPCE and sNMC estimates are of the same EIG as for the other methods, although the ALINE policy is trained and evaluated on the bounded space.

\begin{table}[thbp]
    \centering
    \caption{Training hyperparameters for ALINE. Values apply to all eight benchmark tasks with exceptions explicitly mentioned. Network sizes are given in \cref{tab:aline-sizes}, and the bounded design space and input rescaling used for each task in \cref{tab:aline-task-hparams}.}
    \vspace{1ex}
    \small
    \begin{tabular}{lp{0.58\linewidth}}
        \toprule
        Parameter & Value \\
        \midrule
        Network                        & Shared transformer spine for design and inference networks; embedding dimension $128$, feedforward dimension $384$, $4$ heads, $3$ layers, no dropout, $10$-component Gaussian mixture posterior head \\
        Warm-up iterations             & $20{,}000$ (inference network only, with designs drawn uniformly from the design space) \\
        Outer samples per iteration    & $200$ rollouts \\
        Candidate pool, training       & $200$ designs drawn afresh for each rollout \\
        Candidate pool, evaluation     & $2{,}000$ designs, with the policy acting deterministically \\
        Optimiser                      & AdamW, weight decay $0.01$ \\
        Learning rate                  & Warm-up: $10^{-3}$ with cosine decay over the full $200{,}000$ iterations. Joint phase: optimiser and cosine schedule restarted, with $10^{-3}$ for the acquisition head and $2 \times 10^{-4}$ for all other parameters \\
        Gradient clipping              & $1.0$ (on the $\ell_\infty$ norm of the gradient) \\
        Policy loss weight, $\alpha$   & $1.0$ \\
        Reward discount, $\gamma$      & $1.0$ \\
        Policy reward                  & Incremental posterior log-density gains, clipped at zero and standardised across the batch \\
        Canonicalise $\theta$          & Location Finding only (its two sources are exchangeable) \\
        \bottomrule
    \end{tabular}
    \label{tab:aline-hparams}
\end{table}

\begin{table}[thbp]
    \centering
    \caption{Task-specific ALINE settings. \emph{Design box} is the bounded design space used to train and evaluate ALINE (\cref{appendix_subsect:aline}), and equals the task's own design space except where marked $^\dagger$; as with RL-BOED, the Location Finding and Temporal Discounting tasks are modified. Designs and outcomes are rescaled elementwise as $(x - c)/s$ before being passed to the network; the table gives the centre and scale as $(c;\, s)$. Outcomes are not clipped. A single value applies to every coordinate, and a dash denotes no rescaling.}
    \vspace{1ex}
    \small
    \begin{tabular}{lp{0.22\linewidth}p{0.2\linewidth}p{0.2\linewidth}}
        \toprule
        Task & Design box & Design $(c;\, s)$ & Outcome $(c;\, s)$ \\
        \midrule
        Location Finding$^\dagger$     & $[-4, 4]^2$ & --- & --- \\
        Temporal Discounting$^\dagger$ & $[0.5, 200] \times [1, 99]$ & $(100.25, 50);$\newline $(99.75, 49)$ & $0.5;\, 0.5$ \\
        CES                            & $[0, 100]^6$ & $50;\, 50$ & $0.5;\, 0.5$ \\
        Pharmacokinetic                & $[0, 24]$ & --- & --- \\
        Michaelis-Menten               & $[0, 400]$ & $200;\, 200$ & $100;\, 100$ \\
        Stochastic Pendulum            & $[-1, 1]$ & --- & $0;\, (3, 8)$ \\
        Cart Pole                      & $[-5, 5]$ & $0;\, 5$ & $0;\, (10, 2, 8, 5)$ \\
        Double Pendulum                & $[-4, 4] \times [-2, 2]$ & $0;\, (4, 2)$ & $0;\, (8, 10, 12, 25)$ \\
        \bottomrule
    \end{tabular}
    \label{tab:aline-task-hparams}
\end{table}

To verify that our changes to ALINE do not affect its performance, we train it on the BED task specifications of the original paper~(which differ slightly from those in \testbed), using the original hyperparameters and network sizes, which are smaller than in our benchmark. To make the results in \cref{tab:aline-reproduction} directly comparable, we also follow the original evaluation protocol rather than our own. Since sPCE is a lower bound that grows with the number of contrastive samples $L$, we use the original values: $L=10^6$ for location finding and $L=10^7$ for CES. We also match the original candidate pool: 2000 designs, drawn afresh for each rollout, from which the policy greedily picks one not-yet-chosen design at each step. We train ALINE with 5 random seeds and evaluate the final policy of each on 2048 rollouts. We report the mean sPCE over seeds, with a 95\% confidence interval computed from the standard errors over rollouts.

The CES task in \testbed differs from the one in the original paper in how the scale parameter $u$ of the latent Gaussian is represented: \testbed includes $u \sim \mathrm{LogNormal}(1, 3)$ itself in the hidden parameters, whereas the original includes $\log u \sim \mathcal{N}(1, 3^2)$. This does not change the EIG, but it gives ALINE's inference network a heavy-tailed target. In our benchmark, training on this version of the task was numerically unstable unless the designs and outcomes were rescaled to $[-1, 1]$ before being passed to the network, so we enable this rescaling for CES even though the original method has no such step. The reproduction in \cref{tab:aline-reproduction}, which uses the original $\log u$ parameterization, trained stably without it.

\begin{table}[t]
  \centering
  \caption{Network sizes of ALINE in the original paper, which we also use for the reproduction in \cref{tab:aline-reproduction}, and in our benchmark. Both use 4 attention heads, 3 transformer layers and a 10-component Gaussian mixture posterior head. The design policy comprises every parameter except the posterior head. Parameter counts are for location finding; the counts for CES are within 1\% of these.}
  \vspace{1ex}
  \label{tab:aline-sizes}
  \begin{tabular}{lrr}
    \toprule
    & Original & Benchmark \\
    \midrule
    Embedding dimension   & 32    & 128   \\
    Feedforward dimension & 128   & 384   \\
    Parameters (total)    & 97.5K & 1.15M \\
    Parameters (policy)   & 51.4K & 647K  \\
    \bottomrule
  \end{tabular}
\end{table}

\begin{table}[t]
  \centering
  \caption{EIG lower bound~(sPCE) of ALINE on the location finding and CES tasks of \citet{huang2025aline}. \emph{Original}: numbers from Table~2 of \citet{huang2025aline}. \emph{Ours}: our reimplementation, trained on the same task specifications and network sizes.}
  \vspace{1ex}
  \label{tab:aline-reproduction}
  \begin{tabular}{lcc}
    \toprule
    & Location Finding ($T=30$) & Constant Elasticity of Substitution ($T=10$) \\
    \midrule
    Original & $8.91 \pm 0.04$ & $13.50 \pm 0.15$ \\
    Ours     & $8.92 \pm 0.02$ & $13.50 \pm 0.06$ \\
    \bottomrule
  \end{tabular}
\end{table}

\subsection{ActionBED}

ActionBED \citep{rossa2026actionbed} is a policy learning method for BED which jointly optimises an action policy $\pi_\AA$ alongside the design policy $\pi_\XX$ in order to minimise a downstream loss function $\ell(a, \theta)$ defined between actions $a \in \AA$ and the parameters $\theta \in \Theta$, i.e., the objective is $\min_{\pi_\AA, \pi_\XX} \E_{p(\theta) p(h_T \mid \theta, \pi)}[\ell(\pi_\AA(h_T), \theta)]$. The action policy $\pi_\AA$ is a small MLP, parametrised as described in \cref{appendix_sect:downstream_losses}. Both networks are trained using a single optimiser. 

We compare several different losses for ActionBED in this work: $\ell_{\mathrm{MSE}}$ and $\ell_{\mathrm{CMSE}}$ on the original scale of $\theta$, as well as $\ell_{\mathrm{MSE-L}}$ and $\ell_{\mathrm{CMSE-L}}$, i.e., the MSE and CMSE losses on $\vartheta = \log \theta$ as the prediction target. See \cref{appendix_sect:downstream_losses}. Where available, we follow the optimisation and training procedures described in the original paper. We summarise our hyperparameter choices in~\cref{tab:actionbed-hyperparameters}.

\begin{table}[thbp]
    \centering
    \caption{Hyperparameters for ActionBED method, shared across all tasks unless stated otherwise.}
    \vspace{1ex}
    \small
    \begin{tabular}{lp{0.58\linewidth}}
        \toprule
        Component & Value \\
        \midrule
        Action network       & MLP with widths $[T(d_\XX + d_\YY), 512, 256, 128, d_\Theta]$ and GELU after each hidden layer (no activation on the output) \\
        Action net.\ parameters & $175$k to $319$k depending on the task \\
        Gradient estimator   & Pathwise; REINFORCE with a leave-one-out baseline on Temporal Discounting \\
        Action warm-up       & $25{,}000$ steps training on designs sampled at random from the design space \\
        Batch size           & $2{,}000$ rollouts per step \\
        Optimiser            & Adam, learning rate $5 \times 10^{-4}$, $\beta = (0.8, 0.998)$ \\
        Learning rate schedule & Multiplied by $0.95$ every $2{,}000$ steps after warm-up \\
        Gradient clipping    & $750$ by default; $5$ on Temporal Discounting \\
        \bottomrule
    \end{tabular}
    \label{tab:actionbed-hyperparameters}
\end{table}

\newpage

\section{Network Architectures}
\label{appendix_sect:architectures}

\begin{figure}[!t]
\centering
\resizebox{\linewidth}{!}{%
\begin{tikzpicture}[
  font=\sffamily,
  >={Stealth[length=2.0mm,width=1.4mm]},
  box/.style={
    rounded corners=2pt,
    draw,
    line width=0.6pt,
    align=center,
    inner sep=3pt,
    text centered,
    font=\sffamily\scriptsize,
    minimum width=2.7cm,
    minimum height=0.62cm
  },
  bar/.style={box, fill=polfill, draw=polline},
  embbox/.style={box, fill=taskfill, draw=taskline},
  attnbox/.style={box, fill=objfill, draw=objline, minimum height=0.76cm},
  ffbox/.style={box, fill=taskfill, draw=taskline, minimum height=0.76cm},
  clsbox/.style={box, fill=evfill, draw=evline, minimum width=1.4cm, minimum height=0.60cm},
  timebox/.style={clsbox, fill=white, draw=objline, dashed, dash pattern=on 1.6pt off 1.4pt},
  outbox/.style={box, fill=polfill, draw=polline},
  sumnode/.style={
    circle, draw=black!55, line width=0.6pt, fill=white,
    inner sep=0pt, minimum size=4.2mm, font=\sffamily\tiny
  },
  flow/.style={->, line width=0.7pt, draw=black!58},
  toggle/.style={flow, dashed, dash pattern=on 1.6pt off 1.4pt},
  res/.style={->, line width=0.7pt, draw=black!58, rounded corners=2.5pt},
  back/.style={->, line width=0.75pt, draw=black!52, rounded corners=2.5pt},
  lbl/.style={font=\sffamily\tiny, text=black!62},
  tag/.style={font=\sffamily\tiny, inner sep=1.2pt}
]

\node[lbl] (inp) at (0,-0.72) {history $(x_s, y_s)_{s \le t}$};

\draw[rounded corners=4pt, fill=black!3, draw=taskline, line width=0.7pt]
  (-2.10,-0.30) rectangle (3.66,11.30);
\node[lbl, anchor=west, text=taskline] at (-2.10,11.52) {Encoder};

\draw[rounded corners=3pt, draw=black!45, line width=0.5pt, dashed,
      dash pattern=on 1.6pt off 1.4pt]
  (-1.58,-0.02) rectangle (1.58,2.62);
\node[lbl, anchor=west] at (1.70,1.30) {Token Embedding};

\node[embbox, minimum height=0.52cm] (lin1) at (0,0.45) {Linear \, {\tiny $[d_\XX{+}d_\YY, 256]$}};
\node[embbox, minimum height=0.52cm] (gelu) at (0,1.31) {GELU};
\node[embbox, minimum height=0.52cm] (lin2) at (0,2.17) {Linear \, {\tiny $[256, 128]$}};

\node[sumnode] (plus1) at (0,2.98) {$+$};
\node[timebox] (temb) at (2.55,2.98) {\sd{on / off}\\[-1pt]Time Emb.};

\node[embbox, minimum height=0.52cm] (cat) at (0,3.79) {Append CLS Token};
\node[clsbox] (cls) at (2.55,3.79) {\sd{learned}\\[-1pt]CLS};

\node[bar, minimum height=0.52cm] (ln0) at (0,4.80) {LayerNorm};

\draw[flow] (inp.north)   -- (lin1.south);
\draw[flow] (lin1.north)  -- (gelu.south);
\draw[flow] (gelu.north)  -- (lin2.south);
\draw[flow] (lin2.north)  -- (plus1.south);
\draw[flow] (plus1.north) -- (cat.south);
\draw[toggle] (temb.west) -- (plus1.east);
\draw[flow] (cls.west)    -- (cat.east);
\draw[flow] (cat.north)   -- node[tag, right=1pt] {$t{+}1$ tokens} (ln0.south);

\draw[rounded corners=3pt, fill=black!9, draw=black!30, line width=0.5pt]
  (-1.90,5.33) rectangle (2.34,10.77);
\node[lbl, anchor=west] at (-1.90,10.99) {\texttt{num\_blocks}$\times$};

\node[bar, minimum height=0.52cm] (ln1)  at (0,5.88) {LayerNorm};
\node[attnbox, minimum height=0.64cm] (mha)  at (0,6.80)
{Multi-Head Attention\\[-1pt]{\tiny \texttt{num\_heads}$\times$ heads, $[128, 128]$}};
\node[sumnode] (add1) at (0,7.67) {$+$};
\node[bar, minimum height=0.52cm] (ln2)  at (0,8.48) {LayerNorm};
\node[ffbox, minimum height=0.64cm] (ff)   at (0,9.40) {Feed Forward\\[-1pt]{\tiny $[128, 512, 128]$, GELU}};
\node[sumnode] (add2) at (0,10.27) {$+$};

\draw[flow] (ln0.north)  -- (ln1.south);
\draw[flow] (ln1.north)  -- (mha.south);
\draw[flow] (mha.north)  -- (add1.south);
\draw[flow] (add1.north) -- (ln2.south);
\draw[flow] (ln2.north)  -- (ff.south);
\draw[flow] (ff.north)   -- (add2.south);

\draw[res] (0,5.47) -- (1.82,5.47) -- (1.82,7.67) -- (add1.east);
\draw[res] (0,8.05) -- (1.82,8.05) -- (1.82,10.27) -- (add2.east);

\draw[rounded corners=4pt, fill=black!3, draw=evline, line width=0.7pt]
  (5.50,-0.05) rectangle (10.95,7.25);
\node[lbl, anchor=east, text=evline] at (10.95,7.49) {Design decoder};

\node[bar] (read) at (7.80,0.45) {CLS Readout};
\node[bar] (ln3)  at (7.80,1.60) {LayerNorm};

\draw[rounded corners=3pt, draw=black!45, line width=0.5pt, dashed,
      dash pattern=on 1.6pt off 1.4pt]
  (6.06,2.38) rectangle (9.54,5.72);
\node[lbl, anchor=west] at (9.68,4.05) {MLP};

\node[outbox] (dlin1) at (7.80,2.90) {Linear \, {\tiny $[128, 128]$}};
\node[outbox] (dgelu) at (7.80,4.05) {GELU};
\node[outbox] (dlin2) at (7.80,5.20) {Linear \, {\tiny $[128, d_\XX]$}};

\node[attnbox, minimum width=3.0cm] (space) at (7.80,6.62)
{Design Space\\[-1pt]{\tiny $\tilde{x} \mapsto x$}};
\node[lbl] (out) at (7.80,7.90) {$x_{t+1}$};

\draw[flow] (read.north)  -- (ln3.south);
\draw[flow] (ln3.north)   -- (dlin1.south);
\draw[flow] (dlin1.north) -- (dgelu.south);
\draw[flow] (dgelu.north) -- (dlin2.south);
\draw[flow] (dlin2.north) -- node[tag, right=1pt] {$\tilde{x}_{t+1}$} (space.south);
\draw[flow] (space.north) -- (out.south);

\draw[back]
  (add2.north) -- (0,11.70)
  -- node[tag, above=0.5pt, pos=0.45] {encoded history}
     (4.75,11.70)
  -- (4.75,-0.45) -- (7.80,-0.45) -- (read.south);

\end{tikzpicture}%
}
\caption{The attentive design policy network used throughout the benchmark and
in the architecture ablation (\cref{tab:benchmark-architecture}). The same encoder
is used by the posterior network of \cref{fig:posterior_net}: each history
pair is embedded, a time embedding is added, a learned CLS token is appended,
and the $t+1$ tokens are normalised and passed through three pre-norm
transformer blocks. The CLS position is read out and decoded into an
unconstrained design $\tilde{x}_{t+1}$, which the design space of the task maps
onto the feasible set (\cref{appendix_subsect:design_spaces}). The ablation
switches the time embedding (dashed) on and off, leaving every other component
unchanged. When on, history token $s$ carries the embedding of $s$ and the CLS
token the embedding of $t$; when off, the encoder attends over an unordered set
of tokens and is permutation invariant. On location finding the network has
$663$k trainable parameters with the time embedding and $646$k without.}
\label{fig:policy_net_attentive}
\end{figure}
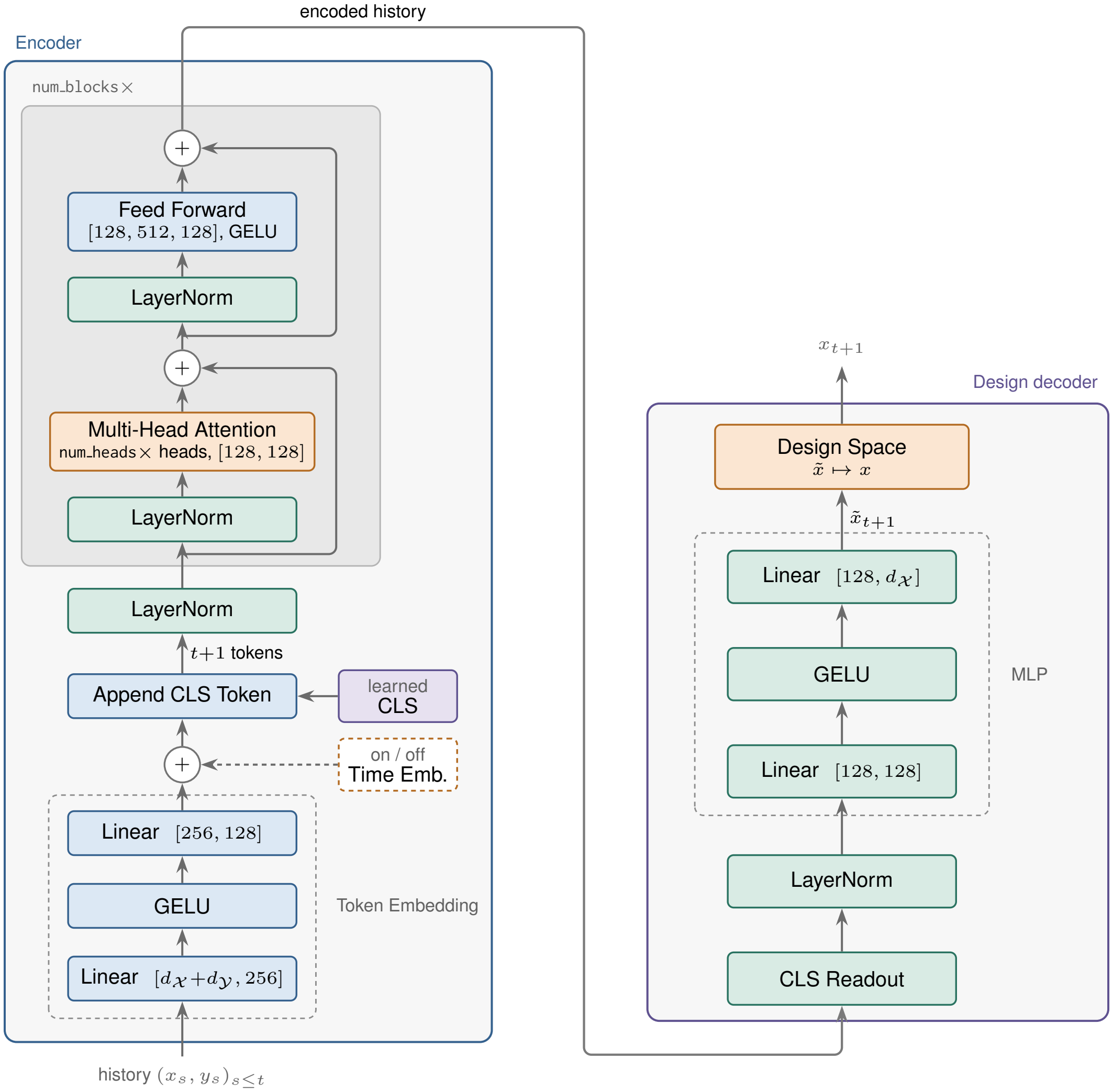

This section describes the various architectures used for our design policies. As discussed in \cref{sect:experiments}, our main benchmark uses a shared attention-based architecture across all methods. This architecture is summarised in \cref{fig:policy_net_attentive}, and is closely inspired by the TNP-style \citep{nguyen2022transformer} architecture used in ALINE \citep{huang2025aline} and other works \citep{rossa2026actionbed}. Each observation $(x_s, y_s)$ in a history $h_t = \{ (x_s, y_s) : s \leq t \}$ is concatenated to give $t$ tokens of dimension $d_\XX + d_\YY$, which are passed through a small MLP embedding block to obtain $t$ fixed-dimensional embeddings. Optionally, learned timestep embeddings are added to these token-wise embeddings. An additional, learned $t+1$'th embedding is concatenated to these. The full stack of $t+1$ embeddings is passed through a transformer block, and the resulting encoding for the $t+1$'th position is used as a readout token, which is passed through a small MLP before being decoded into the resulting design $x_{t+1}$.

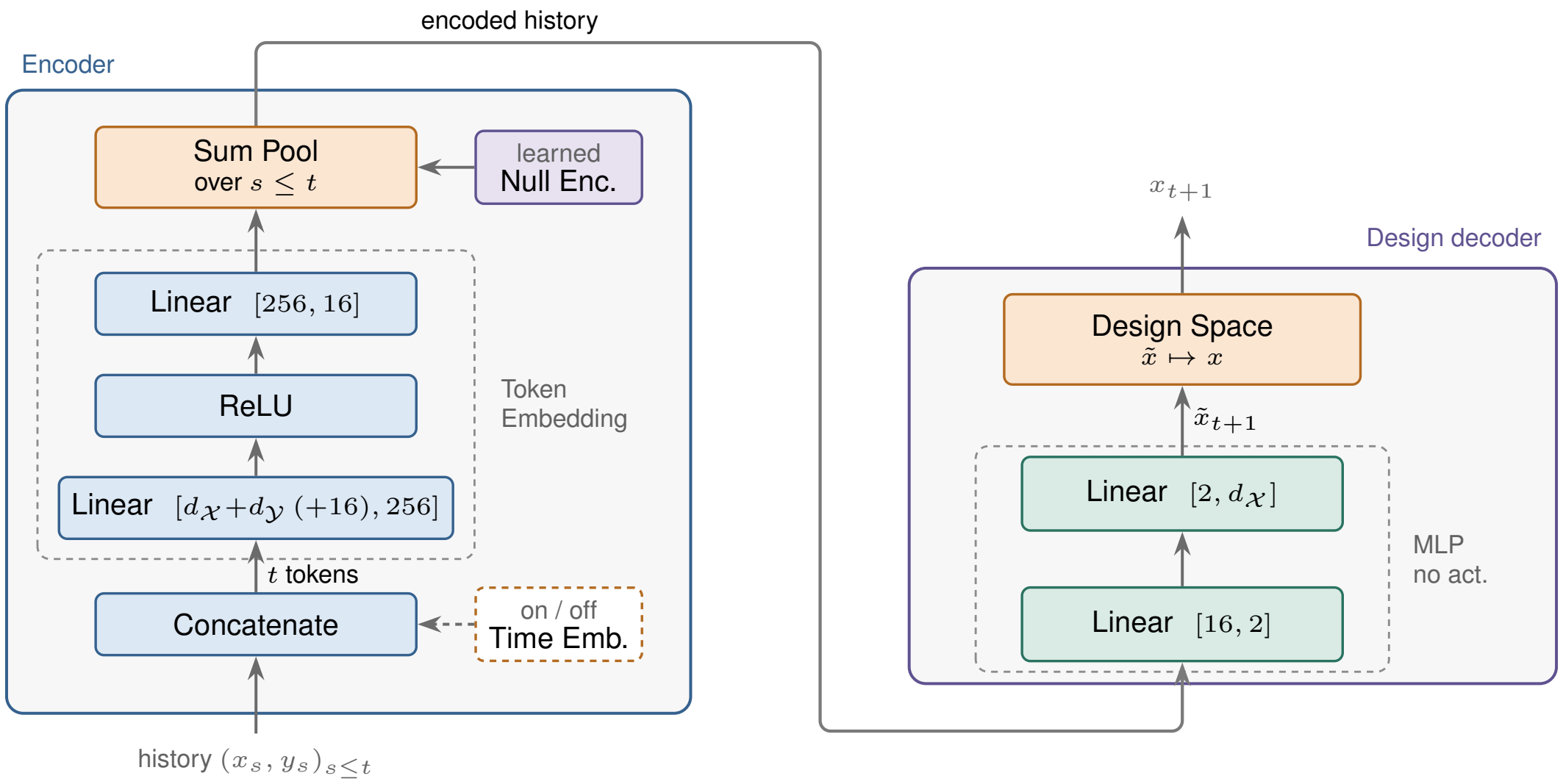
\begin{figure}[!t]
\centering
\resizebox{\linewidth}{!}{%
\begin{tikzpicture}[
  font=\sffamily,
  >={Stealth[length=2.0mm,width=1.4mm]},
  box/.style={
    rounded corners=2pt,
    draw,
    line width=0.6pt,
    align=center,
    inner sep=3pt,
    text centered,
    font=\sffamily\scriptsize,
    minimum width=2.7cm,
    minimum height=0.62cm
  },
  bar/.style={box, fill=polfill, draw=polline},
  embbox/.style={box, fill=taskfill, draw=taskline},
  attnbox/.style={box, fill=objfill, draw=objline, minimum height=0.76cm},
  clsbox/.style={box, fill=evfill, draw=evline, minimum width=1.4cm, minimum height=0.60cm},
  timebox/.style={clsbox, fill=white, draw=objline, dashed, dash pattern=on 1.6pt off 1.4pt},
  outbox/.style={box, fill=polfill, draw=polline},
  flow/.style={->, line width=0.7pt, draw=black!58},
  toggle/.style={flow, dashed, dash pattern=on 1.6pt off 1.4pt},
  back/.style={->, line width=0.75pt, draw=black!52, rounded corners=2.5pt},
  lbl/.style={font=\sffamily\tiny, text=black!62},
  tag/.style={font=\sffamily\tiny, inner sep=1.2pt}
]

\node[lbl] (inp) at (0,-0.72) {history $(x_s, y_s)_{s \le t}$};

\draw[rounded corners=4pt, fill=black!3, draw=taskline, line width=0.7pt]
  (-2.10,-0.30) rectangle (3.66,4.95);
\node[lbl, anchor=west, text=taskline] at (-2.10,5.17) {Encoder};

\node[embbox, minimum height=0.52cm] (cat) at (0,0.45) {Concatenate};
\node[timebox] (temb) at (2.55,0.45) {\sd{on / off}\\[-1pt]Time Emb.};

\draw[rounded corners=3pt, draw=black!45, line width=0.5pt, dashed,
      dash pattern=on 1.6pt off 1.4pt]
  (-1.84,1.00) rectangle (1.84,3.60);
\node[lbl, anchor=west, align=left] at (1.94,2.29) {Token\\Embedding};

\node[embbox, minimum height=0.52cm] (lin1) at (0,1.43) {Linear \, {\tiny $[d_\XX{+}d_\YY\,({+}16), 256]$}};
\node[embbox, minimum height=0.52cm] (relu) at (0,2.29) {ReLU};
\node[embbox, minimum height=0.52cm] (lin2) at (0,3.15) {Linear \, {\tiny $[256, 16]$}};

\node[attnbox, minimum height=0.64cm] (pool) at (0,4.30) {Sum Pool\\[-1pt]{\tiny over $s \le t$}};
\node[clsbox] (null) at (2.55,4.30) {\sd{learned}\\[-1pt]Null Enc.};

\draw[flow] (inp.north)   -- (cat.south);
\draw[toggle] (temb.west) -- (cat.east);
\draw[flow] (cat.north)   -- node[tag, right=1pt, pos=0.35] {$t$ tokens} (lin1.south);
\draw[flow] (lin1.north)  -- (relu.south);
\draw[flow] (relu.north)  -- (lin2.south);
\draw[flow] (lin2.north)  -- (pool.south);
\draw[flow] (null.west)   -- (pool.east);

\draw[rounded corners=4pt, fill=black!3, draw=evline, line width=0.7pt]
  (5.50,-0.05) rectangle (10.95,3.45);
\node[lbl, anchor=east, text=evline] at (10.95,3.69) {Design decoder};

\draw[rounded corners=3pt, draw=black!45, line width=0.5pt, dashed,
      dash pattern=on 1.6pt off 1.4pt]
  (6.06,0.05) rectangle (9.54,1.95);
\node[lbl, anchor=west, align=left] at (9.62,1.00) {MLP\\no act.};

\node[outbox] (dlin1) at (7.80,0.45) {Linear \, {\tiny $[16, 2]$}};
\node[outbox] (dlin2) at (7.80,1.55) {Linear \, {\tiny $[2, d_\XX]$}};

\node[attnbox, minimum width=3.0cm] (space) at (7.80,2.85)
{Design Space\\[-1pt]{\tiny $\tilde{x} \mapsto x$}};
\node[lbl] (out) at (7.80,4.10) {$x_{t+1}$};

\draw[flow] (dlin1.north) -- (dlin2.south);
\draw[flow] (dlin2.north) -- node[tag, right=1pt] {$\tilde{x}_{t+1}$} (space.south);
\draw[flow] (space.north) -- (out.south);

\draw[back]
  (pool.north) -- (0,5.35)
  -- node[tag, above=0.5pt, pos=0.45] {encoded history}
     (4.75,5.35)
  -- (4.75,-0.45) -- (7.80,-0.45) -- (dlin1.south);

\end{tikzpicture}%
}
\caption{The DeepSets design policy network of the architecture ablation
(\cref{tab:benchmark-architecture}), following the location finding network of
\citet{foster2021deep}. Each history pair is embedded by a shared per-step MLP
and the embeddings are sum-pooled into a $16$-dimensional summary, which a
decoder maps to an unconstrained
design $\tilde{x}_{t+1}$ before the design space of the task maps it onto the
feasible set (\cref{appendix_subsect:design_spaces}). An empty history is
encoded by a learned null embedding, which is distinct from \citet{foster2021deep} who fix it to zero.
The ablation switches the time embedding (dashed) off, leaving every
other component unchanged. When on, a learned $16$-dimensional embedding of $s$
is concatenated onto history pair $s$ before the per-step MLP, so it passes
through the nonlinearity and reordering the history changes the pooled
summary; when off, the encoder is permutation invariant. On location finding
the network has $11.4$k trainable parameters with the time embedding and
$5.2$k without.}
\label{fig:policy_net_deepset}
\end{figure}

In \cref{subsect:arch_experiments}, we perform an architecture ablation in which we replace this attention-based architecture with a simpler DeepSets-based architecture \citep{foster2021deep, zaheer2017deepsets}. This is summarised in \cref{fig:policy_net_deepset}. Each observation $(x_s, y_s)$ in a history $h_t = \{ (x_s, y_s) : s \leq t \}$ is again concatenated to give $t$ tokens of dimension $d_\XX + d_\YY$. Optional, learned time embeddings are concatenated with these input tokens. Each of these tokens is passed through a small MLP, and the resulting embeddings are summed over. The result is then passed through a small MLP which decodes this summed embedding into the subsequent design. The $t=1$ case is handled by a learnable null encoding.

\FloatBarrier

\section{Additional Empirical Results}
\label{appendix_sect:additional_results}

This section contains additional experimental results. \cref{appendix_subsect:full_benchmark} expands on the results in our main benchmark in~\cref{subsect:experiments_benchmark} by providing tabular versions of our figures, as well as additional ablations not shown in the main paper. %
\cref{appendix_subsect:var_post_ablation} studies the choice of variational family in JointVarPost.

\subsection{Full Benchmarking Results}
\label{appendix_subsect:full_benchmark}

For our benchmark in \cref{subsect:experiments_benchmark}, we train each method on five random seeds, and report results averaged over these runs. 

In \cref{tab:benchmark}, we present both sNMC upper bounds and sPCE lower bounds of the EIG.  These bounds are estimated using $L=10^{9}$ contrastive samples over $N=2\,048$ rollouts for all designers considered. These yield tight estimates, with the difference between our lower and upper bounds being smaller than the standard error across seeds in many cases.

\cref{tab:posthoc_actions} similarly reports the MSE achieved by the post-hoc action policy fit on top of the frozen designers, as described in \cref{appendix_sect:downstream_losses}. The final loss is estimated using $N=8\,192$ rollouts. For the ActionBED variants, we re-fit the action policy from scratch, discarding the action policy used in the training of ActionBED itself. This is done for the sake of a fair comparison, eliminating any benefit ActionBED may have gained from co-training its design and action networks. We note this then offers a pessimistic view of ActionBED's MSE performance.

We also study several loss functions for ActionBED. Our results in \cref{subsect:experiments_benchmark} use ActionBED (MSE) and ActionBED (CMSE), except on CES and temporal discounting, where we report ActionBED (MSE-L) and ActionBED (CMSE-L). This is because directly predicting $\theta$ on these two tasks is impractical, as there $\theta$ consists of sub-variables which vary widely in terms of their scaling and signal-to-noise-ratios, making MSE an inappropriate metric. On both of these tasks, the prior $p(\theta)$ has components which are log-normal, making a log-scaling natural.

\begin{table}[t]
\centering
\caption{Lower and upper bounds on the EIG (mean $\pm$ standard error over 5 seeds), estimated using $L=10^9, N=2\,048$. All policy-based methods are trained for a fixed 12-hour wall-clock budget, and use an attention-based design policy architecture.}
\vspace{0.2cm}
\label{tab:benchmark}
\scriptsize
\begin{tabular}{@{}lcc@{\hspace{2em}}lcc@{}}
\toprule

\multicolumn{3}{c}{\textbf{Location Finding}}
&
\multicolumn{3}{c}{\textbf{Pharmacokinetic}} \\
\cmidrule(r{2em}){1-3}
\cmidrule(){4-6}

Method & Lower & Upper
& Method & Lower & Upper \\
\cmidrule(r{2em}){1-3}
\cmidrule(){4-6}

Random
& \MeanSD{8.300}{0.019}
& \MeanSD{8.300}{0.019}
&
Random
& \MeanSD{3.370}{0.018}
& \MeanSD{3.370}{0.018} \\

Static
& \MeanSD{8.867}{0.014}
& \MeanSD{8.867}{0.013}
&
Static
& \MeanSD{4.053}{0.018}
& \MeanSD{4.053}{0.018} \\

\addlinespace[3pt]

DAD
& \MeanSD{14.334}{0.223}
& \MeanSD{14.359}{0.228}
&
DAD
& \MeanSD{4.079}{0.008}
& \MeanSD{4.079}{0.008} \\

JointVarPost ($m=10$)
& \MeanSD{17.848}{0.198}
& \MeanSD{18.775}{0.298}
&
JointVarPost ($m=10$)
& \MeanSD{4.041}{0.014}
& \MeanSD{4.041}{0.014} \\

JointVarPost ($m=1$)
& \MeanSD{17.895}{0.165}
& \MeanSD{18.879}{0.270}
&
JointVarPost ($m=1$)
& \MeanSD{3.721}{0.028}
& \MeanSD{3.721}{0.028} \\

ALINE
& \MeanSD{14.693}{0.033}
& \MeanSD{14.723}{0.034}
&
ALINE
& \MeanSD{3.808}{0.005}
& \MeanSD{3.808}{0.005} \\

RL-BOED
& \MeanSD{12.188}{0.022}
& \MeanSD{12.190}{0.022}
&
RL-BOED
& \MeanSD{4.005}{0.025}
& \MeanSD{4.005}{0.025} \\

\addlinespace[3pt]

ActionBED (MSE)
& \MeanSD{12.000}{0.090}
& \MeanSD{12.000}{0.090}
&
ActionBED (MSE)
& \MeanSD{2.754}{0.167}
& \MeanSD{2.754}{0.167} \\

ActionBED (CMSE)
& \MeanSD{18.563}{0.068}
& \MeanSD{20.034}{0.114}
&
ActionBED (CMSE)
& \MeanSD{3.710}{0.058}
& \MeanSD{3.710}{0.058} \\

ActionBED (MSE-L)
& --
& --
&
ActionBED (MSE-L)
& \MeanSD{3.778}{0.044}
& \MeanSD{3.778}{0.044} \\

\addlinespace[2pt]
\cmidrule(r{2em}){1-3}
\cmidrule(){4-6}

\addlinespace[5pt]
\multicolumn{3}{c}{\textbf{Michaelis Menten}}
&
\multicolumn{3}{c}{\textbf{CES}} \\
\cmidrule(r{2em}){1-3}
\cmidrule(){4-6}

Method & Lower & Upper
& Method & Lower & Upper \\
\cmidrule(r{2em}){1-3}
\cmidrule(){4-6}

Random
& \MeanSD{4.975}{0.010}
& \MeanSD{4.975}{0.010}
&
Random
& \MeanSD{9.697}{0.102}
& \MeanSD{9.786}{0.101} \\

Static
& \MeanSD{5.978}{0.015}
& \MeanSD{5.978}{0.015}
&
Static
& \MeanSD{11.254}{0.137}
& \MeanSD{11.350}{0.148} \\

\addlinespace[3pt]

DAD
& \MeanSD{6.010}{0.009}
& \MeanSD{6.010}{0.009}
&
DAD
& \MeanSD{12.636}{0.821}
& \MeanSD{12.705}{0.853} \\

JointVarPost ($m=10$)
& \MeanSD{5.914}{0.073}
& \MeanSD{5.914}{0.073}
&
JointVarPost ($m=10$)
& \MeanSD{14.317}{0.289}
& \MeanSD{14.380}{0.295} \\

JointVarPost ($m=1$)
& \MeanSD{5.859}{0.059}
& \MeanSD{5.859}{0.059}
&
JointVarPost ($m=1$)
& \MeanSD{15.148}{0.277}
& \MeanSD{15.276}{0.292} \\

ALINE
& \MeanSD{5.984}{0.017}
& \MeanSD{5.984}{0.017}
&
ALINE
& \MeanSD{14.656}{0.040}
& \MeanSD{14.754}{0.042} \\

RL-BOED
& \MeanSD{5.946}{0.007}
& \MeanSD{5.946}{0.007}
&
RL-BOED
& \MeanSD{14.940}{0.121}
& \MeanSD{15.193}{0.183} \\

\addlinespace[3pt]

ActionBED (MSE)
& \MeanSD{5.627}{0.116}
& \MeanSD{5.627}{0.116}
&
ActionBED (MSE)
& \MeanSD{5.504}{1.299}
& \MeanSD{5.504}{1.299} \\

ActionBED (CMSE)
& \MeanSD{5.492}{0.137}
& \MeanSD{5.492}{0.137}
&
ActionBED (CMSE)
& \MeanSD{3.829}{0.550}
& \MeanSD{3.829}{0.550} \\

ActionBED (MSE-L)
& \MeanSD{5.697}{0.063}
& \MeanSD{5.697}{0.063}
&
ActionBED (MSE-L)
& \MeanSD{15.767}{0.256}
& \MeanSD{16.239}{0.317} \\

& 
& 
&  

ActionBED (CMSE-L)
& \MeanSD{13.643}{0.566}
& \MeanSD{13.677}{0.584} \\

\addlinespace[3pt]
\cmidrule(r{2em}){1-3}
\cmidrule(){4-6}

\addlinespace[5pt]
\multicolumn{3}{c}{\textbf{Temporal Discounting}}
&
\multicolumn{3}{c}{\textbf{Stochastic Pendulum}} \\
\cmidrule(r{2em}){1-3}
\cmidrule(){4-6}

Method & Lower & Upper
& Method & Lower & Upper \\
\cmidrule(r{2em}){1-3}
\cmidrule(){4-6}

Random
& \MeanSD{2.061}{0.018}
& \MeanSD{2.061}{0.018}
&
Random
& \MeanSD{2.340}{0.016}
& \MeanSD{2.340}{0.016} \\

Static
& \MeanSD{2.572}{0.011}
& \MeanSD{2.572}{0.011}
&
Static
& \MeanSD{3.982}{0.018}
& \MeanSD{3.982}{0.018} \\

\addlinespace[3pt]

DAD
& \MeanSD{4.811}{0.031}
& \MeanSD{4.811}{0.031}
&
DAD
& \MeanSD{3.952}{0.027}
& \MeanSD{3.952}{0.027} \\

JointVarPost ($m=10$)
& \MeanSD{4.612}{0.045}
& \MeanSD{4.612}{0.045}
&
JointVarPost ($m=10$)
& \MeanSD{3.890}{0.036}
& \MeanSD{3.890}{0.036} \\

JointVarPost ($m=1$)
& \MeanSD{4.638}{0.033}
& \MeanSD{4.638}{0.033}
&
JointVarPost ($m=1$)
& \MeanSD{3.930}{0.011}
& \MeanSD{3.930}{0.011} \\

ALINE
& \MeanSD{4.499}{0.057}
& \MeanSD{4.499}{0.057}
&
ALINE
& \MeanSD{2.633}{0.023}
& \MeanSD{2.633}{0.023} \\

RL-BOED
& \MeanSD{4.233}{0.022}
& \MeanSD{4.233}{0.022}
&
RL-BOED
& \MeanSD{3.891}{0.015}
& \MeanSD{3.891}{0.015} \\

\addlinespace[3pt]

ActionBED (MSE)
& \MeanSD{0.854}{0.015}
& \MeanSD{0.854}{0.015}
&
ActionBED (MSE)
& \MeanSD{3.939}{0.034}
& \MeanSD{3.939}{0.034} \\

ActionBED (CMSE)
& \MeanSD{4.238}{0.025}
& \MeanSD{4.238}{0.025}
&
ActionBED (CMSE)
& \MeanSD{3.957}{0.015}
& \MeanSD{3.957}{0.015} \\

ActionBED (MSE-L)
& \MeanSD{3.371}{0.103}
& \MeanSD{3.371}{0.103}
&
ActionBED (MSE-L)
& \MeanSD{3.895}{0.036}
& \MeanSD{3.895}{0.036} \\

ActionBED (CMSE-L)
& \MeanSD{4.712}{0.044}
& \MeanSD{4.712}{0.044} \\

\addlinespace[3pt]
\cmidrule(r{2em}){1-3}
\cmidrule(){4-6}

\addlinespace[5pt]
\multicolumn{3}{c}{\textbf{Double Pendulum}}
&
\multicolumn{3}{c}{\textbf{Cart Pole}} \\
\cmidrule(r{2em}){1-3}
\cmidrule(){4-6}

Method & Lower & Upper
& Method & Lower & Upper \\
\cmidrule(r){1-3}
\cmidrule(l){4-6}

Random
& \MeanSD{8.468}{0.019}
& \MeanSD{8.471}{0.021}
&
Random
& \MeanSD{4.728}{0.020}
& \MeanSD{4.728}{0.020} \\

Static
& \MeanSD{12.310}{0.039}
& \MeanSD{12.317}{0.041}
&
Static
& \MeanSD{7.148}{0.018}
& \MeanSD{7.148}{0.018} \\

\addlinespace[3pt]

DAD
& \MeanSD{12.045}{0.170}
& \MeanSD{12.050}{0.169}
&
DAD
& \MeanSD{7.186}{0.017}
& \MeanSD{7.186}{0.017} \\

JointVarPost ($m=10$)
& \MeanSD{11.568}{0.056}
& \MeanSD{11.572}{0.056}
&
JointVarPost ($m=10$)
& \MeanSD{7.047}{0.021}
& \MeanSD{7.047}{0.021} \\

JointVarPost ($m=1$)
& \MeanSD{11.202}{0.108}
& \MeanSD{11.206}{0.108}
&
JointVarPost ($m=1$)
& \MeanSD{6.917}{0.041}
& \MeanSD{6.917}{0.041} \\

ALINE
& \MeanSD{8.070}{0.106}
& \MeanSD{8.070}{0.106}
&
ALINE
& \MeanSD{4.736}{0.043}
& \MeanSD{4.736}{0.043} \\

RL-BOED
& \MeanSD{11.591}{0.023}
& \MeanSD{11.594}{0.022}
&
RL-BOED
& \MeanSD{6.885}{0.092}
& \MeanSD{6.885}{0.092} \\

\addlinespace[3pt]

ActionBED (MSE)
& \MeanSD{11.330}{0.180}
& \MeanSD{11.332}{0.179}
&
ActionBED (MSE)
& \MeanSD{6.721}{0.039}
& \MeanSD{6.721}{0.039} \\

ActionBED (CMSE)
& \MeanSD{11.254}{0.071}
& \MeanSD{11.256}{0.071}
&
ActionBED (CMSE)
& \MeanSD{6.725}{0.031}
& \MeanSD{6.725}{0.031} \\

ActionBED (MSE-L)
& \MeanSD{11.269}{0.062}
& \MeanSD{11.271}{0.062}
&
ActionBED (MSE-L)
& \MeanSD{6.749}{0.052}
& \MeanSD{6.749}{0.052} \\

\bottomrule
\end{tabular}
\end{table}

\begin{table}[t]
\centering
\caption{Losses attained by fitting an MLP on top of the frozen designer (mean $\pm$ standard error over five seeds), fit by minimising the MSE between the output of the MLP and $\theta$. CES and temporal discounting predict $\log \theta$ instead of $\theta$.}
\vspace{0.2cm}
\label{tab:posthoc_actions}
\resizebox{\textwidth}{!}{%
\begin{tabular}{@{}lcc@{\hspace{2em}}lcc@{}}
\toprule

\multicolumn{3}{c}{\textbf{Location Finding}}
&
\multicolumn{3}{c}{\textbf{Pharmacokinetic}} \\
\cmidrule(r{2em}){1-3}
\cmidrule(){4-6}

Method & MSE & CMSE
& Method & MSE & CMSE \\
\cmidrule(r{2em}){1-3}
\cmidrule(){4-6}

Random
& \MeanSD{0.328}{0.002}
& \MeanSD{-2.522}{0.012}
&
Random
& \MeanSD{0.940}{0.005}
& \MeanSD{-3.961}{0.013} \\

Static
& \MeanSD{0.090}{0.001}
& \MeanSD{-5.151}{0.001}
&
Static
& \MeanSD{0.624}{0.021}
& \MeanSD{-4.967}{0.060} \\

\addlinespace[3pt]

DAD
& \MeanSD{0.068}{0.015}
& \MeanSD{-6.460}{0.167}
&
DAD
& \MeanSD{0.734}{0.046}
& \MeanSD{-4.882}{0.070} \\

JointVarPost ($m=10$)
& \MeanSD{0.030}{0.001}
& \MeanSD{-8.283}{0.125}
&
JointVarPost ($m=10$)
& \MeanSD{0.523}{0.006}
& \MeanSD{-5.311}{0.019} \\

JointVarPost ($m=1$)
& \MeanSD{0.021}{0.003}
& \MeanSD{-8.592}{0.099}
&
JointVarPost ($m=1$)
& \MeanSD{0.375}{0.006}
& \MeanSD{-5.415}{0.021} \\

ALINE
& \MeanSD{0.05121}{0.00047}
& \MeanSD{-4.4137}{0.0153}
&
ALINE
& \MeanSD{0.477}{0.007}
& \MeanSD{-5.0611}{0.0340} \\

RL-BOED
& \MeanSD{0.091}{0.002}
& \MeanSD{-4.299}{0.020}
&
RL-BOED
& \MeanSD{0.716}{0.019}
& \MeanSD{-4.644}{0.068} \\

\addlinespace[3pt]

ActionBED (MSE)
& \MeanSD{0.009}{0.0004}
& \MeanSD{-6.782}{0.036}
&
ActionBED (MSE)
& \MeanSD{0.490}{0.054}
& \MeanSD{-4.635}{0.169} \\

ActionBED (CMSE)
& \MeanSD{0.019}{0.001}
& \MeanSD{-8.876}{0.081}
&
ActionBED (CMSE)
& \MeanSD{0.390}{0.013}
& \MeanSD{-5.325}{0.093} \\

ActionBED (MSE-L)
& --
& --
&
ActionBED (MSE-L)
& \MeanSD{0.726}{0.070}
& \MeanSD{-4.913}{0.096} \\

\addlinespace[2pt]
\cmidrule(r{2em}){1-3}
\cmidrule(){4-6}

\addlinespace[5pt]
\multicolumn{3}{c}{\textbf{Michaelis Menten}}
&
\multicolumn{3}{c}{\textbf{CES}} \\
\cmidrule(r{2em}){1-3}
\cmidrule(){4-6}

Method & MSE & CMSE
& Method & MSE & CMSE \\
\cmidrule(r{2em}){1-3}
\cmidrule(){4-6}

Random
& \MeanSD{145.664}{4.866}
& \MeanSD{148.363}{4.935}
&
Random
& \MeanSD{1.860}{0.138}
& \MeanSD{0.728}{0.192} \\

Static
& \MeanSD{50.198}{0.746}
& \MeanSD{51.375}{0.764}
&
Static
& \MeanSD{0.532}{0.032}
& \MeanSD{-2.540}{0.094} \\

\addlinespace[3pt]

DAD
& \MeanSD{55.331}{1.714}
& \MeanSD{56.460}{1.724}
&
DAD
& \MeanSD{0.311}{0.035}
& \MeanSD{-3.620}{0.207} \\

JointVarPost ($m=10$)
& \MeanSD{64.316}{16.983}
& \MeanSD{65.588}{17.154}
&
JointVarPost ($m=10$)
& \MeanSD{0.266}{0.027}
& \MeanSD{-3.860}{0.200} \\

JointVarPost ($m=1$)
& \MeanSD{59.419}{12.135}
& \MeanSD{60.662}{12.235}
&
JointVarPost ($m=1$)
& \MeanSD{0.141}{0.012}
& \MeanSD{-5.320}{0.189} \\

ALINE
& \MeanSD{59.15}{1.80}
& \MeanSD{60.62}{1.85}
&
ALINE
& \MeanSD{0.8604}{0.0103}
& \MeanSD{-1.894}{0.028} \\

RL-BOED
& \MeanSD{67.947}{0.712}
& \MeanSD{69.392}{0.715}
&
RL-BOED
& \MeanSD{0.534}{0.011}
& \MeanSD{-2.082}{0.073} \\

\addlinespace[3pt]

ActionBED (MSE)
& \MeanSD{53.044}{2.331}
& \MeanSD{54.561}{2.394}
&
ActionBED (MSE)
& \MeanSD{0.863}{0.083}
& \MeanSD{-1.01}{0.325} \\

ActionBED (CMSE)
& \MeanSD{54.825}{4.677}
& \MeanSD{56.449}{4.817}
&
ActionBED (CMSE)
& \MeanSD{1.127}{0.070}
& \MeanSD{-0.254}{0.141} \\

ActionBED (MSE-L)
& \MeanSD{47.175}{1.572}
& \MeanSD{48.531}{1.646}
&
ActionBED (MSE-L)
& \MeanSD{0.053}{0.004}
& \MeanSD{-5.978}{0.174} \\

& 
& 
&
ActionBED (CMSE-L)
& \MeanSD{0.178}{0.027}
& \MeanSD{-5.882}{0.187} \\

\addlinespace[3pt]
\cmidrule(r{2em}){1-3}
\cmidrule(){4-6}

\addlinespace[5pt]
\multicolumn{3}{c}{\textbf{Temporal Discounting}}
&
\multicolumn{3}{c}{\textbf{Stochastic Pendulum}} \\
\cmidrule(r{2em}){1-3}
\cmidrule(){4-6}

Method & MSE & CMSE
& Method & MSE & CMSE \\
\cmidrule(r{2em}){1-3}
\cmidrule(){4-6}

Random
& \MeanSD{0.725}{0.009}
& \MeanSD{-1.435}{0.009}
&
Random
& \MeanSD{0.00312}{0.0000254}
& \MeanSD{-7.377}{0.011} \\

Static
& \MeanSD{0.613}{0.10}
& \MeanSD{-2.569}{0.015}
&
Static
& \MeanSD{0.000555}{0.0000195}
& \MeanSD{-9.344}{0.034} \\

\addlinespace[3pt]

DAD
& \MeanSD{0.290}{0.009}
& \MeanSD{-4.918}{0.059}
&
DAD
& \MeanSD{0.000492}{0.00000842}
& \MeanSD{-9.402}{0.010} \\

JointVarPost ($m=10$)
& \MeanSD{0.318}{0.009}
& \MeanSD{-4.806}{0.058}
&
JointVarPost ($m=10$)
& \MeanSD{0.000560}{0.0000667}
& \MeanSD{-9.290}{0.062} \\

JointVarPost ($m=1$)
& \MeanSD{0.287}{0.009}
& \MeanSD{-4.974}{0.046}
&
JointVarPost ($m=1$)
& \MeanSD{0.000437}{0.0000193}
& \MeanSD{-9.244}{0.0028} \\

ALINE
& \MeanSD{0.6028}{0.0075}
& \MeanSD{-3.326}{0.017}
&
ALINE
& \MeanSD{0.0044}{0.0001}
& \MeanSD{-6.9631}{0.0369} \\

RL-BOED
& \MeanSD{0.548}{0.009}
& \MeanSD{-3.745}{0.014}
&
RL-BOED
& \MeanSD{0.000618}{0.000016}
& \MeanSD{-9.268}{0.016} \\

\addlinespace[3pt]

ActionBED (MSE)
& \MeanSD{1.264}{0.019}
& \MeanSD{-0.546}{0.029}
&
ActionBED (MSE)
& \MeanSD{0.000480}{0.0000239}
& \MeanSD{-9.339}{0.046} \\

ActionBED (CMSE)
& \MeanSD{0.233}{0.004}
& \MeanSD{-4.773}{0.035}
&
ActionBED (CMSE)
& \MeanSD{0.000442}{0.0000151}
& \MeanSD{-9.431}{0.0376} \\

ActionBED (MSE-L)
& \MeanSD{0.157}{0.004}
& \MeanSD{-4.053}{0.093}
&
ActionBED (MSE-L)
& \MeanSD{0.00042}{0.0000215}
& \MeanSD{-9.368}{0.034} \\

ActionBED (CMSE-L)
& \MeanSD{0.290}{0.011}
& \MeanSD{-5.011}{0.040} 
&
& 
& \\

\addlinespace[3pt]
\cmidrule(r{2em}){1-3}
\cmidrule(){4-6}

\addlinespace[5pt]
\multicolumn{3}{c}{\textbf{Double Pendulum}}
&
\multicolumn{3}{c}{\textbf{Cart Pole}} \\
\cmidrule(r{2em}){1-3}
\cmidrule(){4-6}

Method & MSE & CMSE
& Method & MSE & CMSE \\
\cmidrule(r{2em}){1-3}
\cmidrule(){4-6}

Random
& \MeanSD{0.00172}{0.000014}
& \MeanSD{-7.950}{0.008}
&
Random
& \MeanSD{0.00139}{0.000015}
& \MeanSD{-8.009}{0.130} \\

Static
& \MeanSD{0.000201}{0.00000762}
& \MeanSD{-9.859}{0.028}
&
Static
& \MeanSD{0.000222}{0.00000689}
& \MeanSD{-9.772}{0.039} \\

\addlinespace[3pt]

DAD
& \MeanSD{0.000200}{0.00000708}
& \MeanSD{-9.838}{0.023}
&
DAD
& \MeanSD{0.000200}{0.00000603}
& \MeanSD{-9.710}{0.0560} \\

JointVarPost ($m=10$)
& \MeanSD{0.000218}{0.0000170}
& \MeanSD{-9.867}{0.046}
&
JointVarPost ($m=10$)
& \MeanSD{0.00025}{0.00000980}
& \MeanSD{-9.716}{0.055} \\

JointVarPost ($m=1$)
& \MeanSD{0.000180}{0.0000101}
& \MeanSD{-9.955}{0.037}
&
JointVarPost ($m=1$)
& \MeanSD{0.000226}{0.00000742}
& \MeanSD{-9.715}{0.064} \\

ALINE
& \MeanSD{0.0027}{0.0000}
& \MeanSD{-7.386}{0.032}
&
ALINE
& \MeanSD{0.0025}{0.0002}
& \MeanSD{-7.640}{0.070} \\

RL-BOED
& \MeanSD{0.000286}{0.000009}
& \MeanSD{-9.595}{0.038}
&
RL-BOED
& \MeanSD{0.000255}{0.000007}
& \MeanSD{-9.657}{0.023} \\

\addlinespace[3pt]

ActionBED (MSE)
& \MeanSD{0.000146}{0.00000614}
& \MeanSD{-10.084}{0.032}
&
ActionBED (MSE)
& \MeanSD{0.000198}{0.00000271}
& \MeanSD{-9.773}{0.019} \\

ActionBED (CMSE)
& \MeanSD{0.000128}{0.00000484}
& \MeanSD{-10.084}{0.032}
&
ActionBED (CMSE)
& \MeanSD{0.000204}{0.00000226}
& \MeanSD{-9.782}{0.008} \\

ActionBED (MSE-L)
& \MeanSD{0.000136}{0.0000109}
& \MeanSD{-10.174}{0.062}
&
ActionBED (MSE-L)
& \MeanSD{0.000205}{0.00000554}
& \MeanSD{-9.737}{0.035} \\

\bottomrule
\end{tabular}
}
\end{table}

\FloatBarrier

\subsection{Variational Posterior Ablations}
\label{appendix_subsect:var_post_ablation}

Here, we study the choice of variational family in the JointVarPost method. First, we compare JointVarPost using $m=1$ versus $m=10$ mixture components in its Gaussian mixture variational family (\cref{appendix_subsect:jointvarpost}). These results are contained in \cref{tab:benchmark}. Both $m=1$ and $m=10$ both offer relatively close performance in terms of both EIG. However, using only a single mixture component tends to outperform $m=10$ in terms of downstream MSE. While the $m=10$ mixture is more expressive, the variational family only needs to approximate the posterior after all $T$ experiments have been completed, and in our tasks, the resulting posteriors are often sufficiently concentrated that a single mixture component suffices. Given the surprisingly strong performance of this method, studying the choice of variational family more closely is an important direction for future work. 

We further study the effect of transforming the latent Gaussian mixture used to define the variational family. Any $q_\psi$ gives a valid Barber-Agakov bound (\cref{eqn:ba}), but a family that respects the support of $\theta$ gives a tighter bound, and hence a better training signal for the policy. On Temporal Discounting, $\theta = (k, \alpha)$ with $\log k \sim \mathcal{N}(-4.25, 1.5^2)$ and $\alpha \sim \mathrm{HalfNormal}(2)$. We compare a $10$-component mixture placed directly on $\R^2$ with one placed on $(\log k, \alpha)$, the coordinates in which the prior on $k$ is exactly Gaussian, and pushed through an exponential map on $k$. The first wastes mass on $k \le 0$ and fits the skewed posterior on $k$ poorly. We leave $\alpha$ untransformed, since its prior has its mode at zero and $\log \alpha$ has a heavy left tail that a Gaussian mixture covers poorly.

As shown in \cref{tab:var_post_constraint_ablation}, the transformed family raises the EIG of the trained policy from $4.33$ to $4.56$ nats, several standard errors apart. The reported sPCE and sNMC bounds do not depend on $q_\psi$, so this gain comes entirely from better policy training. The two rows differ only in the posterior network. Both use the gated DeepSets design network of previous literature \citep{foster2021deep, pmlr-v267-hedman25a} on this task (\cref{tab:gated-deepset-net}), $1{,}024$ outer samples and a fixed $200$k training steps, so the values are not directly comparable to \cref{tab:benchmark}. We use the transformed family and distinct policy architecture throughout the benchmark (\cref{tab:jointvarpost-posterior-net}).

\begin{table}[ht]
    \centering
        \caption{JointVarPost on Temporal Discounting with two variational families. \emph{Unconstrained} places the Gaussian mixture directly on $\theta = (k, \alpha) \in \R^2$. \emph{Transformed} places it on $(\log k, \alpha)$ and pushes it through an exponential map on $k$, leaving $\alpha$ unchanged. Both give valid Barber-Agakov bounds, but the transformed family puts no mass on $k \le 0$ and so gives a tighter training signal. The rows differ only in the variational family: both use the gated DeepSets design network (\cref{tab:gated-deepset-net}), $1{,}024$ outer samples, a fixed $200$k training steps rather than a time budget, and are evaluated with $N = 2{,}048$ and $L = 10^6$. Mean $\pm$ standard error over five seeds.}
    \begin{tabular}{ccc}
    \toprule
        Variational family & Lower Bound & Upper Bound \\
        \midrule
         Unconstrained &  \MeanSD{4.327}{0.081} & \MeanSD{4.328}{0.081} \\
         Transformed ($\exp$ on $k$) &  \MeanSD{4.561}{0.057} & \MeanSD{4.561}{0.057}  \\
         \bottomrule    
    \end{tabular}
    \label{tab:var_post_constraint_ablation}
\end{table}

\begin{table}[thbp]
    \centering
    \caption{Architecture of the gated DeepSets design network used in the variational posterior ablation (\cref{tab:var_post_constraint_ablation}). It follows the Temporal Discounting network of previous literature \citep{foster2021deep, pmlr-v267-hedman25a}; the only difference is that the null encoding is learned rather than fixed to zero.}
    \vspace{1ex}
    \small
    \begin{tabular}{lp{0.58\linewidth}}
        \toprule
        Component & Value \\
        \midrule
        Token encoder        & MLP on the unconstrained design $\tilde{x}_s$ alone, widths $[2, 256, 256, 2 \times 16]$, Softplus after each hidden layer \\
        Outcome gating       & The two $16$-dimensional heads are mixed by the binary outcome, $y_s h^{(1)}_s + (1 - y_s) h^{(0)}_s$ \\
        Time embedding       & None \\
        Pooling              & Sum over $s \le t$; learned null encoding for the empty history \\
        Emitter              & MLP with widths $[16, 256, 256, 2]$, Softplus after each hidden layer \\
        Output               & Unconstrained design $(\tilde{x}_d, \tilde{x}_r)$, mapped to $(D_1, R_0)$ by the design space \\
        \bottomrule
    \end{tabular}
    \label{tab:gated-deepset-net}
\end{table}

\newpage
\FloatBarrier
\section{sPCE and sNMC Estimators of the EIG}
\label{appendix_sect:spce_snmc}

As outlined in \cref{sect:background}, the EIG is one of the key metrics used in this work to evaluate the quality of a design policy. However, evaluating the EIG (\cref{eqn:eig})
is generally challenging because the marginal likelihood $p(h_T \mid \pi)$ is typically intractable. Consequently, an extensive literature has developed around estimators and bounds for both evaluating and optimising the EIG (\cref{sect:background}).

Among the many approaches available, we focus on two commonly used bounds from the literature: sequential variants of the nested Monte Carlo (sNMC) and prior contrastive estimator (sPCE) \citep{foster2019variational, foster2021deep}. We use these primarily to evaluate the EIG of learned policies, while some of the methods considered also employ them as objectives during training.

Both estimators are constructed from the same set of Monte Carlo samples. We draw $N$ outer samples $(\theta_0^{(n)}, h_T^{(n)}) \sim p(\theta, h_T \mid \pi)$ and, for each outer sample, draw $L$ independent contrastive parameters $\theta_{1:L}^{(n)} \sim p(\theta)$. The two estimators differ in how these samples are used to approximate the marginal likelihood $p(h_T \mid \pi)$. In sPCE, the outer parameter $\theta_0^{(n)}$ is included alongside the $L$ contrastive samples, giving
\begin{equation} \label{eqn:pce}
    \widehat{\II}_{\mathrm{sPCE}}(\pi) = \frac{1}{N}\sum_{n=1}^N \left( \log p(h_T^{(n)} \mid \theta_0^{(n)}, \pi)
    - \log \left(\frac{1}{L+1} \sum_{\ell=0}^L p(h_T^{(n)} \mid \theta_\ell^{(n)}, \pi) \right) \right).
\end{equation}
\looseness=-1
In sNMC, by contrast, the marginal estimate is formed using only the $L$ independent contrastive samples:
\begin{equation} \label{eqn:nmc}
    \widehat{\II}_{\mathrm{sNMC}}(\pi) = \frac{1}{N}\sum_{n=1}^N \left( \log p(h_T^{(n)} \mid \theta_0^{(n)}, \pi)
    - \log \left(\frac{1}{L} \sum_{\ell=1}^L p(h_T^{(n)} \mid \theta_\ell^{(n)}, \pi) \right) \right).
\end{equation}
In expectation, sPCE provides a lower bound on the EIG, whereas sNMC provides an upper bound, with both becoming tight as $L \to \infty$. An important caveat is that sPCE is itself upper bounded by $\log(L+1)$ \citep{foster2019variational}. It can therefore saturate as either a training objective or an evaluation metric when $L$ is not sufficiently large. More generally, accurate estimation of the EIG can require very large values of $L$, making computationally and memory-efficient implementations particularly important. %

\end{document}